\documentclass[dvipsnames]{article}
\usepackage{float}
\usepackage{colm2024_conference}

\usepackage{booktabs}
\usepackage{graphicx}
\usepackage{enumitem}
\usepackage{amsmath,amssymb}
\usepackage{array,tabularx,xltabular,longtable}
\usepackage{colortbl}
\ifPDFTeX\usepackage[utf8]{inputenc}\fi
\usepackage{makecell}
\usepackage{xurl}
\usepackage{marvosym}
\usepackage[section]{placeins}

\usepackage{capt-of}

\usepackage{tcolorbox}
\tcbuselibrary{skins}
\usepackage{tabularx}

\newcolumntype{P}[1]{>{\raggedright\arraybackslash}p{#1}}
\newcolumntype{Z}{>{\raggedright\arraybackslash}X}
\newcommand{\core}{\textsuperscript{*}}
\newcommand{\corresponding}{\textsuperscript{\Letter}}
\usepackage{amsmath,amsfonts,bm}

\def\eqref#1{equation~\ref{#1}}

\def\1{\bm{1}}

\DeclareMathAlphabet{\mathsfit}{\encodingdefault}{\sfdefault}{m}{sl}
\SetMathAlphabet{\mathsfit}{bold}{\encodingdefault}{\sfdefault}{bx}{n}

\newsavebox{\firstpageplots}
\newlength{\firstpageavailable}

\renewcommand{\ghlink}{https://github.com/TraceLite-AI/TraceTrial-Bench}
\title{Trial by Trace: Benchmarking AI Agent Harnesses for Reliability and Failure Attribution}
\author{%
  \textbf{TraceLite AI}\textsuperscript{1,2,3,4,5}\\[4pt]
  {\normalfont\small
    \textsuperscript{1}Institute of Automation, CAS,\quad
    \textsuperscript{2}Zhongguancun Academy\\[2pt]
    \textsuperscript{3}Institute of Information Engineering, CAS\\[2pt]
    \textsuperscript{4}Tsinghua University,\quad
    \textsuperscript{5}Beijing Institute of Technology
  }%
}
\date{October 14, 2026}

\begin{document}
\ifdefined\pdfminorversion
  \pdfminorversion=7
\else
  
\fi
\maketitle
\begin{abstract}
LLM agents are increasingly used to solve real-world problems, and their effectiveness depends on both the underlying models and the harnesses that support their operation. However, most existing benchmarks focus on model capabilities and overall agent performance, with limited attention to the evaluation of agent harnesses. To address this gap, we introduce \textbf{TraceTrial-Bench}, which organizes harness reliability into seven components covering context, execution, governance, lifecycle, observability, tools, and verification. We construct 95 expert-level, mechanism-oriented tasks with executable or state-based ground truth to support semantically equivalent comparisons while preserving the native behavior of each harness. We design a \textbf{Trajectory Tracing (TT) module} to capture agent execution traces and introduce \textbf{a two-layer evaluation framework} that separately assesses agent performance and harness reliability. In addition, we design a \textbf{Multi-source Evidence Acquisition (MEA) module} to provide evidence from the agent context, harness, and execution environment for LLM-based evaluation, and develop a \textbf{Failure Attribution (FA) module} that uses explicit mechanism chains to support attribution at the run level. Across 4 harnesses, 9 models, and 95 tasks, our evaluation of 3,420 runs yields macro-averaged harness scores ranging from 59.8\% to 72.0\%, while 31.1\% of runs achieve task success despite failing at least one harness criterion. TraceTrial-Bench provides an auditable and reproducible foundation for comparing agent harnesses, diagnosing their failures, and studying how harness design interacts with model behavior.
\end{abstract}

\par
\begingroup

\setbox\firstpageplots=\vbox{%
    \hbox{%
        \includegraphics[width=\textwidth]
        {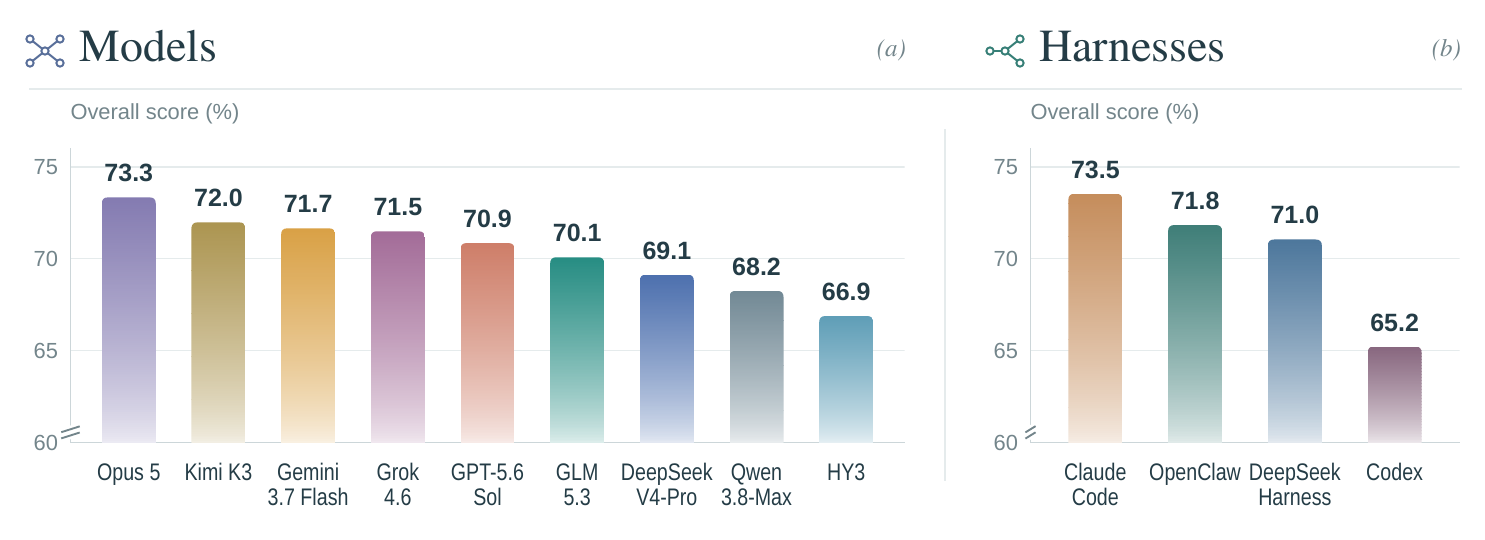}%
    }%
    \vskip 6pt
    \nointerlineskip
    \hbox{%
        \includegraphics[width=\textwidth]
        {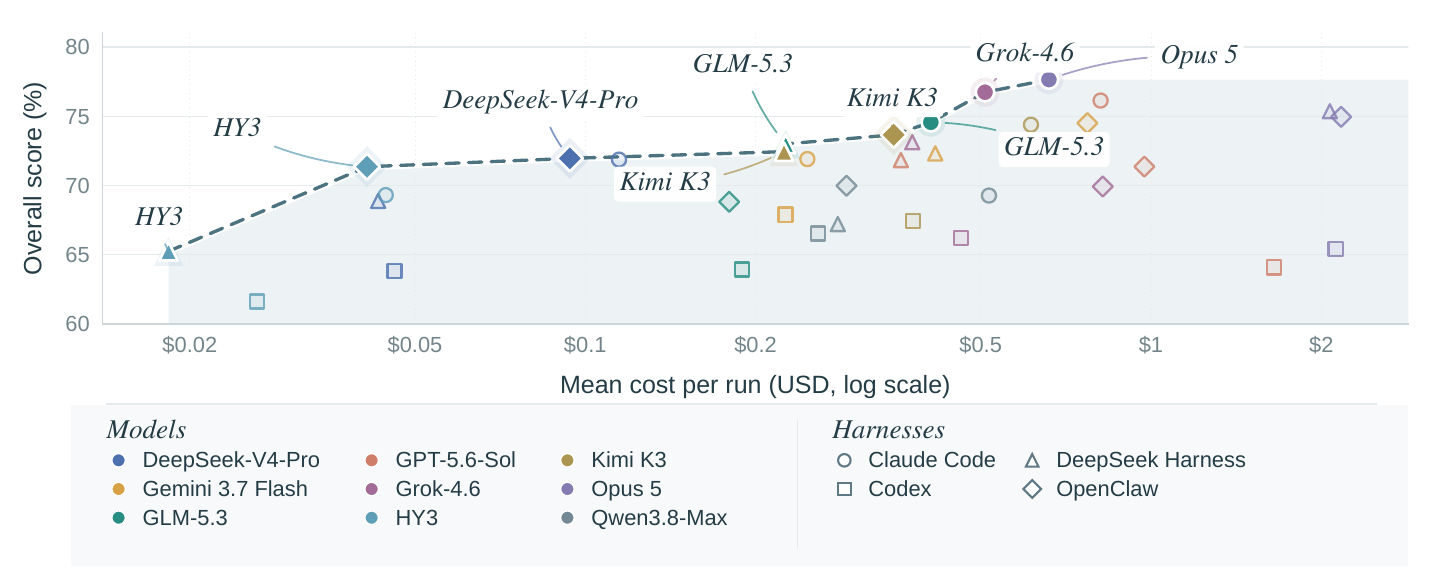}%
    }%
}

\setlength{\firstpageavailable}{%
    \dimexpr\pagegoal-\pagetotal-\pagedepth-2pt\relax
}

\nointerlineskip
\hbox to \textwidth{%
    \hfil
    \ifdim\dimexpr\ht\firstpageplots+\dp\firstpageplots
        \relax>\firstpageavailable
        \resizebox*{!}{\firstpageavailable}{%
            \usebox{\firstpageplots}%
        }%
    \else
        \usebox{\firstpageplots}%
    \fi
    \hfil
}

\endgroup
\newpage

\section{Introduction}
\label{intro}
\label{sec:intro}

\begin{figure}[t]
\centering
\includegraphics[width=\linewidth]{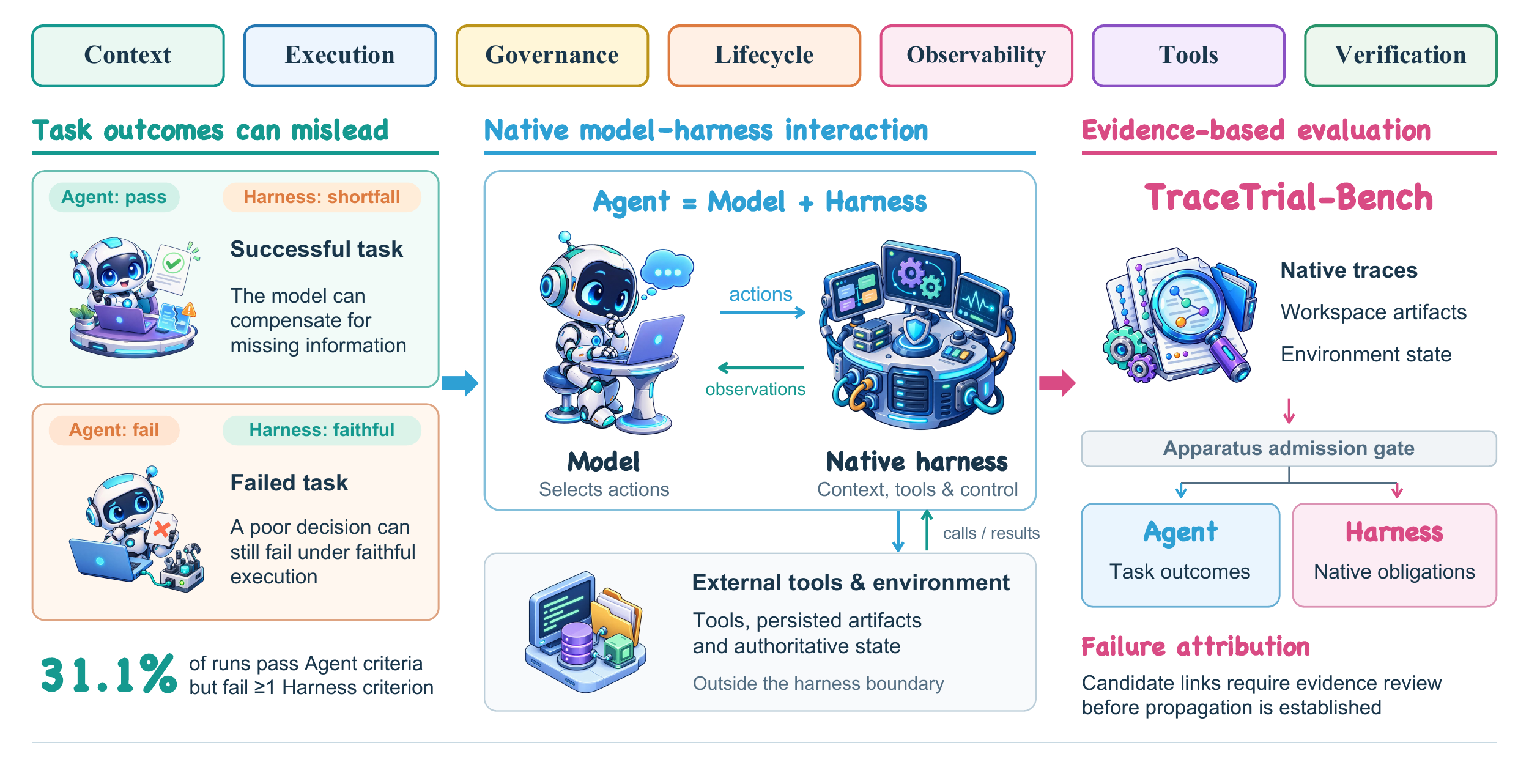}
\caption{\textbf{Background and motivation.} Agent performance depends on both the model and its harness, but task outcomes alone cannot distinguish model errors from harness shortfalls. TraceTrial-Bench uses native execution traces, artifacts, and environment state to assess agent performance and harness reliability separately and support evidence-based failure attribution.}
\label{fig:harness-motivation}
\end{figure}

Large language models have evolved from systems that primarily generate text into the foundation of AI agents that can carry out complex, multi-step tasks~\citep{wang2024survey, yao2022react, agashe2025agent}. By using tools, executing commands, modifying files, and interacting with external environments, these agents can translate user goals into concrete actions and observable outcomes~\citep{schick2023toolformer, yang2024swe, wang2025openhands}. However, an AI agent relies on more than its underlying model~\citep{sumers2023cognitive}. Its behavior also depends on system instructions, context management, available tools, execution environments, persistent state, lifecycle controls, and safety mechanisms~\citep{packer2023memgpt, wu2023autogen, mei2024aios}. As a result, the same model can behave differently when deployed through different system configurations. The \textbf{agent harness} connects the model to these surrounding components and determines what information reaches the model, which actions it can take, and how tool results, errors, and state changes are presented. Agent performance should therefore be understood as a property of the complete system, with the harness treated as a distinct and reliability-critical component~\citep{li2026agent, ning2026code}.

Most existing agent benchmarks evaluate task completion through final answers, test results, or environment states~\citep{merrill2026terminal, deng2025swe, barres2025tau}. These measures capture the combined effects of the model and its harness, but provide limited insight into whether specific harness mechanisms function reliably. As illustrated in Figure~\ref{fig:harness-motivation}, task success may conceal lost context, misleading tool results, or lifecycle violations, while task failure can occur despite faithful harness execution. Even with the model held fixed, outcome-level comparisons alone cannot identify which harness behaviors account for the observed differences~\citep{barke2026agentrx, liu2026auditing}. Recent studies have begun to examine harness configuration effects and safety, motivating complementary evaluation of individual harness mechanisms~\citep{yao2026harness, bai2026harnessrisk}. Such evaluation requires tasks that exercise explicit harness obligations, evidence that reveals whether those obligations were fulfilled, and separate assessments of harness behavior and task outcomes.

Therefore, we introduce \textbf{TraceTrial-Bench}, a mechanism-oriented benchmark that treats the native agent harness as the system under test. It assesses harness behavior through auditable execution traces and external environment states. Building on an existing harness taxonomy~\citep{li2026agent}, we organize reliability into seven components: context, execution, governance, lifecycle, observability, tools, and verification. We construct 95 tasks that exercise specific harness mechanisms through practical workflows, pairing explicit behavioral requirements with agent deliverables and executable or state-based ground truth. Cross-harness adapters preserve common task objectives and evaluation criteria while retaining each harness's native interfaces and execution behavior. Evaluating the same model--task combinations across harnesses provides a basis for comparing their overall reliability and identifying weaknesses in individual components.

Specifically, the Trajectory Tracing (TT) module reconstructs native execution records in a shared Trace Intermediate Representation (Trace IR) while preserving their provenance. A Multi-source Evidence Acquisition (MEA) module combines these records with workspace artifacts and authoritative environment state to assemble criterion-specific evidence. After apparatus integrity checks, the two-layer evaluation framework separately assesses agent task performance and harness reliability. The Failure Attribution (FA) module then connects failed harness and agent criteria through predefined mechanism chains, with further evidence review required to distinguish activated candidate links from validated failure propagation. Across four harnesses, nine models, and 95 tasks, we evaluate 3,420 selected executions. Task success and harness-criterion satisfaction yield different harness rankings, and 31.1\% of runs achieve task success despite failing at least one harness criterion. These findings show how separate assessments of task outcomes and harness behavior expose reliability differences that outcome-only evaluation can overlook.

Our contributions are summarized as follows.
\begin{itemize}[leftmargin=*]
    \item We construct TraceTrial-Bench, a 95-task benchmark covering seven reliability components while preserving native harness behavior in cross-harness comparisons.
    \item We develop a two-layer evaluation framework combining TT, MEA, and Trace IR to separately assess agent performance and harness reliability.
    \item We introduce failure attribution through explicit mechanism chains, distinguishing candidate links from failure propagation validated through evidence review.
    \item We evaluate four harnesses and nine models across 3,420 selected executions and provide tasks, configurations, traces, and verdicts for reproducible analysis.
\end{itemize}

\section{Background}
\label{sec:background}
\label{preliminaries}

\subsection{Agent, Model, and Harness}
A modern agent combines a model's decisions with a harness that manages repeated interaction with tools and the environment~\citep{wei2026agentic,zhong2026ai}. We use the abstraction
\begin{equation}
    \mathrm{Agent}=\mathrm{Model}+\mathrm{Harness}.
\end{equation}
The model interprets available information and selects actions. The harness constructs context, exposes tool interfaces, transports calls and results, maintains persistent state, and enforces execution and permission controls. Tool calls, workspace changes, and output artifacts record the effects of this interaction. Task performance therefore depends on both model decisions and the way those decisions are supported by the harness.

A scaffold usually describes prompts, workflows, and action strategies built around the model, while a runtime supplies processes, file systems, and computing resources. Our harness boundary includes their orchestration and control functions. External tool implementations, task environments, output artifacts, evaluators, and measurement apparatus remain outside that boundary. This separation is necessary when deciding whether an observation concerns a harness obligation, an Agent decision, or a failure of the measurement process.

\subsection{Harness Reliability and Task Outcomes}
Harness functions affect several parts of an execution. Message selection, ordering, truncation, compaction, and session state determine which information remains available to the model~\citep{ye2025agentfold,zhang2026agentic}. Tool inventories, parameter schemas, and routing determine the action interface~\citep{hasan2026model}. Execution and lifecycle controls communicate command outcomes, manage long-running work, and preserve continuity across sessions and delegated tasks~\citep{ma2026longhorizon}. Permissions and safety policies constrain the resulting actions~\citep{buhler2026agentbound}.

Harness reliability concerns faithful information delivery and execution semantics. It does not guarantee task success: a model can make an incorrect decision after receiving accurate information. Conversely, a correct final artifact can coexist with an incomplete result channel, a lost state transition, or a violated interaction requirement. Evaluating only the final deliverable would combine these different situations into a single outcome.

TraceTrial-Bench therefore assigns separate criteria to Agent outcomes and Harness obligations. The Agent layer checks required actions, task compliance, and final deliverables against authoritative state. The Harness layer checks whether the native system fulfilled the information and execution obligations exercised by the task. The seven-component taxonomy in Section~\ref{sec:tt-taxonomy} translates these responsibilities into observable properties.

\subsection{Evidence Boundaries and Levels of Claim}
An external result, a model-facing tool return, a native session record, and a normalized event are related observations, but they need not contain the same information. Environment state establishes what occurred. Changed artifacts identify the persisted deliverable, and native model-facing records establish what the model could observe. Provider-request evidence supports only the captured structured tool declarations, not uncaptured request bodies. Evidence from one boundary cannot automatically substitute for evidence from another.

We distinguish four levels of claim. A criterion verdict records satisfaction under the selected evidence. Confirming a Harness defect additionally requires reconciliation with the authoritative sources for that criterion. An activated candidate link records failed endpoints and a matching predeclared condition. Validating propagation further requires observable intermediate distortion, supported event order, and consideration of competing explanations. Apparatus failure is a measurement limitation rather than a Harness defect. These distinctions govern the scoring and attribution protocol developed next.

\section{TraceTrial-Bench}
\label{sec:tracetrialbench}
\label{TraceTrial-Bench}

\begin{figure}[t]
\centering
\includegraphics[width=\linewidth]{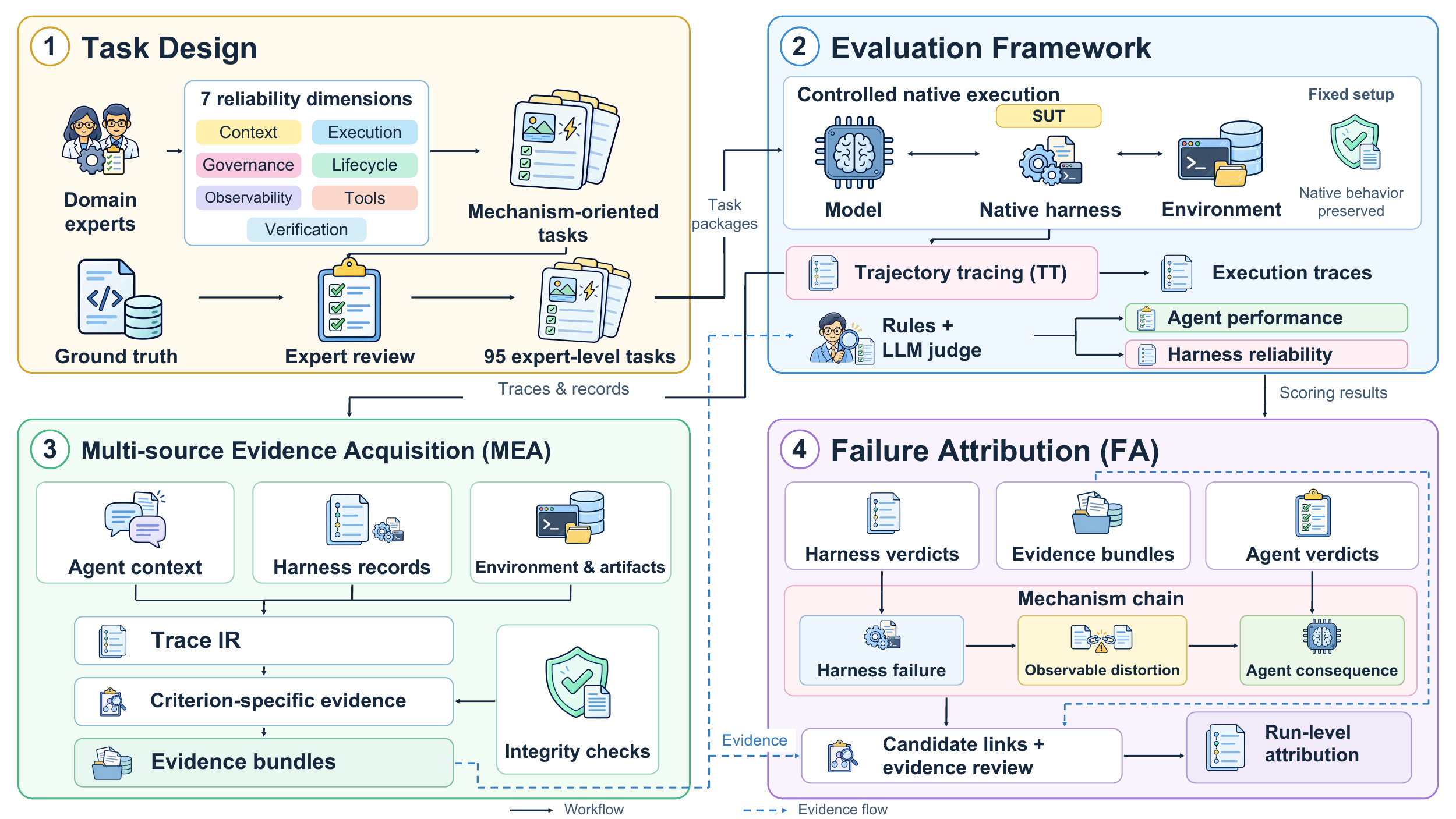}
\caption{\textbf{TraceTrial-Bench overview.} Mechanism-oriented tasks exercise native harnesses; trajectory tracing and multi-source evidence support separate Agent and Harness scores and evidence-based failure attribution. The figure's evidence-extraction stage corresponds to the MEA module.}
\label{fig:tt-bench-overview}
\end{figure}

TraceTrial-Bench evaluates an agent harness as a system, rather than treating it
as a transparent wrapper around a model (Figure~\ref{fig:tt-bench-overview}).  The benchmark asks three distinct
questions for every admitted run: (i) did the agent produce the required    
artifacts and actions, (ii) did the native harness preserve the information,
execution semantics, and lifecycle behavior required by the task, and (iii) is
an observed agent-side failure consistent with a predeclared propagation
mechanism from a Harness-side criterion failure?  The third question is addressed by automatically activating criterion-level candidate links when the corresponding Harness and Agent criteria both fail and the predeclared activation conditions are satisfied.

This chapter describes the implemented measurement pipeline for the released
runs.  It combines apparatus admission, layer-specific Agent and Harness scoring,
provenance-preserving trace reconstruction, deterministic criterion-specific
evidence selection, and criterion-level candidate-mechanism linking.  Run-level outputs summarize criterion verdicts and automatically activated candidate links. These outputs support evidence-based failure analysis, while confirming Harness defects or failure propagation requires additional validation beyond automated scoring.

\subsection{Problem Formulation and System Boundary}
\label{sec:tt-problem}

\paragraph{System under test.}
The system under test (SUT) is the \textbf{native harness}, which manages
model interactions, tool coordination, lifecycle behavior, and permissions.
The model, task, environment, and resource budgets are controlled inputs. The evaluator and the
capture machinery are measurement apparatus.  Apparatus failures are therefore
not counted as harness defects.

\paragraph{Terminology and claim levels.}
We capitalize \emph{Harness layer} and \emph{Agent layer} when referring to the
two score sheets, and use \emph{Judge} for the configured semantic grader.  We
distinguish four claim levels.  A \emph{criterion verdict} records whether the
selected evidence satisfies a declared scoring condition.  A \emph{confirmed
Harness defect} additionally reconciles a failed Harness criterion with the
question-specific authoritative source.  An \emph{activated candidate mechanism
link} records that a predeclared Harness--Agent endpoint pair and its activation
condition match the run.  A \emph{validated propagation claim} additionally
requires confirmation of both endpoint verdicts and manual review of event
order, an observable intermediate distortion, and plausible competing
explanations.  This hierarchy keeps operational scores and propagation claims
auditable without treating an internal runtime label as a causal conclusion.

For harness $h$, model $m$, task $t$, and execution instance $r$, one run is
defined as the ordered record
\begin{equation}
  R_{h,m,t,r} {=}
  (\tau,\Delta W,G,V_H,V_A,L_{\mathrm{act}}),
  \label{eq:run-unit}
\end{equation}
where $\tau$ is the normalized event sequence, $\Delta W$ is the workspace
change relative to the frozen baseline, $G$ is the apparatus-admission state,
$V_H$ and $V_A$ are criterion verdicts for the Harness and Agent layers, and
$L_{\mathrm{act}}$ is the set of activated candidate mechanism links.  The index $r$
distinguishes execution instances; the reported population retains one selected execution per cell.  Run identity additionally binds the task
source, resolved model and harness configuration, container image,
execution-instance identifier, archive manifest, and artifact digests.

\paragraph{Observation is boundary-specific.}
A nominal ``tool result'' may have several non-equivalent representations: the
tool or process produces an external result; the harness may transform it; the
model receives a model-facing payload; a native session archive records some
representation; and the evaluator observes a normalized event.  TraceTrial-
Bench consequently uses question-specific authorities.  Environment or
simulator state is authoritative for what actually happened, changed artifacts
are authoritative for final persisted artifact state, provider-request evidence is
authoritative only for the structured tool declarations it retained, and a
native model-facing result is authoritative for what the model could observe.

\begin{figure}[!htbp]
  \centering
  \resizebox{0.98\textwidth}{!}{%
  \includegraphics{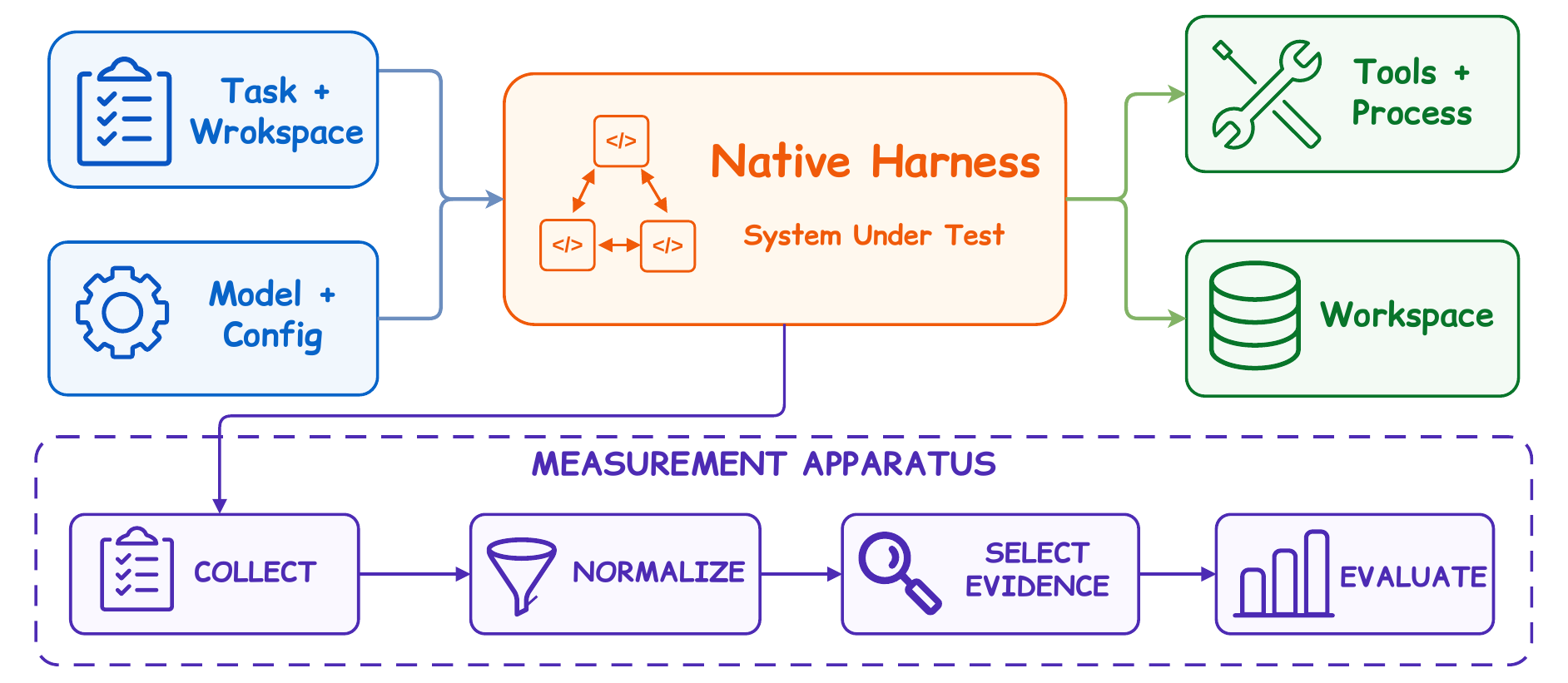}
  }
  \caption{\textbf{TraceTrial-Bench system boundary and implemented pipeline.}
  The native harness is the SUT.  Dashed arrows denote evidence acquisition;
  solid arrows denote task execution or subsequent analysis.  The evaluator
  consumes normalized evidence only after run-identity and apparatus checks.}
  \label{fig:tt-architecture}
\end{figure}

\paragraph{Admission gate and scored layers.}
The implementation has two scored layers after apparatus admission.  The Agent
layer measures task actions, workspace state, and deliverables, whereas the
Harness layer targets native harness behavior using trajectory and probe
evidence.  The two verdict vectors remain distinct; the released protocol
contains one explicitly outcome-conditioned applicability state, described in
Table~\ref{tab:tt-score-states}.  Container lifecycle, archive integrity, trace
scoreability, probe integrity, and private-oracle availability are verified
before semantic scoring is admitted.  Apparatus admission is not a third score
and cannot compensate for either scored layer.
Figure~\ref{fig:tt-architecture} shows the boundary and the resulting dataflow.

\subsection{Benchmark Design}
\label{sec:tt-design}

\subsubsection{Harness Reliability Taxonomy}
\label{sec:tt-taxonomy}

\begin{figure}[!htbp]
  \centering

  \begin{minipage}[c]{0.40\textwidth}
    \centering
    \includegraphics[width=\linewidth]
      {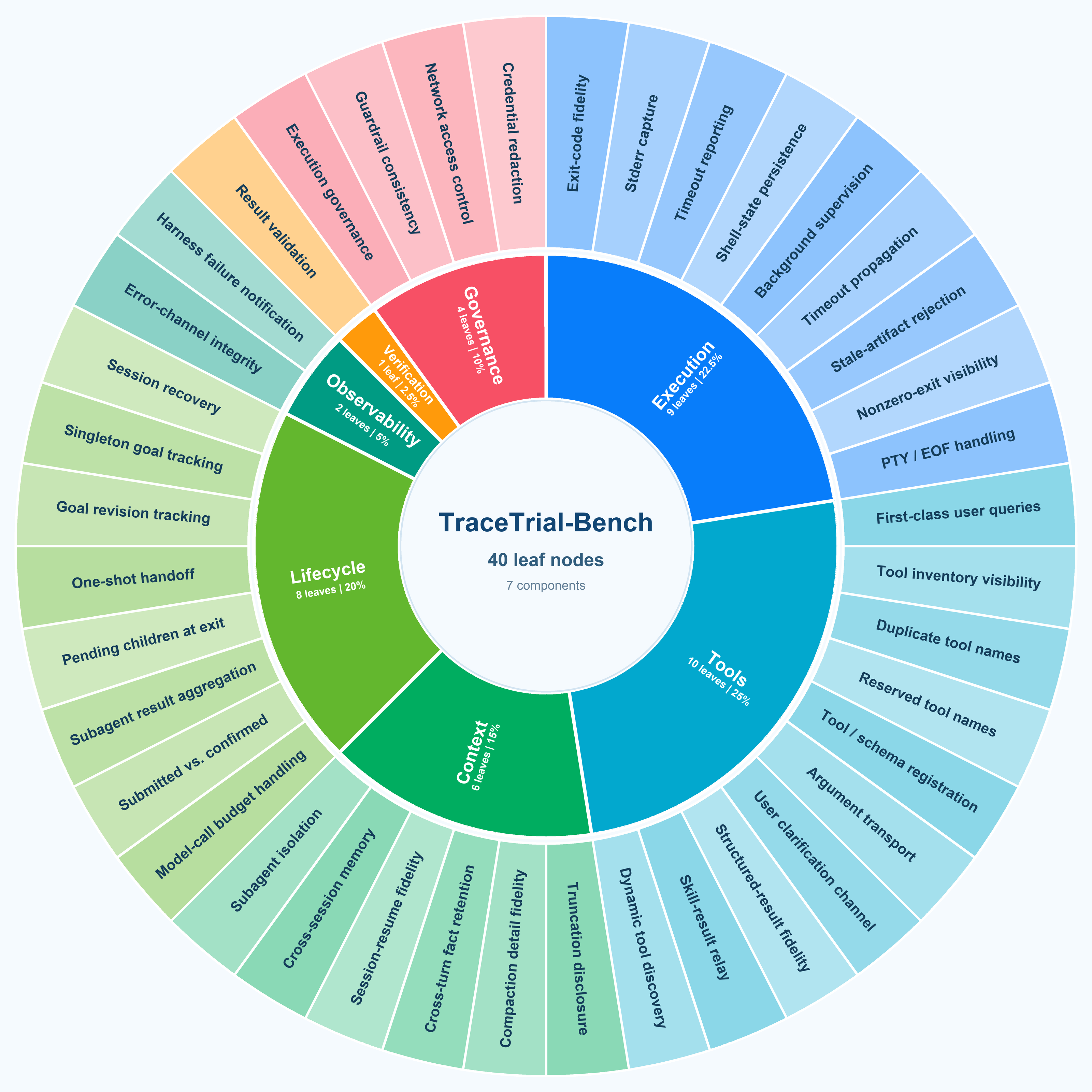}
  \end{minipage}
  \hfill
  \begin{minipage}[c]{0.57\textwidth}
    \centering
    \footnotesize

    \begin{tabularx}{\linewidth}{@{}P{0.25\linewidth}Zr@{}}
  \toprule
  Dimension & Evaluation focus & Tasks \\
  \midrule

  Context (\textsc{C})
  & Information preservation and disclosure across boundaries.
  & 15 \\

  Execution (\textsc{E})
  & Faithful process semantics, state, and result delivery.
  & 27 \\

  Governance (\textsc{G})
  & Approvals, guardrails, secret redaction, and trust boundaries.
  & 9 \\

  Lifecycle (\textsc{L})
  & Correct state transitions, recovery, and idempotent retries.
  & 19 \\

  Observability (\textsc{O})
  & Model-visible diagnostics, replayable logs, and provenance.
  & 7 \\

  Tools (\textsc{T})
  & Valid tool interfaces, faithful transport, and discovery.
  & 17 \\

  Verification (\textsc{V})
  & Consistency of completion claims with authoritative outcomes.
  & 1 \\

  \midrule
  \textbf{Total} & & \textbf{95} \\
  \bottomrule
\end{tabularx}
  \end{minipage}

  \caption{\textbf{Taxonomy and task coverage of TraceTrial-Bench.}
  (Left) Distribution of 40 mechanism-level reliability properties
  across seven Harness components. The inner ring shows each
  component's share, and the outer ring lists individual properties,
  with each property assigned equal weight.
  (Right) Scope of each dimension and its task coverage in the
  95-task evaluation population.}
  \label{fig:tt-taxonomy-overview}
\end{figure}

The formal evaluation organizes Harness reliability into seven top-level
dimensions: Context (\textsc{C}), Execution (\textsc{E}), Governance
(\textsc{G}), Lifecycle (\textsc{L}), Observability (\textsc{O}), Tools
(\textsc{T}), and Verification (\textsc{V})~\citep{li2026agent}. The dimensions are defined by the
Harness obligation under test rather than by the application scenario in which
it is exercised.  Figure~\ref{fig:tt-taxonomy-overview} shows the property distribution
(left) and summarizes the scope of each dimension and its task
coverage in the 95-task evaluation population (right).

Cross-cutting behaviors are assigned according to the property being tested.
For example, state isolation is evaluated under Context when the question is
whether information crosses a session boundary, whereas child-agent completion
is evaluated under Lifecycle when the question concerns creation, delivery, or
termination.  A mechanism is included only when its trigger, expected behavior,
relevant observation boundary, and verdict conditions can be specified before
execution.  Coverage is then tracked separately for task assignment, native
capability availability, mechanism activation, and evidence sufficiency.
Integrity of the measurement apparatus is assessed by the admission gate and is
therefore outside the seven-dimension taxonomy.

\subsubsection{Task Suite}
\label{sec:tt-tasks}

Each task starts from a taxonomy subdimension and identifies the native
mechanism, the property under test, the model-visible consequence, and an
external truth anchor.  The scenario then exercises that mechanism through
ordinary domain work.  Each task pairs Harness-layer criteria with Agent-layer
deliverables and uses deterministic state or artifact checks wherever an oracle
is available.

A task package fixes the prompt or turn sequence, initial workspace, allowed
tools, budgets, expected outputs, deterministic verifier, two score sheets, and
optional simulator or execution/security probes.  Private simulator state and
persisted baseline snapshots remain outside the Agent's writable workspace.
Review checks prompt solvability, mechanism reachability within budget,
verifier agreement with task truth, hidden leakage, and equivalent domain
semantics across eligible harnesses.  The resulting 95-task population is the
formal evaluation set.

The release includes 46 single-turn, 45 fixed multi-turn, and 4 simulator-driven tasks. Task triggers include oversized outputs requiring recovery, nonzero exits paired with success-shaped artifacts, pending jobs, facts crossing session boundaries, and private values available only through an interaction channel. These conditions make the obligation observable while retaining a practical task and an independent truth anchor.

\begin{table}[t]
\centering
\caption{\textbf{Task contracts and evidence boundaries.} Each task combines a Harness obligation with an Agent deliverable and an independent truth anchor. The detailed catalog is in Appendix~\ref{app:task-cards-oracles-gates}.}
\label{tab:task-contract-summary}
\begin{tabularx}{\linewidth}{@{}P{0.23\linewidth}ZZ@{}}
\toprule
Contract element & Fixed before execution & Evaluation role \\
\midrule
Task and pressure & Prompt, fixtures, mechanism trigger, and resource limits & Exercise a declared native behavior within a realistic task. \\
Agent score sheet & Required actions and final deliverables & Check task completion from artifacts and authoritative state. \\
Harness score sheet & Native obligation, observation boundary, and evidence selector & Check faithful information delivery and execution behavior. \\
Oracle and provenance & Private state, workspace baseline, and run identity & Separate environment truth from model-facing claims. \\
\bottomrule
\end{tabularx}
\end{table}

\subsubsection{Harnesses and Cross-Harness Execution}
\label{sec:tt-cross-harness}

The release evaluates Codex, Claude Code, OpenClaw, and DeepSeek Harness through
source-specific adapters.  The runner freezes the task revision, harness
adapter, requested and resolved model, reasoning setting, timeout, container
image, judge model, and resource budgets.  Each adapter launches the native harness
and translates its observable records without replacing native prompt
construction, tool transport, context policy, permissions, or session
behavior.  The task contract and oracle remain fixed while each harness retains
its native execution strategy.

Semantic equivalence permits representation normalization but prohibits
changes that introduce a capability, strengthen model-visible evidence, or
alter task pass/fail conditions.  The release manifest records the selected
population and provenance of every run.  Execution status, trace completeness,
evaluation status, score, and attribution remain separate fields.

Coverage is tracked separately for task assignment, native capability availability, mechanism activation, and evidence sufficiency. A task can be present in the manifest without every selected execution exercising its target path. An unavailable named tool is not automatically a defect if the native harness supplies an equivalent usable path. If an adapter cannot preserve the task's domain semantics or required evidence, the corresponding comparison is non-comparable under the prospective protocol. The current experiment's full-grid inclusion policy is stated separately in Section~\ref{sec:exp-admission}.

\subsubsection{Execution Environment and Measurement Apparatus}
\label{sec:tt-apparatus}

Runs execute in task-scoped containers with a fixed container-internal
workspace.  The runner initializes a frozen task state, captures a private
workspace baseline, and provisions deterministic task tools or stateful
simulator services when declared.  Task metadata controls user turns, tool
calls, model calls, wall time, process containment, and other relevant resource
bounds. The resolved configuration records the effective values.

The measurement apparatus is isolated from the Agent's writable output surface.
Runner-owned lifecycle records, archive manifests, provider-request metadata,
private simulator ledgers, baselines, and execution/security probes carry
explicit provenance and digest bindings.  The release manifest serves as the
authoritative index connecting task, configuration, run identity, evidence,
and checksums.

\subsection{Trajectory Tracing (TT) and Multi-source Evidence Acquisition (MEA)}
\label{sec:tt-evidence}

\subsubsection{Trace Collection and Normalization}
\label{sec:tt-trace-collection}

\paragraph{Harness-specific trace acquisition.}
Native Harnesses expose different observability interfaces, so trace
acquisition is implemented through source-specific adapters.  The Codex
adapter reads its JSON event stream. The Claude Code adapter combines
\texttt{stream-json} output with the session record. The OpenClaw adapter reads
local or gateway-backed JSON output together with session JSONL. The
DeepSeek Harness adapter consumes its persistent JSON-RPC notification stream.
Despite these interface differences, the archival policy is uniform.  When a
native session or notification archive is available, the runner stores a
redacted copy and its digest in the run archive.  Valid records that cannot be
normalized are preserved as provenance-tagged raw stdout or stderr events
rather than discarded.

\paragraph{Canonical event representation.}
To support comparison across Harnesses, the adapters map native observations
to a common event schema while retaining their source records.  The resulting
archive is a JSON array ordered by a contiguous \texttt{seq} index.  Each event
has a common envelope comprising \texttt{schema\_version}, \texttt{seq},
\texttt{timestamp}, \texttt{run\_id}, and \texttt{event\_type}. Optional 
type-specific fields are populated only when the native source exposes them.
For tool interactions, \texttt{tool\_call} records the requested action and
\texttt{tool\_result} records its correlated return.  When available, the
return preserves typed error status, integer exit code, duration, truncation
status, and omitted size.  The parsed source record remains in
\texttt{raw\_event}, allowing the normalization to be audited against the
native observation.

\begin{table}[!htbp]
  \centering
  \footnotesize
  \caption{Canonical event families used by the current tracer.  Optional
  fields are emitted only when the native source supports them.}
  \label{tab:tt-event-schema}
  \begin{tabularx}{\textwidth}{@{}P{25mm}P{40mm}Z@{}}
    \toprule
    Family & Representative event types & Principal normalized content \\
    \midrule
    Lifecycle & \texttt{run\_started}, \texttt{run\_finished},
      \texttt{run\_failed} & Run identity, native session, exit state,
      timeout, and trace-completeness boundary. \\
    Model surface & \texttt{model\_input}, \texttt{tool\_inventory},
      \texttt{assistant\_message}, \texttt{assistant\_delta} & Messages or
      declarations observed at a supported request boundary, assistant content,
      model identity, and finish reason. \\
    \makecell[l]{Action/\\observation} & \texttt{tool} & Correlated call ID, tool name,
      arguments, model-facing result content, error/status fields, timing, and
      the native call/result record. \\
    Stateful control & \texttt{context\_compaction},
      \texttt{subagent\_lifecycle}, \texttt{policy\_event} & Compaction
      boundary, parent/child identity, completion delivery, approval, or policy
      transitions when exposed natively. \\
    \makecell[l]{Accounting/\\archive} & \texttt{token\_usage},
      \path{native_session_archive}, \texttt{raw\_stdout},
      \texttt{raw\_stderr} & Usage, archive provenance and digest, or raw
      material that could not be normalized safely. \\
    \bottomrule
  \end{tabularx}
\end{table}

\paragraph{Call--result correlation and execution lineage.}
Reliable reconstruction requires distinguishing event linkage from event
order.  TraceTrial-Bench therefore associates each tool result
with its call using the Harness's native call identifier whenever one is
available.  Ambiguous observations---including unmatched, delayed, and
duplicate records---remain explicitly unlinked rather than being paired
heuristically.  In multi-agent runs, principal and parent-principal identifiers
record the acting agent and its immediate lineage, while native session and turn
identifiers retain boundaries exported by the Harness.  The tracer organizes events into a unified sequence for consistent
analysis while retaining native timestamps and the original ordering
within each source.
Together, these fields support analysis of execution lineage and concurrent
interleavings when the corresponding native metadata is available.

\paragraph{Provider-level tool declarations.}
Claims about tool availability should be grounded in the interface submitted at
the provider boundary rather than inferred from static Harness configuration.
The runner-side recorder therefore inspects each provider request transiently
and extracts structured \texttt{tools} or \texttt{functions} declarations.  The
archive retains normalized declaration metadata---names, descriptions, input
schemas, request order, path, upstream status, timing, and usage---but excludes
full request bodies and authorization material.  A captured inventory is
treated as authoritative for a request only when the request succeeds and the
declaration channel is present, structurally valid, and complete.  The resulting
record establishes the structured tool surface submitted by the Harness at the
captured request boundary.

\begin{figure}[!htbp]
  \centering
  \resizebox{0.97\textwidth}{!}{%
  \includegraphics{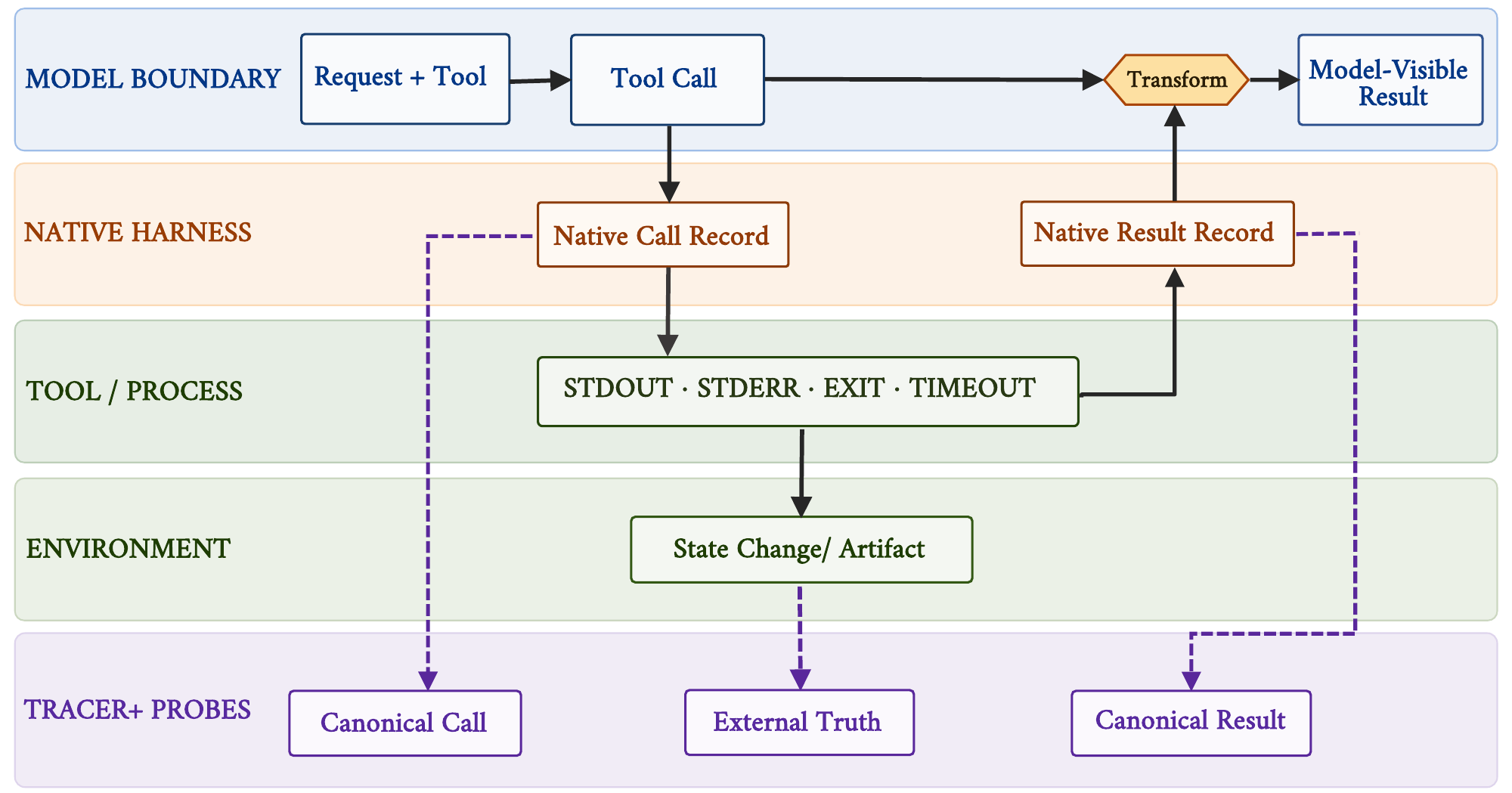}
  }
  \caption{\textbf{Cross-boundary reconstruction of a tool action.}  The
  model-facing result, native record, external execution state, and normalized
  event are related but not interchangeable.  Solid arrows show execution;
  dashed arrows terminate in separately sourced, identity-aligned evidence.}
  \label{fig:tt-trace-swimlane}
\end{figure}

Figure~\ref{fig:tt-trace-swimlane} illustrates how one semantic action is
represented at several boundaries.  The apparatus may know that the external process timed out while the
model-facing result omits that fact. Retaining both observations is
precisely what makes a harness-fidelity criterion possible.

\subsubsection{Evidence Packets and Forensic Reconstruction}
\label{sec:tt-forensic-reconstruction}

The full trace is too large and too heterogeneous to place in every semantic
Judge call.  TraceTrial-Bench therefore separates \emph{archival evidence} from
\emph{criterion evidence}.  Archival evidence is retained for audit.  A
criterion receives a bounded packet produced by a declared selector in its
score sheet.

\paragraph{Question-specific evidence authority.}
Table~\ref{tab:tt-evidence-sources} summarizes the implemented sources.
Authority is claim-specific: external state establishes what occurred,
model-facing records establish what was delivered, and correlated observations
support fidelity comparisons across the two boundaries.

\begin{table}[!htbp]
  \centering
  \footnotesize
  \caption{Evidence sources and their implemented epistemic scope.}
  \label{tab:tt-evidence-sources}
  \begin{tabularx}{\textwidth}{@{}P{31mm}P{31mm}Z Z@{}}
    \toprule
    Source & Collected by & Primary supported claim & Corroborating source for \\
    \midrule
    Simulator ledger and final state & Runner-owned private service & Exact
      requests, issued values, irreversible state, and oracle truth &
      Model-visible presentation of the corresponding result \\
    Workspace baseline and changed artifacts & Persisted scoring inputs and
      post-run workspace & Delivered file existence, structure, content, and
      unauthorized changes & Trajectory and tool evidence explaining their production \\
    Native normalized trajectory & Harness-specific adapter & Model-facing
      calls/results when the native source exposes them, lifecycle, actor
      lineage, and final messages & External process or service truth \\
    Provider-request evidence & Runner-side recorder & Successful structured
      tool declarations, request status, timing, and usage & Native delivery
      and subsequent environment effects \\
    Execution/security probes & Runner-owned post-tool observers & Process
      disposition, timeout, path, approval, and actor observations independent
      of prose reports & Model visibility through a correlated native result \\
    Agent claims & Final messages and output artifacts & What the agent asserted
      or delivered & Harness behavior and external truth through independent evidence \\
    \bottomrule
  \end{tabularx}
\end{table}

\paragraph{Declarative selectors.}
Every semantic rubric declares eligible event types, fields, literal matching
conditions, and a maximum number of returned records.  The finder records the
matched fields and keywords, making the retrieval contract deterministic and
inspectable.  Agent rubrics select changed artifacts relative to the persisted
baseline and may add supporting trajectory records. Harness rubrics
select trajectory and probe evidence, with completed Agent verdicts
supplied separately as downstream context. Digest-validated execution and security
probes carry an explicit provenance tag, so apparatus observations remain
distinguishable from native harness events.

\paragraph{Bounding without replacing the archive.}
Short strings are returned directly, while longer strings retain fixed
boundary segments and a bounded number of keyword-centered windows,
including windows around common reduction markers. Large collections, notably tool catalogs,
retain complete matching entries and append an explicit reduction record.  The
packet reports scanned, matched, and returned counts and a truncation flag.
The archived trace and native evidence remain authoritative for manual audit
and replay.  Global bounds are fixed by the evaluated source revision, and each
criterion's return limit is fixed by its archived score sheet.
The judge model cannot alter either during evaluation.

\begin{figure}[!htbp]
  \centering
  \resizebox{0.96\textwidth}{!}{%
  \includegraphics{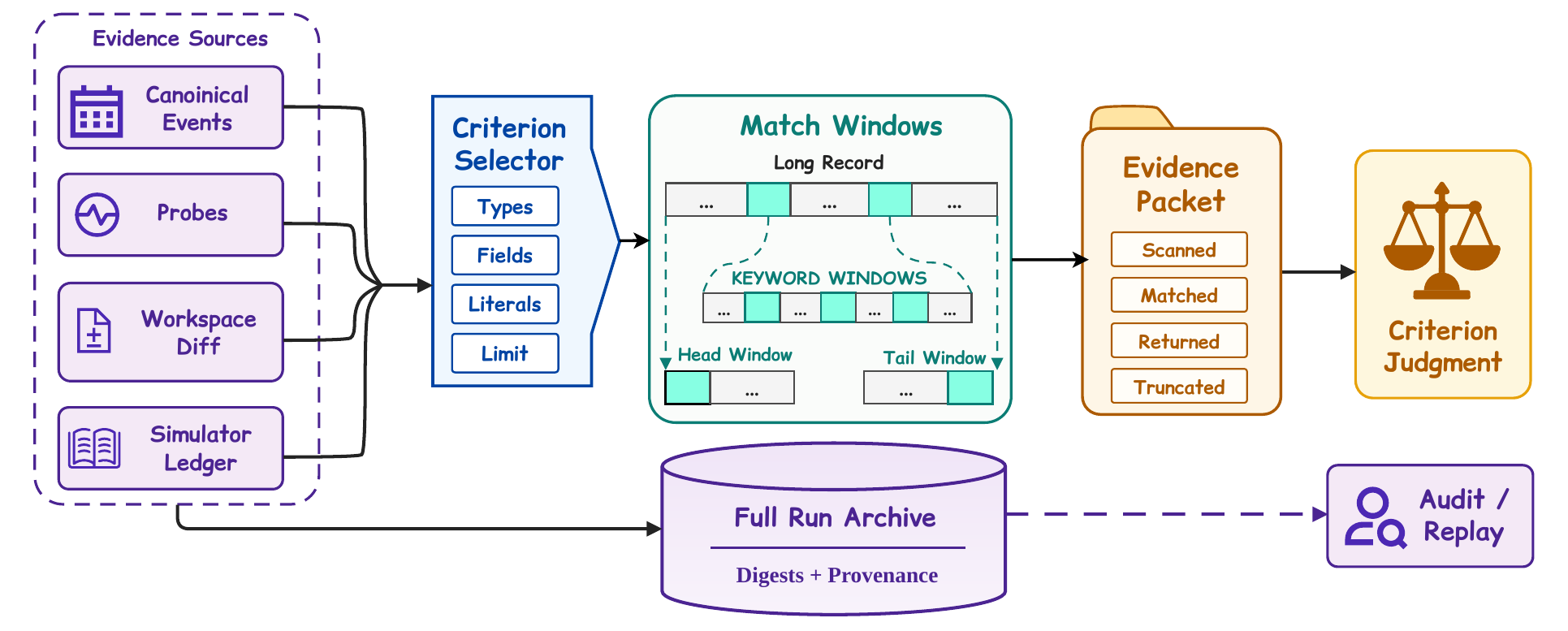}
  }
  \caption{\textbf{Implemented evidence-packet construction.}  Selection is
  deterministic and criterion-specific, while the complete archive remains
  available for audit and replay.}
  \label{fig:tt-evidence-pipeline}
\end{figure}

\paragraph{Integrity, security, and release.}
Event and summary paths must resolve to regular files inside the run's archived
artifact tree and be covered by the archive manifest.  Native archives and
probe evidence are digest-bound.  Task baselines are copied into a private
scoring-input area and checked before artifact comparison.  Known credentials
are redacted recursively by the event writer, provider request bodies are not
persisted, and authorization headers are excluded.  Together these controls
provide software-level path containment, provenance binding, and secret
minimization under the benchmark release policy.

\subsection{Two-Layer Evaluation}
\label{sec:tt-evaluation}

\paragraph{Admission before interpretation.}
A run is scoreable only when its result, resolved configuration, task,
model, and normalized harness identity agree. Lifecycle and cleanup
records must be complete, and the trace and summary must be covered
by the archive manifest. All task-declared baselines, probes, budgets,
and private-oracle records must also pass their integrity checks.
A structured failed trace is admissible only for a declared task in
which the observed lifecycle failure is itself the target property.
Execution, evaluation, and infrastructure states remain separate in
the run record. Admission establishes structural integrity and source
provenance. Criterion evaluation then records selector coverage and
the exact evidence presented for judgment.

\paragraph{Layer-specific scoring protocol.}
Evaluation distinguishes criteria that can be decided by a task-specific
verifier from those that require semantic interpretation.  Each task therefore
defines separate Harness- and Agent-layer score sheets, and designates every
criterion as either a deterministic rule or a semantic rubric.  Before any
Judge call, the deterministic outputs are checked against the declared
criterion identifiers and fixed.  This preserves objective task checks while
restricting semantic judgment to criteria for which a deterministic oracle is
not available.

Semantic criteria are evaluated individually from their criterion-specific
evidence packets.  Agent rubrics are evaluated first, using changed artifacts
as primary evidence.  Once the complete Agent-layer result has been fixed,
Harness rubrics are evaluated from trajectory and probe evidence.  Agent
verdicts are provided only as downstream context: they may describe the impact
observed in the run, but they are not evidence that a Harness criterion failed.
For every semantic criterion, the archive retains the matched packet together
with the Judge prompt, response, parsed score, and rationale.  If a required selector returns no evidence, the current implementation
assigns zero and explicitly records the empty selection. This indicates
that satisfaction was not established under the declared evidence
contract and does not independently prove a Harness defect.

\begin{table}[!htbp]
  \centering
  \footnotesize
  \caption{\textbf{Judge states and their binary resolution.}  Deterministic criteria use states 0--1, while semantic Harness criteria
may additionally use states 2--3. Every finalized criterion remains
in the layer denominator.}
  \label{tab:tt-score-states}
  \begingroup
  \setlength{\tabcolsep}{4.5pt}
  \renewcommand{\arraystretch}{1.18}
  \begin{tabularx}{\textwidth}{@{}
    >{\centering\arraybackslash}p{23mm}P{27mm}Z Z@{}}
    \toprule
    \textbf{State} & \textbf{Scope} & \textbf{Judge-side meaning} &
      \textbf{Final binary treatment} \\
    \midrule
    \cellcolor{red!8}\makecell[c]{\textbf{0}\\Unsupported} &
      Agent or Harness & The criterion is violated, or the selected evidence
      does not establish satisfaction. & Finalize as fail (0). \\
    \addlinespace[2pt]
    \cellcolor{green!9}\makecell[c]{\textbf{1}\\Supported} &
      Agent or Harness & The selected evidence establishes that the criterion
      is satisfied. & Finalize as pass (1). \\
    \addlinespace[2pt]
    \cellcolor{yellow!16}\makecell[c]{\textbf{2}\\Not exercised} &
      Harness rubric only & The native mechanism is absent, or the run did not
      reach the relevant execution path. & Resolve from the Agent outcome:
      pass when all Agent criteria pass, and fail otherwise; retain state 2. \\
    \addlinespace[2pt]
    \cellcolor{orange!12}\makecell[c]{\textbf{3}\\Malfunction} &
      Harness rubric only & The mechanism exists and was reached, but the
      Harness itself malfunctioned. & Finalize as Harness fail (0), independent
      of the Agent outcome; retain state 3. \\
    \bottomrule
  \end{tabularx}
  \endgroup
\end{table}

The two score sheets are distinct but not statistically independent because state 2 depends on the completed Agent outcome. Corpus-level reporting uses the finalized binary verdicts.  Intermediate
diagnostic codes retained in individual rubric artifacts are not analyzed as
separate experimental outcome categories.

\begin{figure}[!htbp]
  \centering
  \resizebox{0.96\textwidth}{!}{%
  \includegraphics{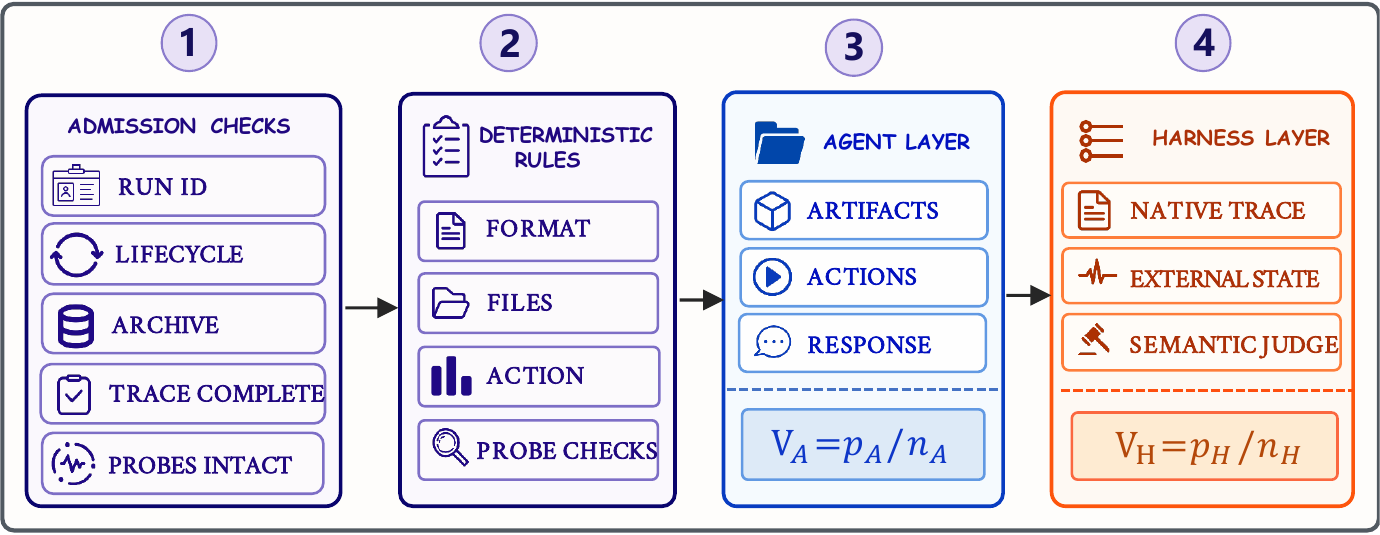}
  }
  \caption{\textbf{Admission and layered-evaluation order.}  Agent results are completed before Harness semantic judging, and the
two layers retain distinct evidence packets and verdict vectors.}
  \label{fig:tt-evaluation-flow}
\end{figure}

For layer $X\in\{H,A\}$ with $n_X$ finalized items and $p_X$ passed items, the
reported score is
\begin{equation}
  S = 50\frac{p_H}{n_H} + 50\frac{p_A}{n_A}.
  \label{eq:layer-score}
\end{equation}
Items are unweighted within a layer.  A run passes the combined evaluator only
at $S=100$, but the continuous score, item verdicts, and attribution are all
retained.  Figure~\ref{fig:tt-evaluation-flow} summarizes the actual order.

\paragraph{Validation scope and reporting claims.}
Every released run is subjected to automated checks of run identity, archive
integrity, and admission status.  Corpus-level analysis reports criterion
verdicts and automatically activated candidate links from the archived evidence.
These are operational outputs: an activated link records agreement with its
predeclared endpoints and activation condition, but does not by itself establish
that the proposed mechanism propagated a failure.

\subsection{Failure Attribution}
\label{sec:tt-attribution}

Attribution is applied only after apparatus admission and both verdict vectors
have been finalized.  It separates two questions that a single run-level label
cannot answer reliably: whether the Harness layer contains a supported failure,
and whether an Agent-layer failure follows a mechanism declared for that task.
Keeping these questions separate prevents failure co-occurrence from being
reported as propagation.

\paragraph{Failure sets and measurement bypass.}
Attribution begins with the criterion-level verdicts, but not every
Harness-layer zero represents a failure attributable to the Harness.  In some
runs, an Agent action makes the target property unobservable.  For example, a
wrapper command may replace the exit status of the command under test with its
own exit status.  Let $F_H$ and $Z_A$ denote the raw sets of zero-scored Harness
and Agent criteria, respectively, and let $B\subseteq F_H$ denote the Harness
criteria affected by such an observation bypass.  We define the Harness-side
failure set used for attribution as
\begin{equation}
  F_H=\operatorname{fail}(V_H),\qquad
  Z_H=F_H\setminus B,\qquad
  Z_A=\operatorname{fail}(V_A).
  \label{eq:attribution-zero-sets}
\end{equation}
Here, $\setminus$ denotes set difference: a criterion belongs to $Z_H$ if and
only if it is zero-scored at the Harness layer and is not marked as
bypass-affected.  Thus, $Z_H$ is obtained by removing every criterion in $B$
from the raw zero set $F_H$.
Bypassed criteria remain visible as zeros in the score report, preserving the
original evaluation outcome, but they do not contribute to $Z_H$.  Membership
in $Z_H$ is itself only an operational Harness-layer result, not a confirmed
defect. The empirical results in Section~\ref{sec:results} use the raw
set $F_H$ without this adjustment, as specified in
Section~\ref{sec:exp-analysis}. The adjusted set and stored runtime
labels remain available as separate implementation outputs.
A criterion enters the confirmed set $D_H\subseteq Z_H$ only after its
verdict has been reconciled with the evidence source appropriate to that
criterion.

\paragraph{Candidate mechanism activation.}
Before execution, a task may declare that failure of one Harness criterion can
affect a particular Agent criterion through a specified intermediate mechanism.
We represent such a declaration as
\begin{equation}
  \ell=(h_\ell,a_\ell,\mu_\ell),
  \label{eq:mechanism-chain}
\end{equation}
where $h_\ell$ and $a_\ell$ are the two criterion endpoints and $\mu_\ell$
describes the intervening information, action, or state distortion.  The implementation also stores a qualitative descriptor for review
prioritization, which is not interpreted as a calibrated probability. Task-specific declarations
define the available endpoint pairs. Some tasks assess faithful reporting of
model-visible information and do not declare a Harness-to-Agent endpoint
pair. The experimental analysis counts activated pairs stored with final
evaluations. It does not infer the size of the full candidate catalog from
the subset of links that happened to activate.

At runtime, the task declarations and applicable compatibility rules form the
candidate pool $\mathcal{C}^{\mathrm{run}}_t$.  For a declared activation
condition $q_\ell(R)$, the implementation constructs
\begin{equation}
  L_{\mathrm{act}}(R)=
  \{\ell\in\mathcal{C}^{\mathrm{run}}_t
  \mid h_\ell\in F_H,\ a_\ell\in Z_A,\ q_\ell(R)=1\}.
  \label{eq:activated-links}
\end{equation}
Activation is computed from the raw Harness set $F_H$ before bypass adjustment,
as in the released implementation.  It establishes that the declared endpoints
and activation condition are present in the run, but does not establish that the
intermediate mechanism actually propagated the failure.

\paragraph{Manual validation.}
A candidate link is promoted to the validated set
$L_{\mathrm{val}}\subseteq L_{\mathrm{act}}$ only after manual review confirms
both endpoint verdicts, reconciles the Harness endpoint with its
criterion-specific evidence source, and finds an observable intermediate
distortion in an event order consistent with the declared mechanism.  The
review also considers provenance and plausible competing explanations.  Hence
the Harness endpoint of every validated link belongs to $D_H$, and only
$L_{\mathrm{val}}$ supports a validated propagation claim.

\paragraph{Run-level operational summary.}
The Agent failures covered by activated candidate links are
\begin{equation}
  Z_{A,\mathrm{linked}}=
  \{a\in Z_A\mid \exists\ell\in L_{\mathrm{act}}:a_\ell=a\}.
  \label{eq:traced-agent-failures}
\end{equation}
TraceTrial-Bench reports $|Z_H|$ independently and partitions $Z_A$ into linked
and unlinked subsets.  These counts are recoverable from the serialized verdicts,
bypass annotations, and candidate links, allowing a run to contain both a
Harness-side zero and an unrelated Agent-side zero.  Table~\ref{tab:tt-attribution-matrix} presents seven ordered decision
rules that yield six distinct stored labels, with two rules producing \texttt{agent\_caused}.

\begin{table}[!htbp]
  \centering
  \footnotesize
  \caption{\textbf{Runtime-label decision table.}  Rows are evaluated from top
  to bottom, and the first matching row determines the stored label.  A
  ``Harness zero'' means $Z_H\neq\emptyset$; an ``outcome issue'' means
  $Z_A\neq\emptyset\lor B\neq\emptyset$.  The labels summarize operational
  evidence states rather than validated causal conclusions.}
  \label{tab:tt-attribution-matrix}
  \begingroup
  \setlength{\tabcolsep}{4pt}
  \renewcommand{\arraystretch}{1.18}
  \begin{tabularx}{\textwidth}{@{}
    >{\centering\arraybackslash}p{12mm}P{54mm}P{43mm}Z@{}}
    \toprule
    \textbf{Order} & \textbf{First matching run state} &
      \textbf{Stored label} & \textbf{Operational meaning} \\
    \midrule
    \textbf{1} & Apparatus admission fails
      ($G.\mathrm{ok}=\mathrm{false}$). &
      \cellcolor{gray!14}\texttt{inconclusive\_device} &
      Exclude the run from Harness--Agent attribution. \\
    \addlinespace[2pt]
    \textbf{2} & Either layer has no measured pass/fail criterion. &
      \cellcolor{gray!14}\texttt{inconclusive\_unmeasured} &
      At least one evaluation layer is unmeasured. \\
    \addlinespace[2pt]
    \textbf{3} & Harness zero and outcome issue, with
      $L_{\mathrm{act}}\neq\emptyset$. &
      \cellcolor{red!8}\texttt{harness\_caused} &
      A candidate Harness-to-Agent mechanism is activated; validation is
      reported separately through $L_{\mathrm{val}}$. \\
    \addlinespace[2pt]
    \textbf{4} & Harness zero and outcome issue, with
      $L_{\mathrm{act}}=\emptyset$. &
      \cellcolor{yellow!14}\texttt{agent\_caused} &
      The outcome issue is unlinked; concurrent Harness zeros remain recorded
      on the Harness axis. \\
    \addlinespace[2pt]
    \textbf{5} & Harness zero, but all Agent criteria pass and no bypass is
      recorded. &
      \cellcolor{orange!12}\texttt{harness\_defect\_absorbed} &
      The Agent delivery is complete despite an operational Harness zero;
      $D_H$ determines defect confirmation. \\
    \addlinespace[2pt]
    \textbf{6} & No Harness zero remains, but an outcome issue is present. &
      \cellcolor{yellow!14}\texttt{agent\_caused} &
      The Agent-side zero or measurement bypass is assigned to the Agent axis. \\
    \addlinespace[2pt]
    \textbf{7} & Neither a Harness zero nor an outcome issue is present. &
      \cellcolor{green!9}\texttt{clean} &
      No measured criterion is zero and no bypass is recorded. \\
    \bottomrule
  \end{tabularx}
  \endgroup
\end{table}

The stored strings are retained for compatibility with the released runtime
record and are interpreted conservatively in the paper.  In particular,
\texttt{agent\_caused} does not assert that the Harness is defect-free when
$Z_H$ is non-empty, and \texttt{harness\_caused} denotes candidate-link
activation rather than a causal conclusion.  Confirmed defect absorption requires the relevant Harness endpoint
to belong to $D_H$, and validated propagation requires
$L_{\mathrm{val}}\neq\emptyset$.

\subsection{Benchmark Outputs and Metrics}
\label{sec:tt-outputs}

\paragraph{Run-level records.}
For each run that passes admission and completes scoring, the archive preserves
three classes of records: (i) configuration and provenance, including the
resolved configuration, run result, and lifecycle and archive manifests; (ii)
execution evidence, including normalized and native traces, provider-request
metadata, applicable simulator, budget, and probe evidence, workspace changes,
and persisted scoring inputs; and (iii) evaluation artifacts, including
verifier outputs, criterion-specific evidence and Judge records, and the
combined layered score.  The release manifest associates each selected run
with its task and resolved configuration and records the corresponding artifact
paths and checksums.  The serialized result contains the apparatus state,
$V_H$, $V_A$, $L_{\mathrm{act}}$, and bypass annotations; together, these
fields determine the two-axis summary without additional heuristic inference.

\paragraph{Aggregate metrics.}
Let $\mathcal{R}_A$, $\mathcal{R}_H$, and $\mathcal{R}_{\mathrm{attr}}$ denote the reporting populations for Agent, Harness, and attribution metrics. In this study all three are the complete selected archive, as specified in Section~\ref{sec:exp-admission}. Agent-level performance is summarized by the task
success rate
\begin{equation}
  M_A=\frac{1}{|\mathcal{R}_A|}
  \sum_{R\in\mathcal{R}_A}\mathbb{I}[Z_A(R)=\emptyset],
  \label{eq:agent-success-rate}
\end{equation}
where a run is successful only if it contains no zero-scored Agent criterion.
Harness-level performance is summarized by the micro- and macro-averaged
criterion pass rates
\begin{equation}
  M_H^{\mathrm{micro}}=
  \frac{\sum_{R\in\mathcal{R}_H}p_H(R)}
       {\sum_{R\in\mathcal{R}_H}n_H(R)},
  \qquad
  M_H^{\mathrm{macro}}=
  \frac{1}{|\mathcal{R}_H|}
  \sum_{R\in\mathcal{R}_H}\frac{p_H(R)}{n_H(R)}.
  \label{eq:harness-pass-rates}
\end{equation}
The micro average assigns equal weight to each finalized Harness criterion,
whereas the macro average assigns equal weight to each run.  Because $p_H(R)$
is derived from finalized verdicts, both quantities incorporate the prescribed
outcome-conditioned resolution of raw state 2 in
Table~\ref{tab:tt-score-states}.  They therefore measure operational
Harness-criterion satisfaction and should not be interpreted as confirmed-
defect rates.

For $\mathcal{R}_{\mathrm{attr}}$, we report three complementary proportions:
(i) runs with at least one activated candidate link
($L_{\mathrm{act}}\neq\emptyset$); (ii) runs with at least one raw Harness-layer
zero but no Agent-layer zero ($F_H\neq\emptyset$ and $Z_A=\emptyset$); and
(iii) runs containing at least one unlinked Agent zero in
$Z_A\setminus Z_{A,\mathrm{linked}}$.  Measurement-quality records---including admission outcomes, trace completeness, and selector match counts---remain separate from performance aggregates. The current analysis retains all finalized cells and does not estimate corpus-level selector recall, Judge agreement, or candidate-link precision.

\section{Experiments}
\label{sec:experiments}

We compare four native agent Harnesses over a fully populated matrix of
95 tasks and nine model configurations. Each cell contains one selected
execution and its final archived evaluation. The experiment asks whether
outcome quality and Harness-criterion satisfaction vary together when the
model and task population are held fixed. This section defines the data
freeze and analysis protocol. Section~\ref{sec:results} reports the results.

\subsection{Evaluated Systems and Experimental Population}
\label{sec:exp-systems}
\label{sec:exp-objectives}

\paragraph{Harnesses and models.}
The evaluated Harnesses are Claude Code, Codex, DeepSeek Harness, and OpenClaw.
Source-specific adapters provision the task and archive native evidence while
retaining each Harness's execution and tool-use interfaces. The nine model
groups are DeepSeek-V4-Pro, Gemini 3.7 Flash, GLM-5.3, GPT-5.6-Sol, Grok-4.6,
HY3, Kimi K3, Opus 5, and Qwen3.8-Max. Reasoning settings follow the archived configurations. Original model identifiers, resolved configurations, and invocation
fields remain in the run records, while the nine names above serve
as the reporting groups used throughout the results.

\paragraph{Task population.}
The analysis covers 95 tasks, each evaluated across all 36
model--Harness combinations.
Taxonomy assignments are recovered from the task paths recorded with the runs
and normalized as described in Appendix~\ref{subsec:A_1}: the four legacy
ask-user tasks are assigned to Tools, while the task measuring truthful
timeout and result handling is classified under Execution.
The resulting population contains 15 Context, 27 Execution,
9 Governance, 19 Lifecycle, 7 Observability, 17 Tools, and 1 Verification task,
covering 40 mechanism-level properties. The generated task ledger records the
exact membership and metadata sources used in this analysis.

\paragraph{Factorial matrix and selection.}
The data freeze contains
\begin{equation}
4\ \text{Harnesses}\times9\ \text{models}\times95\ \text{tasks}
\times1\ \text{selected execution}=3{,}420\ \text{cells}.
\label{eq:exp-factorial-design}
\end{equation}
There are 855 cells per Harness and 380 per model. The unit of analysis is
the root task--model--Harness run directory in \texttt{outputs/traces}.
Nested \texttt{result.json} files emitted by tools or simulators are not
additional experimental runs. Every selected cell has a final
\texttt{combined\_score.json}, and no model block is pending and no missing score
is imputed. The archive contains selected outcomes, including replacement
executions from earlier recovery, rather than independent repetitions
per cell. We therefore describe the observed matrix and do not estimate
within-cell repeatability.

\subsection{Execution and Evaluation Configuration}
\label{sec:exp-protocol}
\label{sec:exp-evaluation}

\paragraph{Execution environment and budgets.}
Runs execute in Docker with the task workspace mounted at
\texttt{/workspace}. Resolved configurations and container records retain
the Harness invocation and image identity. Task packages define the prompt or
turn sequence, fixture, tools, expected deliverables, verifier, and resource
limits. The recorded effective wall-time limits span 120--5,400 seconds.
The corresponding task definitions span 1--120 user-input turns and
5--300 task-tool calls. These are task-specific ceilings, not observed
turn or call counts. The per-run resolved limit, rather than an example
command or directory name, identifies the applicable execution budget.

Task scheduling and recovery conditions vary across the collected runs.
The present analysis consequently compares finalized functional outcomes
and criterion verdicts. It does not interpret duration differences as a
controlled throughput experiment. Task-specific revisions and recovery runs
also limit causal interpretations of a difference between Harnesses.

\paragraph{Fixed Judge identity.}
All 3,420 archived final evaluations use
\texttt{deepseek-v4-flash} as the separate semantic Judge. The stored
per-criterion output limits are 4,096 tokens in 1,966 evaluations,
8,192 in 1,452, and 32,000 in two. Thus Judge identity is fixed, while the
archived scoring budget is not identical across all cells. We preserve these
final verdicts and report their configuration rather than treating the
collection as a uniform Judge-budget experiment.

\paragraph{Final-score authority.}
Agent- and Harness-layer items, their binary verdicts, and activated candidate
links are read from
\texttt{artifacts/llm\_judge/combined\_score.json}.
The root \texttt{result.json} supplies execution identity and historical
metadata, but its score field is not the final-score authority. Execution status, evaluation status, and final criterion verdicts remain separate records.
We verify that every selected combined score has the expected task identity,
both nonempty layer verdict vectors, binary item values, and a total
consistent with the 50/50 formula in Equation~\ref{eq:layer-score}.

\subsection{Inclusion and Aggregation Policy}
\label{sec:exp-admission}
\label{sec:exp-analysis}
\label{sec:exp-design}

\paragraph{Full-grid inclusion.}
All 3,420 final evaluations are included exactly once using their
stored criterion values. Execution-state flags and auxiliary
annotations do not affect inclusion or scores.

\paragraph{Layer metrics.}
For run $r$, let $p_{A,r}$ and $n_{A,r}$ be the passed and total Agent
criteria, and define $p_{H,r}$ and $n_{H,r}$ analogously. Agent task success
requires $p_{A,r}=n_{A,r}$. Agent Score and Harness Macro are the means of
the per-run fractions $p_{A,r}/n_{A,r}$ and $p_{H,r}/n_{H,r}$.
Harness Micro pools passed and total criteria across runs:
$\sum_r p_{H,r}/\sum_r n_{H,r}$. All reported rates are multiplied by 100.
The Layered score averages the two per-run layer scores with equal weight.
A combined pass requires every criterion in both layers to pass; it must not
be confused with Agent-only task success.

Every Harness is evaluated on the same 855 model--task cells.
Fixed-model comparisons use all 95 tasks for each Harness; model averages use
all four Harnesses and 95 tasks. Dimension-level summaries count a task once
under its normalized primary dimension. Since criterion counts vary by task,
we report both micro and macro Harness rates and make the task population
explicit.

\paragraph{Raw failure intersections and candidate links.}
The empirical contingency uses the raw sets
$F_H=\operatorname{fail}(V_H)$ and $Z_A=\operatorname{fail}(V_A)$.
All zero-scored items contribute normally, and auxiliary bypass
annotations do not subtract items from $F_H$. This is the reporting convention for the
current experiments, distinct from the optional adjusted Harness set
$Z_H$ defined in Section~\ref{sec:tt-attribution}.
A jointly clean run has $F_H=Z_A=\emptyset$. An operational absorption
state has $F_H\ne\emptyset$ and $Z_A=\emptyset$.
Crossing the empty/nonempty status of the two sets yields four mutually
exclusive states. Their proportions use all 855 runs per Harness and sum to 100\%.
A shortfall denotes at least one failed criterion, not a runtime
execution error.

Candidate-link counts come from the stored endpoint pairs after final scoring.
We check that each stored link connects a failed raw Harness item to a failed
Agent item. A run can contain both linked and unlinked Agent failures, so these
two evidence categories can overlap. Neither co-occurring zeros nor a stored
link establishes a confirmed Harness defect or causal propagation. Runtime
class labels are retained in the analysis ledger, but the contingency is
reconstructed directly from item verdicts to maintain the declared raw-score
convention.

\paragraph{Conditional candidate-link incidence.}
For primary dimension $d$, let $J_d$ contain the runs with at least one
failed criterion in both layers, and let $L_d\subseteq J_d$ contain those
with at least one stored candidate link. We report $100|L_d|/|J_d|$.
Each run contributes once regardless of its number of failed items or links.
The pooled rate is $100\sum_d|L_d|/\sum_d|J_d|$, not an unweighted mean
of the dimension rates. This statistic measures recorded link activation
among joint failures.

\paragraph{Token usage and normalized cost.}
Resource columns report mean recorded usage per run: input, output, and
total Tokens in thousands, and estimated Agent cost in USD. We apply the
same model-specific prices published on the providers' official websites across all Harnesses. Input cost partitions ordinary input, cache reads, and cache writes
using the available counters. Unspecified cache components use the
ordinary input rate. Output cost includes recorded reasoning Tokens without
counting them again. Total cost is input plus output cost, averaged over
runs with valid recorded usage. These are standardized Token-cost estimates,
not historical invoices. Judge calls, discarded executions, payment fees,
storage, and tool charges are outside their scope.

\subsection{Reproducibility}
\label{sec:exp-reproducibility}

Outcome aggregates can be recomputed from the final score files, their task and model identities, and each task's primary component. Native traces, resolved configurations, task packages, and scoring evidence support inspection of individual verdicts. Resource estimates additionally require recorded usage and the frozen price catalog. The public project repository is linked on the title page.

Recomputation uses the archived final verdicts without rerunning the Agent or introducing a new retrospective admission filter. Identity checks, nonempty binary verdict vectors, and agreement with the 50/50 score establish the inputs to aggregation.

Section~\ref{sec:results} reports the complete selected population, while Appendices~\ref{sec:appendix_A}--\ref{sec:appendix_D} document the task, adapter, evidence, and evaluation contracts. Task-level heatmaps and paired trajectory studies are provided in Appendices~\ref{sec:results-heatmaps} and~\ref{sec:appendix_I}.
\section{Results}
\label{sec:results}

The full matrix separates three quantities that an outcome-only
evaluation would combine: Agent task success, Harness-criterion satisfaction,
and the activation of candidate links between failed criteria.
All comparisons below use the same 95 tasks, nine model groups, and four
Harnesses. Scores are taken from final archived verdicts under the inclusion
policy in Section~\ref{sec:exp-admission}.

\subsection{Evaluation Population and Score Coverage}
\label{sec:results-population}

All 3,420 targeted cells have a final combined score, with nonempty binary
verdict vectors for both layers. Each model--Harness block contains 95 runs; each Harness therefore
contributes 855 and each model 380. Across the population, 14,979 of 20,016
Agent items pass (74.8\%), compared with 19,242 of 28,188 Harness items
(68.3\%). The mean per-run scores are 72.9\% for the Agent layer and
67.9\% for the Harness layer, producing a mean 50/50 score of 70.4. These continuous scores differ substantially from all-criteria success.
A total of 1,351 runs (39.5\%) satisfy every Agent criterion, whereas only
286 (8.4\%) satisfy every criterion in both layers.

\subsection{Overall Layered Outcomes}
\label{sec:results-overall}

Table~\ref{tab:main-summary} gives the primary harness comparison over all model--task cells per harness. Table~\ref{tab:results-main} reports the complete numerical comparison, including all nine fixed-model blocks. Claude Code has the highest Agent
task-success rate (44.0\%) and Agent macro score (75.9\%), as well as the
highest combined score (73.5). OpenClaw has the highest Harness macro rate
(72.0\%), Harness micro rate (71.9\%), and jointly clean rate (11.7\%).
Thus the Harness with the highest observed task success is not the one with
the highest measured Harness-criterion satisfaction.

\begin{table}[t]
\centering
\caption{\textbf{Aggregate results (\%).} All harnesses use 855 matched model--task cells. Overall averages unrounded Agent Score and Harness Macro equally, matching the first-page plots. Agent Pass requires all Agent criteria to pass, while Clean requires
all criteria in both layers to pass. Score and Macro average per-run
fractions, whereas Micro pools Harness items. Full model-level results appear in Table~\ref{tab:results-main}.}
\label{tab:main-summary}
\begin{tabular}{lrrrrrr}
\toprule
& \multicolumn{2}{c}{Agent} & \multicolumn{2}{c}{Harness} & & \\
\cmidrule(lr){2-3}\cmidrule(lr){4-5}
Harness & Pass & Score & Micro & Macro & Clean & Overall \\
\midrule
Claude Code & \textbf{44.0} & \textbf{75.9} & 71.5 & 71.2 & 9.1 & \textbf{73.5} \\
Codex & 35.8 & 70.7 & 60.2 & 59.8 & 2.8 & 65.2 \\
DeepSeek Harness & 40.0 & 73.3 & 69.5 & 68.8 & 9.8 & 71.0 \\
OpenClaw & 38.2 & 71.7 & \textbf{71.9} & \textbf{72.0} & \textbf{11.7} & 71.8 \\
\midrule
All harnesses & 39.5 & 72.9 & 68.3 & 67.9 & 8.4 & 70.4 \\
\bottomrule
\end{tabular}
\end{table}

The Harness macro range is 12.2 percentage points, from Codex at 59.8\% to
OpenClaw at 72.0\%. The Agent task-success range is smaller: 8.2 points,
from Codex at 35.8\% to Claude Code at 44.0\%.
Claude Code and OpenClaw are close on Harness micro performance, differing
by 0.4 points, despite their different Agent success rates. DeepSeek Harness
lies between these systems and Codex on both the mean Harness rate (68.8\%)
and Agent task success (40.0\%).

\subsection{Harness--Model Interaction}
\label{sec:results-interaction}

\begingroup
\scriptsize
\setlength{\tabcolsep}{2.5pt}
\renewcommand{\arraystretch}{1.10}
\setlength{\LTcapwidth}{\textwidth}
\begin{xltabular}{\linewidth}{@{}>{\raggedright\arraybackslash}X*{12}{r}@{}}
\caption{\textbf{Model $\times$ Harness main results.} Each fixed-model block evaluates 95 tasks per Harness. The aggregate pools nine models. Outcome columns are percentages: Pass requires all Agent criteria, Agent Score and Harness Macro average run-level fractions, Micro pools criteria. Clean requires both layers to pass, Link records candidate links, Abs.\ denotes Harness zeros without Agent zeros, Layered is the 50/50 mean. Resource columns average valid recorded usage: input, output, and total Tokens in thousands (k), and estimated Agent cost in USD at the frozen model-specific prices. Judge costs are excluded. Within each block, \textbf{bold} and \underline{underlining} mark the highest and second-highest displayed outcomes, with ties sharing marks. Link, Abs., resources, and the All Harnesses row are unranked.}\label{tab:results-main} \\
\toprule
& \multicolumn{2}{c}{\textbf{Agent layer}} & \multicolumn{2}{c}{\textbf{Harness layer}} & \multicolumn{3}{c}{\textbf{Attribution evidence}} & \textbf{Layered} & \multicolumn{4}{c}{\textbf{Agent resources (mean/run)}} \\
\cmidrule(lr){2-3}\cmidrule(lr){4-5}\cmidrule(lr){6-8}\cmidrule(lr){9-9}\cmidrule(l){10-13}
\textbf{Harness} & \textbf{Pass} & \textbf{Score} & \textbf{Micro} & \textbf{Macro} & \textbf{Clean} & \textbf{Link} & \textbf{Abs.} & \textbf{Score} & \shortstack{\textbf{Input}\\\textbf{(k)}} & \shortstack{\textbf{Output}\\\textbf{(k)}} & \shortstack{\textbf{Total}\\\textbf{(k)}} & \shortstack{\textbf{Est.\ cost}\\\textbf{(\$)}} \\
\midrule
\endfirsthead
\multicolumn{13}{c}{\tablename\ \thetable\ (continued)} \\
\toprule
& \multicolumn{2}{c}{\textbf{Agent layer}} & \multicolumn{2}{c}{\textbf{Harness layer}} & \multicolumn{3}{c}{\textbf{Attribution evidence}} & \textbf{Layered} & \multicolumn{4}{c}{\textbf{Agent resources (mean/run)}} \\
\cmidrule(lr){2-3}\cmidrule(lr){4-5}\cmidrule(lr){6-8}\cmidrule(lr){9-9}\cmidrule(l){10-13}
\textbf{Harness} & \textbf{Pass} & \textbf{Score} & \textbf{Micro} & \textbf{Macro} & \textbf{Clean} & \textbf{Link} & \textbf{Abs.} & \textbf{Score} & \shortstack{\textbf{Input}\\\textbf{(k)}} & \shortstack{\textbf{Output}\\\textbf{(k)}} & \shortstack{\textbf{Total}\\\textbf{(k)}} & \shortstack{\textbf{Est.\ cost}\\\textbf{(\$)}} \\
\midrule
\endhead
\midrule
\multicolumn{13}{r}{Continued on next page} \\
\endfoot
\bottomrule
\endlastfoot
\rowcolor{gray!14}
\multicolumn{13}{l}{\textbf{All models} \enspace (9 models; 855 runs per Harness)} \\*
Claude Code & \textbf{44.0} & \textbf{75.9} & \underline{71.5} & \underline{71.2} & 9.1 & 22.7 & 34.9 & \textbf{73.5} & 569.4 & 9.5 & 578.8 & 0.436 \\*
Codex & 35.8 & 70.7 & 60.2 & 59.8 & 2.8 & 28.1 & 33.0 & 65.2 & 606.9 & 10.2 & 617.2 & 0.595 \\*
DeepSeek Harness & \underline{40.0} & \underline{73.3} & 69.5 & 68.8 & \underline{9.8} & 21.8 & 30.2 & 71.0 & 361.5 & 11.2 & 372.7 & 0.446 \\*
OpenClaw & 38.2 & 71.7 & \textbf{71.9} & \textbf{72.0} & \textbf{11.7} & 26.0 & 26.5 & \underline{71.8} & 670.8 & 9.2 & 680.0 & 0.633 \\
\addlinespace[1pt]
\textbf{All Harnesses} & 39.5 & 72.9 & 68.3 & 67.9 & 8.4 & 24.6 & 31.1 & 70.4 & 552.1 & 10.0 & 562.1 & 0.528 \\
\midrule
\rowcolor{blue!7}
\multicolumn{13}{l}{\textbf{DeepSeek-V4-Pro} \enspace (95 tasks per Harness)} \\*
Claude Code & \textbf{43.2} & \textbf{72.5} & \underline{71.6} & \underline{71.2} & 8.4 & 22.1 & 34.7 & \underline{71.9} & 603.5 & 12.9 & 616.5 & 0.115 \\*
Codex & 30.5 & 68.4 & 59.9 & 59.3 & 0.0 & 27.4 & 30.5 & 63.8 & 580.1 & 13.0 & 593.1 & 0.046 \\*
DeepSeek Harness & 37.9 & 69.9 & 68.6 & 67.9 & \underline{10.5} & 25.3 & 27.4 & 68.9 & 325.0 & 11.4 & 336.4 & 0.043 \\*
OpenClaw & \underline{38.9} & \underline{72.0} & \textbf{71.9} & \textbf{71.9} & \textbf{11.6} & 25.3 & 27.4 & \textbf{72.0} & 605.0 & 10.8 & 615.8 & 0.094 \\
\midrule
\rowcolor{blue!7}
\multicolumn{13}{l}{\textbf{Gemini 3.7 Flash} \enspace (95 tasks per Harness)} \\*
Claude Code & \underline{40.0} & 74.1 & 70.4 & 69.7 & \underline{8.4} & 21.1 & 31.6 & 71.9 & 590.7 & 10.3 & 601.0 & 0.247 \\*
Codex & 38.9 & 73.8 & 62.5 & 62.0 & 7.4 & 23.2 & 31.6 & 67.9 & 622.2 & 13.1 & 635.3 & 0.226 \\*
DeepSeek Harness & \textbf{44.2} & \textbf{74.6} & \underline{70.8} & \underline{70.0} & \underline{8.4} & 25.3 & 35.8 & \underline{72.3} & 496.0 & 11.6 & 507.6 & 0.416 \\*
OpenClaw & 38.9 & \underline{74.2} & \textbf{75.1} & \textbf{74.8} & \textbf{10.5} & 26.3 & 28.4 & \textbf{74.5} & 975.9 & 10.6 & 986.5 & 0.772 \\
\midrule
\rowcolor{blue!7}
\multicolumn{13}{l}{\textbf{GLM-5.3} \enspace (95 tasks per Harness)} \\*
Claude Code & \underline{43.2} & \textbf{76.0} & \textbf{73.7} & \textbf{73.1} & \underline{9.5} & 21.1 & 33.7 & \textbf{74.6} & 639.3 & 13.9 & 653.2 & 0.408 \\*
Codex & 28.4 & 68.6 & 59.1 & 59.3 & 2.1 & 28.4 & 26.3 & 63.9 & 627.1 & 8.2 & 635.3 & 0.189 \\*
DeepSeek Harness & \textbf{44.2} & \underline{75.5} & \underline{71.4} & \underline{70.5} & \underline{9.5} & 17.9 & 34.7 & \underline{73.0} & 380.6 & 14.5 & 395.1 & 0.226 \\*
OpenClaw & 41.1 & 67.5 & 69.7 & 70.1 & \textbf{12.6} & 25.3 & 28.4 & 68.8 & 606.1 & 10.2 & 616.3 & 0.180 \\
\midrule
\rowcolor{blue!7}
\multicolumn{13}{l}{\textbf{GPT-5.6-Sol} \enspace (95 tasks per Harness)} \\*
Claude Code & \textbf{44.2} & \textbf{79.6} & \textbf{72.8} & \textbf{72.6} & \underline{11.6} & 21.1 & 32.6 & \textbf{76.1} & 559.5 & 4.7 & 564.2 & 0.814 \\*
Codex & 41.1 & 71.8 & 57.6 & 56.4 & 3.2 & 25.3 & 37.9 & 64.1 & 1427.8 & 10.9 & 1438.7 & 1.649 \\*
DeepSeek Harness & 36.8 & \underline{73.2} & \underline{71.3} & 70.4 & \underline{11.6} & 22.1 & 25.3 & \underline{71.8} & 298.3 & 2.6 & 300.9 & 0.361 \\*
OpenClaw & \underline{42.1} & 71.5 & 71.1 & \underline{71.2} & \textbf{14.7} & 29.5 & 27.4 & 71.4 & 606.7 & 5.1 & 611.8 & 0.974 \\
\midrule
\rowcolor{blue!7}
\multicolumn{13}{l}{\textbf{Grok-4.6} \enspace (95 tasks per Harness)} \\*
Claude Code & \textbf{55.8} & \textbf{81.6} & \textbf{71.9} & \textbf{71.9} & \textbf{10.5} & 18.9 & 45.3 & \textbf{76.7} & 531.1 & 7.2 & 538.3 & 0.509 \\*
Codex & \underline{44.2} & 72.9 & 59.6 & 59.5 & 5.3 & 21.1 & 38.9 & 66.2 & 292.6 & 4.2 & 296.9 & 0.462 \\*
DeepSeek Harness & 43.2 & \underline{75.6} & \underline{71.4} & \underline{70.6} & \textbf{10.5} & 18.9 & 32.6 & \underline{73.1} & 315.7 & 11.9 & 327.6 & 0.378 \\*
OpenClaw & 33.7 & 70.8 & 68.5 & 69.0 & \underline{8.4} & 24.2 & 25.3 & 69.9 & 541.6 & 4.4 & 546.0 & 0.822 \\
\midrule
\rowcolor{blue!7}
\multicolumn{13}{l}{\textbf{HY3} \enspace (95 tasks per Harness)} \\*
Claude Code & \textbf{32.6} & \underline{69.5} & \underline{69.7} & \underline{69.2} & \underline{6.3} & 30.5 & 26.3 & \underline{69.3} & 798.9 & 15.0 & 813.9 & 0.044 \\*
Codex & \underline{30.5} & 64.2 & 59.3 & 59.0 & 1.1 & 35.8 & 29.5 & 61.6 & 415.9 & 13.7 & 429.6 & 0.026 \\*
DeepSeek Harness & 28.4 & 64.8 & 65.8 & 65.7 & \underline{6.3} & 30.5 & 22.1 & 65.2 & 247.0 & 11.2 & 258.2 & 0.018 \\*
OpenClaw & 29.5 & \textbf{70.0} & \textbf{72.7} & \textbf{72.7} & \textbf{11.6} & 29.5 & 17.9 & \textbf{71.4} & 659.5 & 18.5 & 678.0 & 0.041 \\
\midrule
\rowcolor{blue!7}
\multicolumn{13}{l}{\textbf{Kimi K3} \enspace (95 tasks per Harness)} \\*
Claude Code & \textbf{49.5} & \textbf{78.2} & 71.1 & \underline{70.6} & \textbf{11.6} & 21.1 & 37.9 & \textbf{74.4} & 358.1 & 4.8 & 362.9 & 0.613 \\*
Codex & 42.1 & 74.0 & 61.3 & 60.9 & 2.1 & 28.4 & 40.0 & 67.4 & 573.9 & 7.7 & 581.6 & 0.379 \\*
DeepSeek Harness & \underline{43.2} & \underline{74.4} & \underline{71.5} & 70.5 & \underline{10.5} & 16.8 & 32.6 & 72.4 & 281.7 & 6.1 & 287.8 & 0.225 \\*
OpenClaw & 40.0 & 73.8 & \textbf{73.6} & \textbf{73.5} & \underline{10.5} & 22.1 & 29.5 & \underline{73.7} & 580.8 & 5.7 & 586.6 & 0.351 \\
\midrule
\rowcolor{blue!7}
\multicolumn{13}{l}{\textbf{Opus 5} \enspace (95 tasks per Harness)} \\*
Claude Code & \textbf{53.7} & \textbf{83.4} & 71.8 & \underline{71.9} & 7.4 & 16.8 & 46.3 & \textbf{77.6} & 379.4 & 3.7 & 383.1 & 0.660 \\*
Codex & 30.5 & 70.3 & 60.8 & 60.5 & 2.1 & 34.7 & 28.4 & 65.4 & 482.8 & 7.6 & 490.4 & 2.121 \\*
DeepSeek Harness & \underline{49.5} & \underline{79.3} & \underline{72.3} & 71.4 & \textbf{13.7} & 17.9 & 35.8 & \underline{75.4} & 551.5 & 16.9 & 568.5 & 2.071 \\*
OpenClaw & 41.1 & 76.1 & \textbf{73.8} & \textbf{73.9} & \underline{11.6} & 23.2 & 29.5 & 75.0 & 822.3 & 7.7 & 830.0 & 2.168 \\
\midrule
\rowcolor{blue!7}
\multicolumn{13}{l}{\textbf{Qwen3.8-Max} \enspace (95 tasks per Harness)} \\*
Claude Code & 33.7 & 68.3 & \underline{70.5} & \underline{70.2} & \underline{8.4} & 31.6 & 25.3 & \underline{69.3} & 663.8 & 12.6 & 676.4 & 0.517 \\*
Codex & \underline{35.8} & \textbf{72.1} & 61.4 & 61.0 & 2.1 & 28.4 & 33.7 & 66.5 & 436.7 & 13.4 & 450.2 & 0.258 \\*
DeepSeek Harness & 32.6 & \underline{71.8} & 62.5 & 62.6 & 7.4 & 21.1 & 25.3 & 67.2 & 357.9 & 14.2 & 372.1 & 0.279 \\*
OpenClaw & \textbf{38.9} & 69.0 & \textbf{70.6} & \textbf{70.9} & \textbf{13.7} & 28.4 & 25.3 & \textbf{70.0} & 638.8 & 9.7 & 648.5 & 0.290 \\
\end{xltabular}
\endgroup

The fixed-model blocks of Table~\ref{tab:results-main} compare all four
Harnesses on the same 95 tasks. Codex has the lowest Harness macro rate in
every model group. OpenClaw has the highest rate for six models, while
Claude Code leads for GLM-5.3, GPT-5.6-Sol, and Grok-4.6.
The within-model Harness range spans 10.0--16.2 percentage points:
Qwen3.8-Max has the smallest spread and GPT-5.6-Sol the largest.

Agent success exhibits a different interaction. Grok-4.6 ranges from
33.7\% with OpenClaw to 55.8\% with Claude Code, a 22.1-point difference.
Opus 5 has an even wider range, from 30.5\% with Codex to 53.7\% with
Claude Code, a 23.2-point difference.
For Gemini 3.7 Flash, DeepSeek Harness has the highest Agent success
(44.2\%), while OpenClaw has the highest Harness macro rate (74.8\%).
For Qwen3.8-Max, OpenClaw leads on both measures. These fixed-population
contrasts show that a model's observed task success depends on its Harness
configuration, and that the strongest task outcome need not coincide with
the strongest execution-layer score.

\subsection{Reliability by Taxonomy Dimension}
\label{sec:results-taxonomy}

Table~\ref{tab:results-dimensions} decomposes the full population by the
task's primary Harness dimension. Each column uses the same tasks and nine
models within a dimension. Execution has the lowest pooled Harness micro
rate (64.6\%), followed by Lifecycle (65.5\%) and Context (65.6\%).
Observability reaches 84.3\%. Verification reaches 90.1\%.

\begin{table}[t]
\centering
\caption{\textbf{Harness Micro by reliability dimension (\%).} All four harnesses use the same tasks and nine models within each row.}
\label{tab:main-dimensions}
\begin{tabular}{lrrrrr}
\toprule
Dimension & Tasks & \shortstack{Claude\\Code} & Codex & \shortstack{DeepSeek\\Harness} & OpenClaw \\
\midrule
Context & 15 & \textbf{74.5} & 62.0 & 64.3 & 61.6 \\
Execution & 27 & 64.7 & 51.5 & 65.3 & \textbf{77.1} \\
Governance & 9 & 72.3 & 74.2 & 68.8 & \textbf{80.5} \\
Lifecycle & 19 & 71.3 & 53.0 & \textbf{71.7} & 65.9 \\
Observability & 7 & \textbf{88.7} & 76.5 & 86.8 & 85.0 \\
Tools & 17 & \textbf{71.5} & 65.6 & \textbf{71.5} & 70.2 \\
Verification & 1 & 93.7 & \textbf{95.2} & 82.5 & 88.9 \\
\bottomrule
\end{tabular}
\end{table}

\begin{table}[t]
\centering
\caption{\textbf{Reliability by taxonomy dimension.} Criterion pass rates and Agent task success on the full 3,420-run grid (\%, higher is better).}
\label{tab:results-dimensions}
\begingroup
\footnotesize
\setlength{\tabcolsep}{3.5pt}
\renewcommand{\arraystretch}{1.20}
\begin{tabular*}{\linewidth}{@{\extracolsep{\fill}}lrrrrrrrr@{}}
\toprule
& \multicolumn{2}{c}{\textbf{Coverage}} & \multicolumn{4}{c}{\textbf{Harness Micro (\%)}} & \multicolumn{2}{c}{\textbf{Pooled (\%)}} \\
\cmidrule(lr){2-3}\cmidrule(lr){4-7}\cmidrule(l){8-9}
\textbf{Dimension} & \textbf{Tasks} & \textbf{Runs/H} & \multicolumn{1}{c}{\shortstack{\textbf{Claude}\\\textbf{Code}}} & \multicolumn{1}{c}{\textbf{Codex}} & \multicolumn{1}{c}{\shortstack{\textbf{DeepSeek}\\\textbf{Harness}}} & \multicolumn{1}{c}{\textbf{OpenClaw}} & \shortstack{\textbf{H}\\\textbf{Micro}} & \shortstack{\textbf{Agent}\\\textbf{Pass}} \\
\midrule
Context & 15 & 135 & \cellcolor{blue!12}\textbf{74.5} & 62.0 & \underline{64.3} & 61.6 & 65.6 & 45.2 \\
\rowcolor{gray!4}
Execution & 27 & 243 & 64.7 & 51.5 & \underline{65.3} & \cellcolor{blue!12}\textbf{77.1} & 64.6 & 49.2 \\
Governance & 9 & 81 & 72.3 & \underline{74.2} & 68.8 & \cellcolor{blue!12}\textbf{80.5} & 74.0 & 27.2 \\
\rowcolor{gray!4}
Lifecycle & 19 & 171 & \underline{71.3} & 53.0 & \cellcolor{blue!12}\textbf{71.7} & 65.9 & 65.5 & 43.6 \\
Observability & 7 & 63 & \cellcolor{blue!12}\textbf{88.7} & 76.5 & \underline{86.8} & 85.0 & 84.3 & 43.3 \\
\rowcolor{gray!4}
Tools & 17 & 153 & \cellcolor{blue!12}\textbf{71.5} & 65.6 & \cellcolor{blue!12}\textbf{71.5} & \underline{70.2} & 69.7 & 21.4 \\
Verification & 1 & 9 & \underline{93.7} & \cellcolor{blue!12}\textbf{95.2} & 82.5 & 88.9 & 90.1 & 8.3 \\
\midrule
\rowcolor{blue!7}
\textbf{All tasks} & 95 & 855 & \underline{71.5} & 60.2 & 69.5 & \cellcolor{blue!12}\textbf{71.9} & 68.3 & 39.5 \\
\bottomrule
\end{tabular*}
\par\smallskip
\begin{minipage}{\linewidth}
\footnotesize
\textit{Note.} Runs/H counts runs per Harness: nine models per task. Within each row, \textbf{bold} with shading marks the highest Harness Micro rate and \underline{underlining} the second-highest; ties at the displayed precision share marks. Pooled H Micro aggregates criterion counts across Harnesses; Agent Pass requires all Agent criteria. Pooled columns are not ranked. All tasks pools the full grid, not the seven displayed dimension rates. Each task belongs to one dimension.
\end{minipage}
\endgroup
\end{table}

Execution also exhibits a large cross-harness spread: Codex records 51.5\% and OpenClaw 77.1\%. Context reverses part of the overall ordering: Claude Code
reaches 74.5\%, whereas OpenClaw records 61.6\%.
Lifecycle similarly separates DeepSeek Harness (71.7\%) and Claude Code
(71.3\%) from Codex (53.0\%). These results suggest that Harness selection should reflect the
reliability requirements of the target task.

\begin{figure}[t]
\centering
\includegraphics[width=\linewidth]{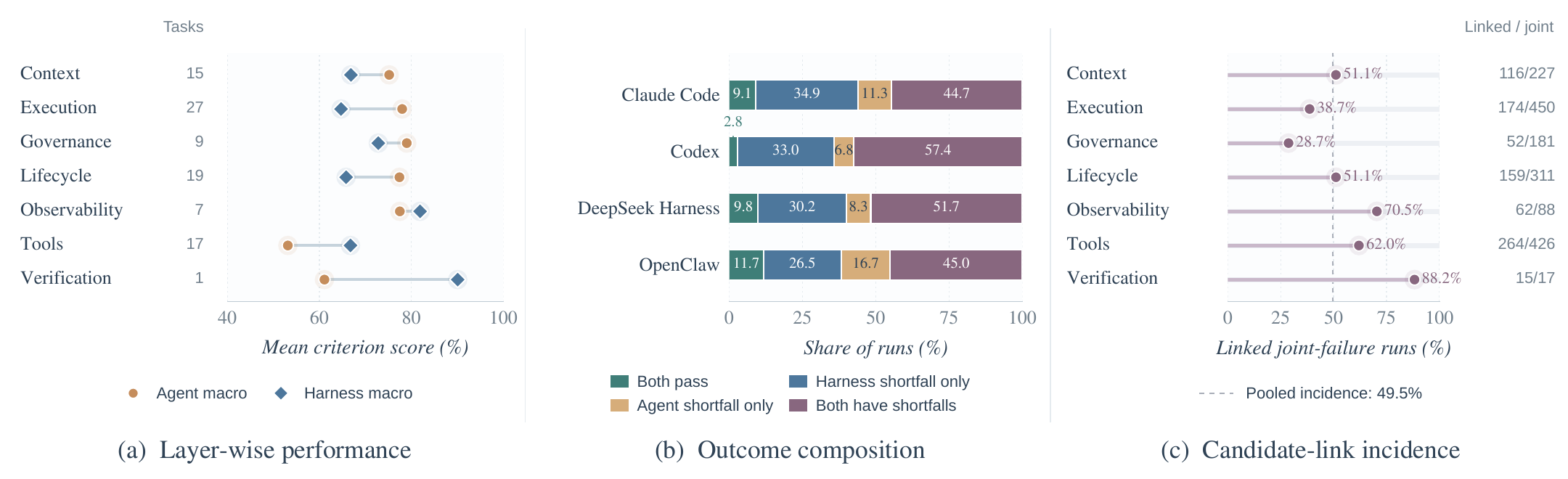}
\caption{\textbf{Reliability and candidate links on the 3,420-run grid.} (a) Mean per-run layer scores by dimension. (b) Four criterion-verdict states over 855 runs per harness. (c) Runs with a candidate link among joint-failure runs, with linked/joint-failure counts and the pooled 49.5\% rate.}
\label{fig:results-reliability-analysis}
\end{figure}

Figure~\ref{fig:results-reliability-analysis}(a) compares both layers using
the same per-run macro aggregation. Execution records an Agent score of
78.0\% versus a Harness score of 64.7\%. Lifecycle similarly records
77.4\% versus 65.8\%. Tools reverses this gap (53.2\% versus 66.8\%). The layer gap therefore
changes direction across dimensions rather than following a uniform offset.

All-criteria Agent success is a stricter outcome measure. Execution has the
highest Agent success among the multi-task dimensions (49.2\%) despite its lowest
pooled Harness micro rate. Tools has a 69.7\% Harness micro rate but only
21.4\% Agent success over 612 runs. These contrasts
show why a criterion-level execution measure cannot substitute for an
all-criteria task-outcome measure.

Among the 40 properties, none reaches a 100\% Harness micro
rate across all nine models and four Harnesses.
Network restriction is lowest at 38.9\%. Tool exposure reaches 45.8\% Harness satisfaction
and no all-criteria Agent successes across its two scenarios.
Subagent result aggregation reaches only 46.2\% Harness
satisfaction while achieving 72.2\% Agent success over 144 runs.
Cross-turn fact retention shows an even sharper contrast:
48.4\% Harness satisfaction and 95.8\% Agent success. These contrasts demonstrate the diagnostic value of TraceTrial-Bench:
its two-layer evaluation reveals mechanism-level shortfalls even when
Agent task-success rates are high.

\subsection{Failure States and Candidate Attribution}
\label{sec:results-attribution}

Table~\ref{tab:results-failure-analysis} reports failure states and
candidate-link activation across Harnesses and primary dimensions.
We partition runs using all raw zero-scored items,
$F_H=\operatorname{fail}(V_H)$ and
$Z_A=\operatorname{fail}(V_A)$, and report linked runs separately from
the four mutually exclusive states. The raw criterion intersections contain 286 jointly clean runs, 1,065 Harness-only shortfalls, 369 Agent-only shortfalls, and 1,700 joint shortfalls. Thus 31.1\% of all runs achieve Agent success despite at least one failed Harness criterion. Among the 1,700 joint-shortfall runs, 842 (49.5\%) contain at least one stored candidate link, totaling 1,633 activated endpoint pairs. Of 2,069 Agent-failure runs, 1,773 (85.7\%) retain at least one failed Agent criterion not reached by a link.

\begin{table}[t]
\centering
\caption{\textbf{Failure states and candidate-link activation.} Results across Harnesses and primary dimensions on the full 3,420-run grid.}
\label{tab:results-failure-analysis}
\begingroup
\footnotesize
\setlength{\tabcolsep}{3pt}
\renewcommand{\arraystretch}{1.15}
\begin{tabular*}{\linewidth}{@{\extracolsep{\fill}}l*{9}{r}@{}}
\toprule
& & \multicolumn{2}{c}{\textbf{Outcomes}} & \multicolumn{4}{c}{\textbf{Criterion states (runs)}} & \multicolumn{2}{c}{\textbf{Candidate links}} \\
\cmidrule(lr){3-4}\cmidrule(lr){5-8}\cmidrule(l){9-10}
\textbf{Group / Harness} & \textbf{Runs} & \shortstack{\textbf{Agent}\\\textbf{Pass}} & \shortstack{\textbf{H Micro}\\\textbf{(\%)}} & \textbf{Clean} & \textbf{H-only} & \textbf{A-only} & \textbf{Both} & \textbf{Linked} & \shortstack{\textbf{Link/Both}\\\textbf{(\%)}} \\
\midrule
\rowcolor{blue!7}
\multicolumn{10}{l}{\textbf{Full grid}} \\
\textbf{All Harnesses} & 3420 & 1351 & 68.3 & 286 & 1065 & 369 & 1700 & 842 & 49.5 \\
\rowcolor{gray!4}
Claude Code & 855 & 376 & 71.5 & 78 & 298 & 97 & 382 & 194 & 50.8 \\
Codex & 855 & 306 & 60.2 & 24 & 282 & 58 & 491 & 240 & 48.9 \\
\rowcolor{gray!4}
DeepSeek Harness & 855 & 342 & 69.5 & 84 & 258 & 71 & 442 & 186 & 42.1 \\
OpenClaw & 855 & 327 & 71.9 & 100 & 227 & 143 & 385 & 222 & 57.7 \\
\midrule
\rowcolor{blue!7}
\multicolumn{10}{l}{\textbf{Primary dimension}} \\
Context & 540 & 244 & 65.6 & 29 & 215 & 69 & 227 & 116 & 51.1 \\
\rowcolor{gray!4}
Execution & 972 & 478 & 64.6 & 76 & 402 & 44 & 450 & 174 & 38.7 \\
Governance & 324 & 88 & 74.0 & 14 & 74 & 55 & 181 & 52 & 28.7 \\
\rowcolor{gray!4}
Lifecycle & 684 & 298 & 65.5 & 72 & 226 & 75 & 311 & 159 & 51.1 \\
Observability & 252 & 109 & 84.3 & 50 & 59 & 55 & 88 & 62 & 70.5 \\
\rowcolor{gray!4}
Tools & 612 & 131 & 69.7 & 42 & 89 & 55 & 426 & 264 & 62.0 \\
Verification & 36 & 3 & 90.1 & 3 & 0 & 16 & 17 & 15 & 88.2 \\
\bottomrule
\end{tabular*}
\par\smallskip
\begin{minipage}{\linewidth}
\footnotesize
\end{minipage}
\endgroup
\end{table}

Joint failure is the largest state, but Harness and Agent verdicts
frequently disagree. Claude Code has the most H-only runs, whereas
OpenClaw has the most A-only runs and the most jointly clean runs
(Figure~\ref{fig:results-reliability-analysis}(b)).
The H-only state represents \emph{operational absorption}: measured
Harness shortfalls coexist with a fully passing Agent outcome.
H-only runs therefore provide a focused set of cases for diagnosing
Harness-level shortfalls in tasks the Agent successfully completes.

Co-occurrence is broader than candidate-link activation. Fewer than half
of jointly failing runs contain a stored link, and most Agent-failure
runs retain at least one failed criterion not reached by any link.
Conditional activation is lower in Execution and Governance than in
Tools and Observability (Figure~\ref{fig:results-reliability-analysis}(c)).
The table distinguishes this conditional rate from incidence over all
runs: a Harness can have more linked runs because it has more joint
failures, without having a higher conditional activation rate.

\subsection{Recorded Resource Use}
\label{sec:results-resources}

At the frozen model-specific prices, mean estimated Agent costs per selected run are \$0.436 for Claude Code, \$0.595 for Codex, \$0.446 for DeepSeek Harness, and \$0.633 for OpenClaw. Mean total recorded tokens are 578.8k, 617.2k, 372.7k, and 680.0k, respectively. Table~\ref{tab:results-main} retains the corresponding input, output, total-token, and cost columns for each model group.

\subsection{Findings and Interpretation Limits}
\label{sec:results-summary}

The full-grid results support three conclusions. First, outcome quality
and Harness-criterion satisfaction are empirically distinct: Agent
success reaches 39.5\%, but only 8.4\% of runs pass every item in both
layers, and different Harnesses lead the two aggregate measures.
Second, differences depend on both the model and taxonomy dimension:
within-model Harness macro ranges reach 16.2 points, and Harness micro
orderings vary across dimensions.
Third, failure co-occurrence exceeds candidate-link activation:
842 of 1,700 both-zero runs contain a stored link, while 1,065 runs
retain fully passing Agent outcomes despite Harness zeros.

Appendix~\ref{sec:results-heatmaps} shows the task-level score grid, and Appendix~\ref{sec:appendix_I} provides five paired trajectory studies. These selected contrasts support inspection of specific behaviors rather than a population-level estimate of attribution accuracy.

\section{Discussion}
\label{sec:discussion}

\subsection{Limitations}

The evaluation uses a targeted task suite with uneven mechanism
coverage and one selected execution per task-model-harness
combination. The results characterize the collected configurations,
with limited evidence about generalization to deployment workloads
or stability across repeated runs. Variation in task revisions,
runtime settings, and scoring budgets further constrains causal
conclusions about individual harness design choices.

\subsection{Future Directions}

\paragraph{Harness Evolution.}
TraceTrial-Bench could support a trace-guided improvement loop:
identify mechanism failures, revise the relevant Harness behavior,
and evaluate the changes on held-out tasks under fixed models and
budgets. Controlled interventions and repeated executions would
help assess whether each modification resolves the targeted failure
and introduces regressions elsewhere. Tracking versions,
configurations, and supporting evidence would connect Harness
changes to measurable improvements in execution reliability
and task performance.

\paragraph{Evaluation for Recursive Self-Improvement.}
Future evaluation methods could combine task outcomes, Harness
behavior, and evidence quality to provide actionable feedback for
Recursive Self-Improvement (RSI). Crossed evaluations of Agent and Harness
versions would help distinguish their individual contributions
and interactions. Criterion-level feedback could then guide
updates to Agent planning, tool use, and recovery strategies.
Testing these updates on held-out tasks and unfamiliar Harnesses
would assess whether the improvements transfer beyond the
configurations used during adaptation.
\section{Related Works}
\label{sec:relatedworks}

\subsection{Agent Harnesses}

Harness research addresses context, tools, state, verification, and permissions~\citep{zhong2026ai,barbaste2026harness}. ACE, ACDL, and SkillAlign improve context and interfaces~\citep{zhang2026agentic,peleg2026language,ren2026skillalign}; LongHorizon-Harness and Crab address continuity and recovery~\citep{ma2026longhorizon,wu2026crab}; AgentBound and SafeHarness enforce policies and safeguards~\citep{buhler2026agentbound,lin2026safeharness}. HARBOR, retrospective optimization, and SHAPER optimize configurations, skills, or harness code~\citep{sengupta2026harbor,pan2026evolving,wang2026self}, while HarnessFix diagnoses and patches implementation flaws~\citep{chen2026failed}. Interpreting improvements requires matched budgets and held-out evaluation~\citep{wang2026rethinking}. TraceTrial-Bench supplies mechanism-level criteria and evidence while preserving native harness behavior. Configuration search, skill evolution, and implementation repair target different system components. TraceTrial-Bench preserves native execution strategies and records obligation-specific evidence to examine whether task-level gains coincide with more reliable harness behavior.

\subsection{Agent Evaluation and Failure Analysis}

Terminal-Bench and SWE-Bench Pro assess verifiable task
outcomes~\citep{merrill2026terminal,deng2025swe}.
Harness-Bench compares model--harness configurations under shared
conditions~\citep{yao2026harness}, while HarnessAudit and HarnessRisk
examine execution fidelity and adversarial
risks~\citep{liu2026auditing,bai2026harnessrisk}. For failure analysis, Who\&When localizes responsible agents and
error steps~\citep{zhang2025agent}.
AgentRx diagnoses constraint violations~\citep{barke2026agentrx},
while REFLECT uses interventions and controlled
replay~\citep{lin2026reflect}. TraceTrial-Bench complements these approaches by combining
separate Agent and Harness criteria with evidence-based failure
analysis. The two-layer evaluation exposes successful deliveries
that coexist with Harness shortfalls. Its attribution framework
connects failed criteria through declared mechanisms and
source-specific evidence.

\section{Conclusion}
\label{conclusion}
\label{sec:conclusion}

TraceTrial-Bench evaluates native agent harnesses through 95 mechanism-oriented tasks across seven reliability components. Trajectory tracing and multi-source evidence support separate Agent and Harness scores, while explicit mechanism chains guide auditable failure analysis. Across 3,420 selected runs, Harness Macro differs by 12.2 percentage points and the leaders in task success and harness satisfaction differ. In particular, 1,065 of 3,420 runs (31.1\%) achieve Agent success despite at least one failed Harness criterion. These successful deliveries with failed Harness criteria show that outcome-only evaluation cannot characterize harness reliability. The released tasks, configurations, traces, and verdicts support evidence-based comparison and diagnosis of harness behavior.

\section{Authors}

Zhiyong Cao\core, Hengjie Zhu\core, Yongxin Wang\core, Xuchen Li\core, Jintao Li, Shuxia Liu, Dunqiang Liu, Bo Wang, Kairan Zhang, Cong Lin, JiangDi Yu, Siyuan Liu, Sijia Yao, Yi-Liang Wei, Lingxu Ma, Aijia Wang, Anxin Chen, Zixian Wei, Xiang Ruan, Ruizhi Xu, Ruqian Shi, Baihua Xiao\corresponding, Sicheng Zhou\corresponding

\medskip
\noindent
\core\ Core contributor.
\corresponding\ Corresponding author.

\label{main-text-end}

\clearpage
\bibliography{biblio}
\bibliographystyle{colm2024_conference}

\clearpage
\appendix
\section*{Appendix}
\label{sec:appendix}

\section{Benchmark Taxonomy and Task Suite}
\label{sec:appendix_A}

\subsection{Harness Reliability Taxonomy}
\label{subsec:A_1}

The taxonomy adapts the ETCLOVG decomposition~\citep{li2026agent} to observable harness responsibilities in the 95-task release. A property belongs in the benchmark when the harness mediates it, a task can exercise it under comparable conditions, and its outcome can be checked against execution evidence. The seven primary components are

\begin{equation}
\mathcal{D}_{H}
=
\{C, E, G, L, O, T, V\},
\end{equation}

Each task contributes to one primary component and one mechanism-level property. Secondary tags describe interactions without duplicating observations in component totals. Table~\ref{tab:harness-taxonomy-components} defines the responsibilities and their scope.

\begingroup
\small
\renewcommand{\arraystretch}{1.12}
\begin{xltabular}{\linewidth}{@{}P{0.13\textwidth} P{0.34\textwidth} Z@{}}
\caption{The seven components of the TraceTrial-Bench harness reliability taxonomy.}
\label{tab:harness-taxonomy-components} \\

\toprule
\textbf{Component} &
\textbf{Harness responsibility} &
\textbf{Representative properties in the final task suite} \\
\midrule
\endfirsthead

\toprule
\textbf{Component} &
\textbf{Harness responsibility} &
\textbf{Representative properties in the final task suite} \\
\midrule
\endhead

Context (C) &
Manage the information made available to the model and preserve task-relevant content across context reduction, turns, sessions, and agent boundaries. &
Truncation disclosure, rolling-compaction fidelity, cross-turn fact retention, session-resume fidelity, persistent-memory behavior, and subagent context isolation. \\

Execution (E) &
Execute requested actions and return process state, output, and failure information without changing or concealing their operational meaning. &
Exit-code fidelity, standard-error capture, timeout behavior, shell-state persistence, background-process supervision, timeout propagation, stale-artifact handling, and PTY termination. \\

Governance (G) &
Apply permissions, safety policies, and environmental restrictions consistently while preserving a clear relationship between declared and enforced behavior. &
Execution governance, approval and guardrail consistency, network restrictions, destructive-action control, and secret or credential redaction. \\

Lifecycle (L) &
Coordinate state and control flow across turns, sessions, delegated work, recovery operations, and resource boundaries. &
Model-call budgets, submitted-versus-confirmed state, subagent result aggregation, pending-child handling, workflow handoff, goal revision, and session recovery. \\

Observability (O) &
Expose sufficient and accurate information for the model or operator to identify failures, distinguish their sources, and reconstruct important execution events. &
Tool-error relay, harness-error and refusal notification, event provenance, diagnostic state, and auditable completion information. \\

Tools (T) &
Expose external capabilities through stable model-facing interfaces and faithfully transport tool definitions, arguments, results, and interaction requests. &
Tool availability, naming and registration, argument transport, structured-result fidelity, skill-result relay, dynamic tool discovery, and structured user-clarification channels. \\

Verification (V) &
Support evidence-based confirmation of execution progress and completion, and distinguish verified results from unsupported success claims. &
Claim-to-execution consistency, citable completion receipts, disclosure of incomplete work, and consistency between reported values and produced artifacts. \\

\bottomrule

\end{xltabular}
\endgroup

Ownership follows the tested responsibility: tool declarations and argument transport belong to Tools; process outcomes belong to Execution; child context belongs to Context; child completion belongs to Lifecycle. Verification concerns harness support for checking progress, while the independent oracle belongs to the measurement apparatus.

Component and property counts appear in Table~\ref{tab:taxonomy-coverage}.

\subsection{Task Design and Benchmark Populations}
\label{app:task-design-populations}

Each task couples a harness mechanism with a practical deliverable. Examples include oversized outputs, masked exit failures, pending jobs, session handoffs, and structured clarification. Private or one-time values make correct delivery depend on the intended interaction; tasks that test recovery provide a bounded recovery route.

The task package fixes the prompt or turn sequence, initial workspace, tools, budgets, required outputs, verifier, and two score sheets. Cross-harness adapters preserve this contract while retaining native execution behavior (Appendix~\ref{sec:appendix_B}). A missing target capability can be a measured outcome; an adapter that changes the task semantics creates a different comparison.

The release population is the 95 unique paths in \texttt{lean95.manifest}. The construction groups in Table~\ref{tab:task-source-groups} describe task origins and receive no separate scoring weights.

\begingroup
\begin{xltabular}{\linewidth}{@{}P{0.31\linewidth} r Z@{}}
\caption{Construction groups represented in the final 95-task release.}
\label{tab:task-source-groups} \\

\toprule
\textbf{Construction group} & \textbf{Tasks} & \textbf{Design purpose} \\
\midrule
\endfirsthead

\toprule
\textbf{Construction group} & \textbf{Tasks} & \textbf{Design purpose} \\
\midrule
\endhead

Obstruction and attribution cases
& 28
& Expose harness-created obstacles, available recovery paths, and the possible effect of those obstacles on task completion. \\

Shared-mechanism comparison cases
& 27
& Exercise mechanisms that are broadly available across harnesses so that comparisons focus on differences in implementation quality. \\

Capability-gap cases
& 21
& Examine situations in which harnesses differ in whether they provide a usable path for clarification, delegation, approval, context handling, or lifecycle control. \\

Facet-balancing supplement
& 19
& Improve the representation of under-covered mechanisms and execution surfaces, particularly coding-oriented and terminal-oriented scenarios. \\
\midrule
Total
& 95
& Authoritative release population defined by \texttt{lean95.manifest}. \\
\bottomrule

\end{xltabular}
\endgroup

The protocol distinguishes task construction from execution and evaluation:

\begin{equation}
\begin{aligned}
\text{Authored}
&\rightarrow \text{Audited}
\rightarrow \text{Cross-harness eligible} \\
&\rightarrow \text{Executed}
\rightarrow \text{Apparatus valid}
\rightarrow \text{Measured}
\rightarrow \text{Attributable}.
\end{aligned}
\end{equation}

\subsection{Task Cards, Oracles, and Hard Gates}
\label{app:task-cards-oracles-gates}

Task cards summarize the contract, execution conditions, and verification sources defined by the task files. All released packages provide a prompt, turn specification, runtime profile, workspace, executable verifier, and separate Harness and Agent score sheets. Table~\ref{tab:task-card-fields} gives the normalized documentation fields; the archived task package remains authoritative.

\begingroup
\begin{xltabular}{\linewidth}{@{}P{0.20\linewidth} P{0.29\linewidth} Z@{}}
\caption{Normalized task-card fields for the released benchmark tasks.}
\label{tab:task-card-fields} \\

\toprule
\textbf{Field} & \textbf{Primary source} & \textbf{Content} \\
\midrule
\endfirsthead

\toprule
\textbf{Field} & \textbf{Primary source} & \textbf{Content} \\
\midrule
\endhead

Task identity
& \texttt{task.yaml}, release manifest
& Task ID, title, released path, and task revision. \\

Scenario and target
& \texttt{task.yaml}
& Application domain, primary scenario, target component, and mechanism-level property. \\

Agent-visible contract
& \texttt{prompt.md}, \texttt{turns.json}
& User instructions, task goal, operational rules, working boundaries, and expected response. \\

Conversation protocol
& \texttt{turns.json}, \path{runtime/execution.json}
& Single-turn, fixed multi-turn, or simulator-driven interaction, including any session boundary or replay condition. \\

Initial state
& \path{workspace/}, simulator configuration
& Files, directories, services, records, and controlled state available when execution begins. \\

Available interfaces
& \texttt{task.yaml}, runtime profile, tool manifests
& Native tools, task-specific tools, business services, process controls, and other interfaces required by the task. \\

Target pressure
& \texttt{task.yaml}, \texttt{isolation.yaml}, simulator configuration
& The controlled condition used to exercise the target behavior, such as truncation, timeout, pending work, state recovery, schema conflict, or policy denial. \\

Milestones and restrictions
& Prompt, turn schedule, simulator, verifier
& Required intermediate states and actions that would bypass or invalidate the intended task path. \\

Deliverables
& \path{expected_outputs}, prompt, verifier
& Required files or responses, together with their paths, formats, schemas, and completeness conditions. \\

Resource budget
& \texttt{task.yaml}, resolved run configuration
& Maximum user-input turns, tool calls, and wall-clock time, together with any task-specific limit. \\

Execution contract
& \path{runtime/execution.json}, probes
& Execution mode, runtime requirements, process behavior, and evidence that must be collected during the run. \\

Oracle and verification
& Private truth, simulator state, trace evidence, \texttt{score.py}
& Authoritative facts, expected states, criterion definitions, and the executable verifier entry point. \\
\bottomrule

\end{xltabular}
\endgroup

Agent-visible inputs are separated from private truth, simulator state, and verifier logic. For example, \texttt{C1-escalation-export-001} requires one complete ticket export and bounded recovery requests against the same snapshot to obtain twelve escalation codes. The private ledger records the issued codes and requests. The verifier compares them with delivered tool results and the final JSON mapping. This binds task success to the information channel under test without relying on the Agent's account of its own actions.

\paragraph{Oracle sources.}
An oracle may be a truth file, protected fixture, simulator ledger, process outcome, or executable condition. Table~\ref{tab:oracle-sources} lists these forms.

\begingroup
\begin{xltabular}{\linewidth}{@{}P{0.23\linewidth} P{0.30\linewidth} Z@{}}
\caption{Main oracle sources used by TraceTrial-Bench.}
\label{tab:oracle-sources} \\

\toprule
\textbf{Oracle source} & \textbf{Examples} & \textbf{Evaluation role} \\
\midrule
\endfirsthead

\toprule
\textbf{Oracle source} & \textbf{Examples} & \textbf{Evaluation role} \\
\midrule
\endhead

Private structured truth
& Identifiers, codes, receipts, mappings, expected hashes
& Supports exact comparison between required values and final deliverables. \\

Simulator state and ledger
& Accepted requests, issued values, state transitions, call counts
& Records what reached the controlled service independently of the model-visible result. \\

Workspace and artifact state
& File contents, database rows, manifests, output schemas
& Determines whether required changes occurred and whether protected inputs remained unchanged. \\

Execution state
& Exit status, standard output, standard error, signals, timeouts, process state
& Establishes the actual outcome of commands and long-running operations. \\

Trace and probe evidence
& Call-result identity, message delivery, session boundaries, child completion, policy decisions
& Determines what the harness delivered to the model and how execution events were represented. \\
\bottomrule

\end{xltabular}
\endgroup

\paragraph{Milestones and prohibited events.}
A correct final artifact may still violate an interaction contract, for example by repeating a one-time operation or reading private state. Required actions, ordering constraints, and forbidden shortcuts are encoded in prompts, turn schedules, score sheets, and verifiers. These checks preserve the distinction between Agent actions, harness-mediated behavior, and collection failures.

\paragraph{Admission and hard gates.}
Ordinary criteria measure individual requirements; runner hard gates can override an execution-time verifier result; apparatus checks assess whether the evidence is usable. Table~\ref{tab:runner-hard-gates} documents these runner records.

\begingroup
\begin{xltabular}{\linewidth}{@{}P{0.25\linewidth} P{0.29\linewidth} Z@{}}
\caption{Runner-level admission and hard-gate policy, distinct from the final-score aggregation used in the present analysis.}
\label{tab:runner-hard-gates} \\

\toprule
\textbf{Condition} & \textbf{Evidence} & \textbf{Treatment} \\
\midrule
\endfirsthead

\toprule
\textbf{Condition} & \textbf{Evidence} & \textbf{Treatment} \\
\midrule
\endhead

Apparatus admission failure
& Missing or inconsistent run identity, execution record, required trace, oracle evidence, or verifier result
& Admission is withheld before scoring. This protocol state is not an additional exclusion from the finalized archive analyzed in this report. \\

Task-specific prohibited event
& Workspace state, simulator record, process evidence, or normalized trace
& The relevant task criterion fails. The event does not override the whole score unless an explicit policy gives it hard-gate status. \\

Simulator evidence integrity
& Private simulator manifest and its integrity result
& For an otherwise scoreable simulator-backed run, failed integrity triggers a score-zero gate. If integrity cannot be established because of an apparatus failure, the run is invalid and the score override is not applied. \\

Non-simulator tool-call ceiling
& Task-level \path{max_tool_calls} and a complete structured trace
& If the trace proves that the limit was exceeded and no gateway prevented the extra calls from taking effect, the effective score is set to zero. \\

Simulator request ceiling
& Immutable task limit and simulator-side request count
& Requests beyond the limit are rejected before they can affect business state. The excess is recorded as a budget violation and is not charged again through a second score-zero gate. \\

Wall-clock and conversation limits
& Resolved invocation timeout, turn schedule, and execution status
& The limits are enforced during execution. The resulting termination state is retained as run evidence. \\
\bottomrule

\end{xltabular}
\endgroup

\paragraph{Validation and revisions.}
Task validation covers structural consistency, oracle and verifier behavior, trigger reachability, solvability within budget, and shortcuts that could bypass the target mechanism. Cross-harness review checks the same task contract and observable effects. Validation assets are summarized below.

\subsection{Benchmark Coverage}
\label{app:coverage-gaps}

\paragraph{Taxonomy coverage.}
Table~\ref{tab:taxonomy-coverage} counts each task once under its primary component.

\begin{table}[t]
\centering
\caption{Primary component coverage in \texttt{lean95.manifest}.}
\label{tab:taxonomy-coverage}
\begin{tabular}{lrrr}
\toprule
\textbf{Component} & \textbf{Tasks} & \textbf{Share} & \textbf{Properties} \\
\midrule
Context       & 15 & 15.8\% & 6 \\
Execution     & 27 & 28.4\% & 9 \\
Governance    & 9  & 9.5\%  & 4 \\
Lifecycle     & 19 & 20.0\% & 8 \\
Observability & 7  & 7.4\%  & 2 \\
Tools         & 17 & 17.9\% & 10 \\
Verification  & 1  & 1.1\%  & 1 \\
\midrule
Total         & 95 & 100.0\% & 40 \\
\bottomrule
\end{tabular}
\end{table}

\paragraph{Scenario and execution coverage.}
Table~\ref{tab:scenario-execution-coverage} summarizes the task surfaces and interaction forms. Native execution means use of the harness's native interfaces inside the benchmark container. It does not mean execution outside Docker. The L3 label describes task pressure, not model reasoning intensity.

\begin{table}[t]
\centering
\caption{Scenario and execution coverage of the 95 released tasks.}
\label{tab:scenario-execution-coverage}
\begin{tabular}{llrr}
\toprule
\textbf{Facet} & \textbf{Category} & \textbf{Tasks} & \textbf{Share} \\
\midrule
Scenario surface
& Workflow & 59 & 62.1\% \\
& Terminal & 22 & 23.2\% \\
& Coding & 14 & 14.7\% \\
\midrule
Conversation form
& Single-turn & 46 & 48.4\% \\
& Fixed multi-turn & 45 & 47.4\% \\
& Simulator-driven & 4 & 4.2\% \\
\midrule
Execution mode
& Native & 93 & 97.9\% \\
& Docker & 2 & 2.1\% \\
\bottomrule
\end{tabular}
\end{table}

\paragraph{Oracle and evidence coverage.}
The artifact categories in Table~\ref{tab:evidence-coverage} overlap and reflect task-specific oracle and evidence requirements.

\begin{table}[t]
\centering
\caption{Task metadata, oracle, probe, and validation coverage.}
\label{tab:evidence-coverage}
\begin{tabular}{lrr}
\toprule
\textbf{Artifact or declaration} & \textbf{Tasks} & \textbf{Share} \\
\midrule
Executable verifier and both score sheets & 95 & 100.0\% \\
Structured \path{expected_outputs} field & 83 & 87.4\% \\
Private \texttt{truth.json} & 66 & 69.5\% \\
Explicit \texttt{oracle.json} & 19 & 20.0\% \\
Isolation specification & 68 & 71.6\% \\
Business simulator declaration & 44 & 46.3\% \\
Execution probe declaration & 73 & 76.8\% \\
Security probe declaration & 3 & 3.2\% \\
Dedicated \texttt{selftest.py} & 57 & 60.0\% \\
\bottomrule
\end{tabular}
\end{table}

The score sheets contain 783 Harness and 556 Agent criteria: 909 programmatic rules and 430 semantic rubrics in total. Broader limits of this stress-task population and its grading procedure are consolidated in Section~\ref{sec:discussion}.

\section{Harness Configurations and Cross-Harness Adapters}
\label{sec:appendix_B}

\subsection{Harness Selection and Configuration}
\label{app:harness-selection-configuration}

The four harnesses support repeated model--tool interaction, configurable model access, isolated task execution, and observable native activity. Table~\ref{tab:frozen-harness-configurations} identifies the configured packages and interfaces. Resolved configurations are retained with each run.

\begingroup
\begin{xltabular}{\linewidth}{@{}P{0.18\linewidth} P{0.14\linewidth} P{0.29\linewidth} Z@{}}
\caption{Harness versions and controlled execution paths used in TraceTrial-Bench.}
\label{tab:frozen-harness-configurations} \\

\toprule
\textbf{Harness} & \textbf{Version} & \textbf{Controlled execution path} & \textbf{Dockerfile} \\
\midrule
\endfirsthead

\toprule
\textbf{Harness} & \textbf{Version} & \textbf{Controlled execution path} & \textbf{Dockerfile} \\
\midrule
\endhead

OpenClaw
& \texttt{2026.6.9}
& Native OpenClaw runtime with an isolated loopback Gateway
& \path{docker/harnessbench/openclaw.Dockerfile} \\

Codex
& \texttt{0.145.0}
& Native Codex CLI through a loopback LiteLLM gateway
& \path{docker/harnessbench/codex.Dockerfile} \\

Claude Code
& \texttt{2.1.220}
& Native Claude Code CLI through a loopback LiteLLM gateway
& \path{docker/harnessbench/claudecode.Dockerfile} \\

DeepSeek Harness
& \path{0.1.0-rc.6}
& Official SDK with the native JSON-RPC agent runtime
& \path{docker/harnessbench/deepseek.Dockerfile} \\
\bottomrule

\end{xltabular}
\endgroup

Codex and Claude Code use a local LiteLLM gateway. OpenClaw uses its native Gateway. DeepSeek Harness uses its SDK and JSON-RPC runtime. Docker isolates the task workspace and services while native harness interfaces control execution. Multi-turn tasks continue through the same native session unless the task declares a session boundary.

The model registry binds reporting aliases to provider endpoints. The adapter records the resolved configuration and preserves native defaults unless a task, measurement requirement, or compatibility condition requires an override. Reporting groups and the selected execution population are defined in Section~\ref{sec:exp-systems}.

\paragraph{Configuration boundary.}
Common task contracts, initial states, and external limits are benchmark-controlled. System instructions, context policy, built-in tools, permissions, and session behavior remain native. Gateways and adapters provide transport and observation; they must not solve the task, repair model decisions, or supply a missing native capability.

\subsection{Native Capability Discovery and Harness Profiles}
\label{app:native-capability-discovery}

Capability profiles distinguish documented support, exposure to the model, exercise in a run, and behavior verified against external state. Static configuration or the absence of a tool call alone cannot establish availability. Table~\ref{tab:harness-capability-contracts} lists adapter collection contracts: A denotes formal-run support and T task-specific support.

\begingroup
\small
\begin{xltabular}{\linewidth}{@{}Zcccc@{}}
\caption{Measurement-facing capability contracts of the evaluated harnesses.}
\label{tab:harness-capability-contracts} \\

\toprule
\textbf{Capability or evidence contract}
& \textbf{OpenClaw}
& \textbf{Codex}
& \textbf{Claude Code}
& \textbf{DeepSeek Harness} \\
\midrule
\endfirsthead

\toprule
\textbf{Capability or evidence contract}
& \textbf{OpenClaw}
& \textbf{Codex}
& \textbf{Claude Code}
& \textbf{DeepSeek Harness} \\
\midrule
\endhead

Native tool calls and correlated tool results
& A & A & A & A \\

Provider-boundary tool inventory with an explicit completeness status
& A & A & A & A \\

Terminal completion, timeout, process containment, and background-process evidence
& A & A & A & A \\

Native context-compaction boundary observation
& A & A & A & A \\

Task-scoped business tools, service simulators, and MCP interfaces
& T & T & T & T \\

Native child delegation and child-lifecycle observation
& T & T & T & T \\

Asynchronous child dispatch, child-return boundary, and pushed completion
& T & T & T & T \\

Approval, policy, principal-lineage, and managed-path evidence
& T & T & T & T \\
\bottomrule

\end{xltabular}
\endgroup

\paragraph{Native evidence paths.}
OpenClaw combines Gateway results with session JSONL and native parent--child relationships. Codex combines JSONL execution events with thread-resume and session records. Claude Code combines stream-JSON with session archives and native tool-call identifiers. DeepSeek Harness uses persistent JSON-RPC notifications across turns. Task-specific contracts enable observation of compaction, permissions, and asynchronous child return where the runtime exposes them.

\subsection{Adapter Boundary and Canonicalization}
\label{app:adapter-boundary}

Adapters launch the selected harness, expose common task tools through native extensions, collect source records, and normalize them into Trace IR. Benchmark-supervised timeouts, probes, and container events remain distinguishable from native harness actions. Native session continuation is used for multi-turn tasks.

Table~\ref{tab:canonical-event-types} summarizes the event families, and Table~\ref{tab:harness-adapter-mapping} identifies their native inputs. Appendix~\ref{app:trace-ir} supplies the detailed schema.

\begingroup
\small
\begin{xltabular}{\linewidth}{@{}P{0.23\textwidth}Z@{}}
\caption{Main event types produced by the cross-harness adapters.}
\label{tab:canonical-event-types} \\

\toprule
\textbf{Event type} & \textbf{Normalized meaning} \\
\midrule
\endfirsthead

\toprule
\textbf{Event type} & \textbf{Normalized meaning} \\
\midrule
\endhead

\path{model_input}
& A model-visible input or request surface, including available messages and tool definitions when they can be observed safely. \\

\path{tool_inventory}
& The structured tools exposed at a provider-request boundary, together with source, visibility, and completeness information. \\

\path{assistant_delta}
& A native fragment of streamed assistant output. \\

\path{assistant_message}
& A complete or terminal assistant message produced by the harness. \\

\texttt{tool}
& A native tool call and its correlated result, including identifiers, arguments, status, output, and error information when available. \\

\path{token_usage}
& Native or provider-level input, output, reasoning, and cache-token counts, with unavailable fields kept distinct from zero. \\

\path{context_compaction}
& A native context-compaction boundary and its replacement summary when the harness exposes one. \\

\path{subagent_lifecycle}
& The creation, start, return, completion, or failure of a child agent, together with parent-child identity information. \\

\path{policy_event}
& A native approval, denial, permission, sandbox, or policy decision associated with an attempted action. \\

\path{native_session_archive}
& The location, identity, and completeness status of an archived native session or notification stream. \\

\path{run_started}, \path{run_finished}, and \path{run_failed}
& Benchmark lifecycle events that record process status without being treated as model or harness actions. \\
\bottomrule

\end{xltabular}
\endgroup

\begingroup
\small
\begin{xltabular}{\linewidth}{@{}P{0.14\textwidth}P{0.28\textwidth}Z@{}}
\caption{Native evidence sources and their adapter mappings.}
\label{tab:harness-adapter-mapping} \\

\toprule
\textbf{Harness} & \textbf{Native evidence source} & \textbf{Adapter mapping} \\
\midrule
\endfirsthead

\toprule
\textbf{Harness} & \textbf{Native evidence source} & \textbf{Adapter mapping} \\
\midrule
\endhead

OpenClaw
& Gateway-backed CLI output and session JSONL records
& Parses the structured CLI result and recovers missing trajectory information from the session bound to the run. Parent and child sessions remain linked through native session identifiers. \\

Codex
& Native \texttt{exec} JSONL stream and session archives
& Maps command execution, MCP calls, collaboration events, assistant messages, and usage records. Native thread identifiers are retained for continuation and child-session discovery. \\

Claude Code
& Native stream-JSON output and session archives
& Matches tool calls and results through native identifiers, removes duplicate stream records, and retains messages, reasoning output, usage, compaction, and child-agent evidence. \\

DeepSeek Harness
& SDK results and persistent JSON-RPC notifications
& Reconstructs the model-visible sequence from incremental notifications and retains tool interactions, compaction, subagent lineage, usage, and policy events across turns. \\
\bottomrule

\end{xltabular}
\endgroup

\paragraph{Canonicalization rules.}
The following rules preserve source meaning while allowing a common representation.

\begin{enumerate}
    \item \textbf{Native content is preserved.}
    Prompts, assistant messages, tool arguments, tool outputs, errors, and compaction summaries are not rewritten for semantic consistency. Secret values may be redacted, but the surrounding record and its structure remain unchanged.

    \item \textbf{Tool calls and results are matched conservatively.}
    Native tool-call identifiers are used whenever they are available. A call without an observable result remains in the trace with a missing result rather than being removed or assigned a successful outcome. An unmatched result is also retained as incomplete evidence.

    \item \textbf{Native identifiers remain recoverable.}
    Session, message, tool-call, and child-agent identifiers are preserved in their native form. If a harness reuses the same tool-call identifier across conversation turns, the merged trace assigns a turn-scoped normalized identifier and stores the original value as \path{native_tool_call_id}.

    \item \textbf{Event order follows the strongest available native order.}
    Stream order, native sequence numbers, and session-record order take priority over adapter observation timestamps. The final trace receives a contiguous sequence number for inspection. Source sequence numbers and timestamps remain attached to imported events.

    \item \textbf{Distinct principals are not flattened.}
    Actions performed by the main agent and actions performed by child agents retain separate identities. Parent-child relationships are recorded explicitly, and a child action is not reassigned to the parent merely because both appear in the same session archive.

    \item \textbf{Unknown and damaged records remain visible.}
    A valid but unrecognized native event is retained as a generic native or raw event. If a stream contains malformed structured output, the affected record is preserved and the structured-trace status is downgraded. The adapter does not silently discard the parsing failure.

    \item \textbf{Missing information is not reconstructed without evidence.}
    An adapter does not generate hidden reasoning, omitted tool results, approval decisions, or lost context. A native compaction event records only the summary and boundary exposed by the harness, but it does not restore messages removed by the compaction process.

    \item \textbf{Apparatus evidence remains separate.}
    Process timeouts, container enforcement, provider-request records, and task-tool transport logs are labeled according to their actual source. They may support evaluation or attribution, but they are not converted into native harness events.
\end{enumerate}

For Codex command events, normalization removes only the fixed outer shell wrapper added by the evaluated executor. Agent-supplied pipes, redirections, and nested wrappers remain unchanged, and the full native command remains available. Task-tool transport may change its envelope but not arguments, responses, state effects, or error meaning.

\subsection{Semantic Equivalence and Eligibility}
\label{app:semantic-equivalence}

This subsection defines prospective comparison requirements. It does not introduce an additional filter on the archived results in Section~\ref{sec:results}. Equivalent paths preserve the task goal, information, action opportunities, state effects, and evaluation conditions while leaving harness-controlled differences observable.

\begin{enumerate}
    \item \textbf{Goal equivalence.}
    The model receives the same task objective, required deliverables, constraints, and success conditions. Formatting changes required by a native input interface must not add, remove, or clarify task information.

    \item \textbf{Initial-state equivalence.}
    The workspace, files, service records, credentials, environment variables, and simulated external state are initialized from the same task specification. Harness-specific configuration files may differ, but they must not contain task answers or additional problem-solving information.

    \item \textbf{Task-tool equivalence.}
    Benchmark-supplied tools must represent the same task-level operations. Their required inputs, returned information, errors, and effects on external state must remain equivalent even when the harness uses a different transport mechanism.

    \item \textbf{Action-effect equivalence.}
    The same valid task action must produce the same externally observable state change. An adapter must not execute an operation on behalf of the model, repeat a failed operation automatically, or convert a rejected operation into a successful one.

    \item \textbf{Observation equivalence.}
    Tool results, errors, completion signals, and service responses must preserve the information returned by the task environment. Harness-specific presentation may differ, but the adapter must not summarize, expand, or repair the information before it reaches the model.

    \item \textbf{Lifecycle equivalence.}
    Turn limits, wall-clock limits, model-call limits, process boundaries, and session-reset rules must follow the same task specification. A native continuation, compaction, background process, or child-agent mechanism must not be replaced with a stronger benchmark-controlled alternative.

    \item \textbf{Policy equivalence.}
    Managed paths, approval requirements, prohibited actions, and task-specific access restrictions must apply to the same operations. A benchmark safety backstop may prevent damage, but it cannot be counted as evidence that the harness itself enforced the policy.

    \item \textbf{Evidence equivalence.}
    The evidence required by the task oracle must be observable with comparable strength. A structured native event cannot be replaced by an adapter-generated statement that the event probably occurred. When one harness exposes weaker evidence, the affected comparison is restricted or marked non-comparable.
\end{enumerate}

\paragraph{Pair-level eligibility.}
Let $R_t$ be the capabilities needed to administer and observe task $t$, $C_h$ the harness's measurement contract, $S_{t,h}$ semantic equivalence, and $O_{t,h}$ the required observation path. Then

\begin{align}
E_{t,h}
=
\mathbb{I}
\left[
R_t \subseteq C_h
\land S_{t,h}
\land O_{t,h}
\right],
\end{align}

A target capability need not be present when its absence is the behavior under test. A pair is unsupported if the test cannot be administered or observed, and non-comparable if the adapted path changes its semantics.

\paragraph{Run-level validity.}
For an eligible pair, let $A_{t,h,r}$ denote apparatus health and $I_{t,h,r}$ source and archive integrity. Protocol admission is

\begin{align}
V_{t,h,r}
=
E_{t,h}
\land A_{t,h,r}
\land I_{t,h,r}.
\end{align}

Admission concerns evidence integrity, not task success. A model action that bypasses a measurement channel is annotated separately, but it does not erase a finalized zero. The full-grid reporting policy is specified in Section~\ref{sec:exp-admission}.

A collector or archive failure can make an observation inconclusive. A harness timeout or absent event remains measurable when the observation path itself is intact. Table~\ref{tab:eligibility-treatment} summarizes the protocol distinctions.

\begingroup
\small
\begin{xltabular}{\linewidth}{@{}P{0.18\textwidth}P{0.48\textwidth}Z@{}}
\caption{Treatment of eligibility and validity outcomes.}
\label{tab:eligibility-treatment} \\

\toprule
\textbf{Status} & \textbf{Meaning} & \textbf{Statistical treatment} \\
\midrule
\endfirsthead

\toprule
\textbf{Status} & \textbf{Meaning} & \textbf{Statistical treatment} \\
\midrule
\endhead

Eligible
& A semantically valid and measurable native execution path is available.
& Included in the cross-harness eligible population. \\

Unsupported
& The benchmark lacks a valid administration or observation path; this is distinct from an absent target capability.
& Outside the prospective protocol population; not an additional filter on the current archive. \\

Non-comparable
& A possible adapter path would change task semantics, native behavior, or evidence strength.
& Report the protocol limitation; the current analysis retains its declared full grid. \\

Measurement bypass
& An Agent action obscured a target Harness property.
& Preserve raw verdicts and annotate the affected criteria; adjust only the optional attribution set. \\

Apparatus-invalid
& The benchmark could not execute, collect, or verify the run as specified.
& Withhold protocol admission; do not infer a Harness defect. No new archival exclusion is applied in this study. \\

Measured failure
& The pair was eligible, the apparatus was valid, and the observed behavior failed one or more criteria.
& Included as a failure in the applicable denominator. \\
\bottomrule

\end{xltabular}
\endgroup

\paragraph{Matched comparisons.}
For a prospective study using protocol-based admission, the matched population for two harnesses is

\begin{align}
\mathcal{D}_{h_1,h_2}
=
\left\{
(t,m,r)
\mid
E_{t,h_1}=E_{t,h_2}=1,
V_{t,h_1,r}=V_{t,h_2,r}=1
\right\}.
\end{align}

The present study instead retains all finalized cells in its fixed archive, with 855 model--task cells per harness. Each model--task cell contributes one selected execution.

\section{Evidence Acquisition and Trace IR}
\label{sec:appendix_C}

\subsection{Evidence Sources and Acquisition Architecture}
\label{app:evidence-acquisition}

The collection pipeline preserves native harness records, provider-boundary metadata, task-tool ledgers, environment state, and apparatus records. Table~\ref{tab:evidence-layers} specifies their producers, visibility, and uses. The runner captures an initial workspace, collects observations during execution, and archives the final workspace and source records for offline evaluation.

\begingroup
\small
\begin{xltabular}{\linewidth}{@{}P{0.16\textwidth}P{0.18\textwidth}P{0.18\textwidth}Z@{}}
\caption{Evidence layers collected for each benchmark run.}
\label{tab:evidence-layers} \\

\toprule
\textbf{Evidence layer}
& \textbf{Producer}
& \textbf{Model visibility}
& \textbf{Examples and primary use} \\
\midrule
\endfirsthead

\toprule
\textbf{Evidence layer}
& \textbf{Producer}
& \textbf{Model visibility}
& \textbf{Examples and primary use} \\
\midrule
\endhead

Native harness evidence
& Harness runtime
& Mixed
& Assistant messages, native tool calls and results, session records, compaction events, child-agent events, policy decisions, and native usage reports. Used to reconstruct harness behavior. \\

Provider-boundary evidence
& Request recorder
& Mixed
& Model-visible tool declarations, request counts, numeric timing observations, and provider usage fields. Used to confirm the request surface and derive comparable efficiency measurements. \\

Task-tool evidence
& Task tools and service simulators
& Responses are visible, transport records are observer-only
& Tool requests, returned values, errors, committed operations, and service-state transitions. Used to verify the effects of model-initiated actions. \\

Environment-state evidence
& Workspace and task environment
& Depends on task permissions
& Baseline and final files, workspace differences, output artifacts, and final external-service state. Used to determine whether required results were produced. \\

Apparatus evidence
& Benchmark runner and container runtime
& Not visible
& Process status, exit code, timeout, image identity, resource limits, archive status, cleanup status, and collector health. Used to determine whether the run was measured correctly. \\
\bottomrule

\end{xltabular}
\endgroup

\paragraph{Visibility and authority.}
A task-tool ledger establishes what a service received, produced, or committed. Only a correlated model-facing record establishes what the model received through the harness. Likewise, a workspace or simulator snapshot establishes external state, while an assistant message records a claim about that state. Observer-only records, including private truth and apparatus logs, are not treated as information available to the Agent.

The provider recorder retains structured tool declarations and numeric request metadata rather than full prompts or conversation content. A complete inventory supports a claim about the structured tool surface at that request boundary; but it does not describe every possible native action interface. Native traces contain only exposed reasoning or assistant content, without reconstructing hidden reasoning.

\paragraph{Artifacts and partial collection.}
The run archive stores normalized events in \texttt{events.json}, a derived overview in \texttt{summary.json}, native records, workspace changes, and applicable service or probe evidence. The root \texttt{result.json} indexes execution artifacts, and final scoring authority is specified in Section~\ref{sec:exp-protocol}. Collected records remain available after errors or timeouts, and malformed or unsupported native records are retained as raw evidence. Missing events are never synthesized to complete a trajectory.

\subsection{Trace Reconstruction and Source Precedence}
\label{app:trace-reconstruction}

Reconstruction parses native records, binds them to the run and session, correlates exchanges, merges resumed turns, and preserves supported ordering. Native identities and explicit relationships take precedence over text similarity or timestamp proximity.

\paragraph{Calls, turns, and principals.}
Tool calls and results are matched by native call identifiers. Repeated exports are merged only when identities and semantic content agree. Unmatched calls or results remain incomplete. Multi-turn events retain \path{conversation_turn}, \path{source_event_seq}, and \path{source_event_timestamp}. If a later turn reuses an identifier for a different call, the normalized identifier becomes turn-scoped while \path{native_tool_call_id} preserves the source value.

Child sessions retain their own principal and a typed parent relationship. Spawn, return, and completion records establish cross-session dependencies,
while local event order is preserved within each principal. Background jobs are linked through exposed job or process identifiers. External completion and delivery to the parent or model are distinct observations.

\paragraph{Context and ordering.}
Persisted history is not necessarily active model context. Native model-input records, when available, establish the messages
delivered at a step, while compaction records establish only the
boundary and summary actually exposed. Timestamps may describe creation, observation, or export and cannot establish cross-session causality by themselves. The final contiguous \texttt{seq} index is a presentation order, not a stronger causal claim.

\paragraph{Claim-specific precedence.}
Table~\ref{tab:source-precedence} identifies the preferred source for each question. Conflicting lower-precedence observations remain in the archive: for example, a native success message and a simulator's failed transaction describe different boundaries and may reveal a delivery problem. Ambiguous session bindings instead require integrity review.

\begingroup
\small
\begin{xltabular}{\linewidth}{@{}P{0.23\textwidth}P{0.31\textwidth}Z@{}}
\caption{Claim-specific source precedence used during trace reconstruction.}
\label{tab:source-precedence} \\

\toprule
\textbf{Claim}
& \textbf{Preferred evidence}
& \textbf{Fallback and limitation} \\
\midrule
\endfirsthead

\toprule
\textbf{Claim}
& \textbf{Preferred evidence}
& \textbf{Fallback and limitation} \\
\midrule
\endhead

Tool availability
& Complete model-visible provider-boundary inventory
& An observed native call proves presence only. It cannot establish that other tools were absent. \\

Model-visible context
& Native model-input or request-surface record
& Persisted session history may support reconstruction, but it does not prove that every message remained active. \\

Native tool invocation
& Native tool-call event with a stable identifier
& A task-service request proves that an operation reached the service, but it may not identify the responsible principal. \\

Tool-result delivery
& Native tool-result or later model-input record containing the result
& A service response proves that the service produced a result, but not that the model received it. \\

External state change
& Simulator ledger, authoritative service state, or final workspace
& Tool output and assistant text describe the claimed effect but do not override authoritative state. \\

Compaction, subagent, or policy event
& Typed native event bound to the correct session or principal
& Assistant text may support interpretation but cannot create the missing native event. \\

Model-request count
& Complete provider-request sequence
& Per-call native usage may provide a derived count. Turn-level or run-level aggregates do not. \\

Token usage
& Complete per-request provider usage
& Native per-call usage is used when provider evidence is incomplete. Run aggregates remain labeled as aggregate evidence. \\

Timing
& Provider-boundary request and response observations
& Native CLI stream timing is used only when its scope and coverage are known. \\

Process completion and timeout
& Process supervisor and container lifecycle
& A harness message may explain the stop condition but does not override the observed process status. \\

Apparatus validity
& Container lifecycle, archive manifest, integrity checks, and collector status
& Harness output cannot establish that the measurement apparatus completed correctly. \\
\bottomrule

\end{xltabular}
\endgroup

Provider usage replaces a native aggregate only when request coverage is complete. Timing and token quantities must cover the same requests before they are combined.

\subsection{Trace IR Schema and Event Dictionary}
\label{app:trace-ir}

Trace IR separates a common envelope from event-specific payloads. Tables~\ref{tab:trace-ir-envelope} and~\ref{tab:trace-ir-core-events} define the required fields and core events. Optional identities and values are populated only when supported by native evidence.

\begingroup
\small
\begin{xltabular}{\linewidth}{@{}P{0.21\textwidth}P{0.16\textwidth}Z@{}}
\caption{Common fields in the Trace IR event envelope.}
\label{tab:trace-ir-envelope} \\

\toprule
\textbf{Field} & \textbf{Type} & \textbf{Definition} \\
\midrule
\endfirsthead

\toprule
\textbf{Field} & \textbf{Type} & \textbf{Definition} \\
\midrule
\endhead

\path{schema_version}
& integer
& Trace IR schema version. The current release uses version 1. \\

\texttt{seq}
& positive integer
& Contiguous event position in the final reconstructed trace. \\

\texttt{timestamp}
& string
& UTC timestamp assigned when the normalized event is written. This is an observation timestamp rather than a native execution timestamp. \\

\path{run_id}
& string
& Identifier shared by all events and artifacts belonging to one benchmark run. \\

\path{event_type}
& string
& Canonical type that determines the event-specific payload. \\

\texttt{source}
& string or absent
& Native stream, session archive, provider boundary, or apparatus component from which the event was obtained. \\

\path{conversation_turn}
& positive integer or absent
& Benchmark-managed user-input turn to which the event belongs. \\

\path{session_id}
& string or absent
& Native harness session, thread, or conversation identifier. \\

\path{turn_id}
& string or integer or absent
& Native identifier for a model or harness turn. \\

\path{step_id}
& string or integer or absent
& Native identifier for an internal model step when the harness exposes one. \\

\path{principal_id}
& string or absent
& Identity of the main agent or child agent responsible for the event. \\

\path{parent_principal_id}
& string or absent
& Identity of the parent principal when the event belongs to a child agent. \\

\path{principal_role}
& string or absent
& Role of the responsible principal, normally \texttt{main} or \texttt{child}. \\

\path{source_event_seq}
& integer or absent
& Original sequence number before a turn or session trace was merged. \\

\path{source_event_timestamp}
& string or absent
& Observation timestamp copied from the source trace during merging. \\

\path{source_timestamp}
& string or absent
& Timestamp supplied directly by the native harness record. \\

\path{raw_event}
& JSON value or absent
& Native source record associated with the normalized event after required redaction. \\
\bottomrule

\end{xltabular}
\endgroup

The normalized \texttt{timestamp} records collection time, while source
timestamps remain in \texttt{source\_timestamp} or \texttt{raw\_event}. Run, session, turn, step, and principal identifiers locate each observation without collapsing concurrent agents into one execution stream.

\begingroup
\small
\begin{xltabular}{\linewidth}{@{}P{0.20\textwidth}P{0.31\textwidth}Z@{}}
\caption{Core Trace IR event types and payload fields.}
\label{tab:trace-ir-core-events} \\

\toprule
\textbf{Event type}
& \textbf{Main payload fields}
& \textbf{Meaning} \\
\midrule
\endfirsthead

\toprule
\textbf{Event type}
& \textbf{Main payload fields}
& \textbf{Meaning} \\
\midrule
\endhead

\path{model_input}
& \texttt{model}, \texttt{provider}, \texttt{messages}, \texttt{tools}, \path{request_params}
& A model invocation surface exposed by the native harness. The completeness of messages and tools depends on the source. \\

\path{tool_inventory}
& \path{model_visible}, \texttt{complete}, \texttt{scope}, \texttt{tools}, \path{request_count}, \path{inventory_fingerprint}
& Structured tool declarations observed at the provider boundary. Each tool contains a name, type, description, and input schema. \\

\path{assistant_delta}
& \texttt{delta} or \path{content_delta}, \path{reasoning_delta}, \path{content_kind}, \path{fragment_seq}
& A fragment of streamed assistant text or reasoning content exposed by the harness. \\

\path{assistant_message}
& \path{message_id}, \texttt{content}, \path{content_kind}, \texttt{reasoning}, \texttt{model}, \path{finish_reason}
& A complete assistant message or complete reasoning block. A reasoning block is not treated as the final answer. \\

\texttt{tool}
& \path{tool_call}, \path{tool_result}
& A native tool call and its correlated result. The result may be null when no matching native result was observed. \\

\path{token_usage}
& \texttt{usage}, \path{usage_scope}, \path{usage_aggregate}, \path{reported_cost_usd}
& Token usage associated with a model call, turn, complete run, or internal harness operation. \\

\path{context_compaction}
& \path{native_event_type}, \path{native_event_id}, \path{native_event_key}, \texttt{trigger}, \texttt{summary}, \path{summary_present}, \path{notice_visible_to_model}
& A native context-compaction, pruning, or replacement boundary. \\

\path{subagent_lifecycle}
& \texttt{action}, \texttt{status}, \path{stop_reason}, principal fields, \path{boundary_kind}, \path{pending_child_ids}
& Creation, dispatch, start, return, completion, or failure of a native child agent. \\

\path{task_completion}
& \texttt{status}, \texttt{content}, \path{model_visible_content}, principal fields, \path{parent_reactivated}
& A native child or delegated task result that reaches a completion boundary. \\

\path{policy_event}
& \path{policy_kind}, \texttt{decision}, \path{reason_code}, \path{approval_id}, \path{tool_call_id}, \path{tool_name}, \path{reason_visible_to_model}, \path{execution_started}
& A native approval, permission, sandbox, or policy decision. \\

\path{native_event}
& \path{native_event_type} or \path{native_method}, \texttt{status}
& A valid native record that does not map to a more specific core event. \\

\path{native_diagnostic}
& \path{diagnostic_type}, diagnostic-specific fields
& A native diagnostic that may explain a process result but does not itself represent a model action. \\
\bottomrule

\end{xltabular}
\endgroup

\paragraph{Tool exchanges.}
A \texttt{tool} event contains a \path{tool_call} with its identity, name, and arguments, and a correlated \path{tool_result} when available. Result fields preserve content, source-indicated error status, integer exit code, duration, and known truncation. Error status is not inferred from negative wording. A missing result is represented as

\begin{align}
\mathtt{tool\_result}=\mathtt{null}.
\end{align}

This differs from a successfully returned empty string.

\paragraph{Usage scope.}
Usage retains whether it describes an individual call or an aggregate:

\begin{itemize}
    \item \texttt{call} for one model request
    \item \texttt{turn} for a native harness turn
    \item \texttt{run} for a complete execution aggregate
    \item \texttt{internal} for a harness operation such as a compaction call
\end{itemize}

Cached input is included in total input, and reasoning output is included
in total output. Neither is added again. Missing usage remains unavailable,
and an aggregate is not expanded into estimated calls.

\paragraph{Compaction, child lifecycle, and policy.}
Compaction records retain a native identity or record hash, the exposed replacement summary, and whether a summary was present. A visible compaction notice does not establish preservation of all relevant context. Child lifecycle events retain action, principal, and parent identities.
A child completion does not establish parent reactivation without a
corresponding native observation. Policy events retain the native decision, supported reason, visibility, and whether execution began. An outer-container denial remains apparatus evidence rather than a native harness policy decision.

\paragraph{Support and extension events.}
Table~\ref{tab:trace-ir-support-events} documents collection and orchestration events, which are not counted as model actions.

\begingroup
\small
\begin{xltabular}{\linewidth}{@{}P{0.23\textwidth}P{0.29\textwidth}Z@{}}
\caption{Lifecycle, orchestration, and evidence-support events.}
\label{tab:trace-ir-support-events} \\

\toprule
\textbf{Event type}
& \textbf{Main payload fields}
& \textbf{Meaning} \\
\midrule
\endfirsthead

\toprule
\textbf{Event type}
& \textbf{Main payload fields}
& \textbf{Meaning} \\
\midrule
\endhead

\path{run_started}
& \texttt{harness}, \texttt{model}, \texttt{cwd}, \texttt{argv}, \texttt{isolation}
& Start of the harness execution or task protocol. \\

\path{run_finished}
& \path{exit_code}, \path{trace_complete}, \path{summary_path}
& Normal end of the observed execution process. \\

\path{run_failed}
& \path{error_type}, \path{error_message}, \path{exit_code}, \path{timed_out}
& Abnormal termination, timeout, unsupported protocol, or infrastructure failure. \\

\path{conversation_started}
& \texttt{mode}, \path{planned_turns}, \path{max_turns}
& Start of a benchmark-managed multi-turn conversation. \\

\path{conversation_turn_started}
& \path{turn_index}, \texttt{input}, \path{user_input}
& Delivery of one benchmark-managed user turn. \\

\path{conversation_turn_finished}
& \path{turn_index}, \texttt{status}, \path{native_session_id}
& Completion status of one conversation turn. \\

\path{conversation_simulator_decision}
& \path{turn_index}, \texttt{done}, \path{next_message}
& Decision produced by a task-defined conversation simulator. \\

\path{conversation_finished}
& \texttt{status}, \path{turns_completed}, \path{native_session_id}
& End of the complete benchmark-managed conversation. \\

\path{session_boundary}
& previous and new session identifiers, \path{before_turn}, \path{workspace_preserved}, \path{history_replayed}
& Explicit transition from one native session to another. \\

\path{context_handoff}
& \texttt{action}, \path{before_turn}, \path{model_visible}, history count and hash
& Benchmark-controlled delivery of permitted prior context at a declared session boundary. \\

\path{provider_request_count}
& \texttt{count}, \texttt{source}
& Native or provider-supported count of model requests when exposed as a trace event. \\

\path{native_session_archive}
& \texttt{status}, \texttt{format}, \texttt{artifact}, \texttt{sha256}, \path{size_bytes}, \texttt{redacted}
& Metadata for an archived native session or notification stream. \\

\path{trace_export}
& \texttt{status}, \texttt{format}, \texttt{path}, \texttt{redacted}, export-specific checks
& Status of a native trace-export operation. \\

\path{raw_stdout}
& \texttt{text}, \path{stream_seq}, \texttt{parseable}, \path{parse_error}
& Standard output that could not be represented by a more specific event. \\

\path{raw_stderr}
& \texttt{text}, \path{stream_seq}, \texttt{severity}
& Standard error retained for diagnosis after redaction. \\
\bottomrule

\end{xltabular}
\endgroup

Task-declared memory and lifecycle protocols may additionally use:

\begin{itemize}
    \item \path{session_created}
    \item \path{episode_injected}
    \item \path{fresh_session_completed}
    \item \path{delegated_parent_created}
    \item \path{delegated_child_created}
    \item \path{delegated_child_completed}
    \item \path{final_prompt_completed}
    \item \path{app_server_notifications}
\end{itemize}

These receipts retain native request, session, child, or episode identities and content hashes. They supplement the core events only for the corresponding protocol.

\paragraph{Mapping and missing values.}
Table~\ref{tab:raw-canonical-mapping} summarizes normalization. Missing fields, explicitly unavailable values, measured zeros, and empty successful outputs retain their distinct meanings. The complete native record remains associated with its canonical event.

\begingroup
\small
\begin{xltabular}{\linewidth}{@{}P{0.31\textwidth}P{0.22\textwidth}Z@{}}
\caption{General mapping from native records to Trace IR events.}
\label{tab:raw-canonical-mapping} \\

\toprule
\textbf{Native record category}
& \textbf{Trace IR event}
& \textbf{Mapping rule} \\
\midrule
\endfirsthead

\toprule
\textbf{Native record category}
& \textbf{Trace IR event}
& \textbf{Mapping rule} \\
\midrule
\endhead

Request start or turn start
& \path{model_input}
& Preserve the exposed model, messages, tools, parameters, and native turn identity. \\

Structured provider tool declaration
& \path{tool_inventory}
& Normalize name, type, description, and input schema while retaining completeness and visibility fields. \\

Streamed text or reasoning fragment
& \path{assistant_delta}
& Preserve the text fragment, content kind, fragment order, and source timing. \\

Completed assistant content
& \path{assistant_message}
& Preserve message identity, visible content, content kind, and finish reason. \\

Tool-use start followed by result
& \texttt{tool}
& Match through the native call identifier and retain both source records. \\

Tool-use start without result
& \texttt{tool}
& Preserve the call and set \path{tool_result} to null. \\

Native usage record
& \path{token_usage}
& Preserve token fields and record whether their scope is a call, turn, run, or internal action. \\

Native compaction or pruning record
& \path{context_compaction}
& Preserve the boundary, replacement summary, session identity, and model-visibility status. \\

Child-agent lifecycle record
& \path{subagent_lifecycle}
& Preserve child identity, parent identity, action, status, and return boundary. \\

Delegated result delivered to a parent
& \path{task_completion}
& Preserve the child result and whether the parent was reactivated. \\

Approval, permission, or sandbox record
& \path{policy_event}
& Normalize the decision while retaining the complete native policy record. \\

Valid but unrecognized native record
& \path{native_event}
& Preserve the source type and complete raw record without assigning unsupported semantics. \\

Malformed or non-JSON stream content
& \path{raw_stdout} or \path{raw_stderr}
& Preserve the original text and record parsing or severity information. \\
\bottomrule

\end{xltabular}
\endgroup

\subsection{Evidence Quality, Integrity, and Privacy}
\label{app:evidence-quality}

Table~\ref{tab:evidence-quality-dimensions} defines the quality checks and reporting dimensions. Structural validity, criterion coverage, and support for attribution
are separate properties. The table is a measurement specification,
not a claim that every dimension has a corpus-level estimate in this study.

\begingroup
\small
\begin{xltabular}{\linewidth}{@{}P{0.18\textwidth}P{0.33\textwidth}Z@{}}
\caption{Evidence-quality dimensions used by TraceTrial-Bench.}
\label{tab:evidence-quality-dimensions} \\

\toprule
\textbf{Dimension}
& \textbf{Question}
& \textbf{Main checks} \\
\midrule
\endfirsthead

\toprule
\textbf{Dimension}
& \textbf{Question}
& \textbf{Main checks} \\
\midrule
\endhead

Completeness
& Were the records required for the intended measurement captured?
& Trace termination, structured-event availability, session coverage, call-result coverage, provider-request coverage, and required probe artifacts. \\

Fidelity
& Does the normalized event preserve the meaning and content of its native source?
& Native identifiers, arguments, outputs, errors, status fields, raw records, and allowed normalization rules. \\

Source traceability
& Can a normalized claim be traced to a native record or an authoritative state artifact?
& Source labels, session and principal identities, source sequence numbers, artifact paths, and hashes. \\

Integrity
& Are the stored artifacts complete, unchanged, and bound to the correct run?
& File-set manifests, SHA-256 values, run identity, image identity, path safety, lifecycle checks, and evidence-schema validation. \\

Privacy
& Were credentials and private benchmark materials excluded from released evidence?
& Field-level redaction, output-tree scans, path scans, provider-record restrictions, and separation of private verifier inputs. \\

Replayability
& Can the run be inspected and scored again without rerunning the agent?
& Preserved workspace, trace, task state, scoring-input manifest, frozen configuration, and verifier-gate checks. \\

Bounded overhead
& Did evidence collection remain within declared time, memory, output, and storage limits?
& End-to-end timing, peak memory, captured output size, archive limits, and collector status. \\
\bottomrule

\end{xltabular}
\endgroup

\paragraph{Completeness and scoreability.}
The \path{trace_complete} field records whether the configured collection path reached its expected end, not whether all harness internals were observable. \path{structured_trace_status} records whether usable structured events were extracted. Table~\ref{tab:evidence-quality-states} summarizes these states. The current archive inclusion policy appears in Section~\ref{sec:exp-admission}. criterion handling appears in Appendix~\ref{app:verdict-rules}.

\begingroup
\small
\begin{xltabular}{\linewidth}{@{}P{0.20\textwidth}P{0.48\textwidth}Z@{}}
\caption{Run-level evidence-quality states.}
\label{tab:evidence-quality-states} \\

\toprule
\textbf{State}
& \textbf{Condition}
& \textbf{Treatment} \\
\midrule
\endfirsthead

\toprule
\textbf{State}
& \textbf{Condition}
& \textbf{Treatment} \\
\midrule
\endhead

Complete
& Structured extraction succeeded, the normal acquisition path finished, and all evidence required by the reported measurements passed validation.
& All supported criteria may be evaluated. \\

Partial
& Some valid structured or state evidence was preserved, but the run ended early or one or more expected sources were incomplete.
& Check admission and retain coverage limitations. Admitted criteria follow the fixed scoring rules. \\

Unavailable
& No usable structured trajectory was recovered, although raw output or final artifacts may remain.
& Check whether required-source admission fails. Do not silently remove finalized trace criteria from a score. \\

Integrity failure
& Evidence cannot be bound to the run, does not match its manifest, contains an unsafe artifact, or fails a required schema check.
& The affected measurement is rejected and may make the run apparatus-invalid. \\
\bottomrule

\end{xltabular}
\endgroup

\paragraph{Integrity and release boundary.}
Archive manifests bind file paths, sizes, and hashes to the run configuration and source identities. Task-specific probes and service ledgers retain their own provenance. These checks detect archive changes and incorrect bindings, but they do
not establish the truth of every source record. Permitted representation changes are listed in Appendix~\ref{app:adapter-boundary}.

Private truth and verifier inputs are separated from the Agent's writable workspace. The export policy excludes credentials and full provider request bodies and requires redaction of sensitive archived content. Offline scoring reads preserved evidence and verifies a workspace copy.
It does not replay the Agent or guarantee reproduction of its decisions.

\section{Layered Evaluation and Failure Attribution}
\label{sec:appendix_D}

\subsection{Evaluation Layers and Criterion Catalog}
\label{app:evaluation-layers}

The Agent layer scores task actions and deliverables, and the Harness layer scores the mediated execution behavior. Apparatus checks establish evidence integrity without contributing a third score (Table~\ref{tab:evaluation-layers}). The two verdict vectors remain separate, with one explicit dependency: Harness state 2 uses the completed Agent outcome.

\begingroup
\small
\begin{xltabular}{\linewidth}{@{}P{0.18\linewidth} P{0.19\linewidth} Z@{}}
\caption{Evaluation layers used in TraceTrial-Bench.}
\label{tab:evaluation-layers} \\

\toprule
\textbf{Layer} & \textbf{Role} & \textbf{Main evaluation target} \\
\midrule
\endfirsthead

\toprule
\textbf{Layer} & \textbf{Role} & \textbf{Main evaluation target} \\
\midrule
\endhead

Agent & Scored, 50 points & Correctness, completeness, and task compliance of the agent's actions, claims, and final deliverables. \\
Harness & Scored, 50 points & Faithful delivery of context, tools, execution results, state, errors, permissions, and lifecycle events. \\
Apparatus & Admission gate & Integrity of traces, probes, runner metadata, private evidence, and archived artifacts used by the evaluator. \\
\bottomrule

\end{xltabular}
\endgroup

The task score sheets are \path{verifier/harness_score.yaml} and \path{verifier/agent_score.yaml}. A criterion specifies its identifier, type, statement, and required evidence:

\begin{align}
c = \langle id, layer, type, statement, evidence, component, trigger, role \rangle .
\end{align}

\begingroup
\small
\begin{xltabular}{\linewidth}{@{}P{0.20\linewidth} Z@{}}
\caption{Fields used to describe an evaluation criterion.}
\label{tab:criterion-fields} \\

\toprule
\textbf{Field} & \textbf{Meaning} \\
\midrule
\endfirsthead

\toprule
\textbf{Field} & \textbf{Meaning} \\
\midrule
\endhead

\texttt{id} & A stable identifier that is unique across the two score sheets of a task. \\
\texttt{layer} & The scored layer to which the criterion belongs, either Harness or Agent. \\
\texttt{type} & The evaluation method, either a deterministic rule or an evidence-bounded rubric. \\
\texttt{statement} & A direct description of the condition that must hold for the criterion to pass. \\
\texttt{evidence} & The event types, fields, keywords, and matching rules used to select relevant evidence. \\
\texttt{component} & The harness taxonomy component examined by the criterion. \\
\texttt{trigger} & The condition under which a mechanism-specific criterion becomes applicable. \\
\texttt{role} & Whether the criterion contributes to a core metric or provides diagnostic information. \\
\bottomrule

\end{xltabular}
\endgroup

A \emph{rule} uses deterministic checks of task state, artifacts, or exact trace properties. A \emph{rubric} requires semantic judgment of a bounded evidence packet. Taxonomy ownership and task construction are defined in Appendix~\ref{sec:appendix_A}.

\paragraph{Score composition.}
Let $P_H,P_A$ be passed criterion counts and $N_H,N_A$ the corresponding totals. Each layer contributes up to 50 points to the combined score:

\begin{align}
S_H &= 50 \times \frac{P_H}{N_H}, \\
S_A &= 50 \times \frac{P_A}{N_A}, \\
S_{\mathrm{total}} &= S_H + S_A.
\end{align}

These contributions differ from the separately reported layer percentages, $100P_H/N_H$ and $100P_A/N_A$. A combined pass requires $S_{\mathrm{total}}=100$; Agent task success requires only $P_A=N_A$.

\subsection{Verdict Rules and Evaluation Procedure}
\label{app:verdict-rules}

Deterministic rules are binary. Semantic Harness rubrics may return diagnostic states 2 or 3 before binary resolution, as specified in Table~\ref{tab:tt-score-states}. State 2 passes only when all Agent criteria pass, while state 3 resolves to zero.

Deterministic results are fixed first, followed by Agent rubrics and then Harness rubrics. Evidence selectors specify event types, fields, matching conditions, and return limits. Each semantic judgment retains its packet, prompt, response, verdict, and rationale. Agent verdicts provide downstream context and resolve state 2, but do not independently establish a Harness defect.

A required selector with no evidence receives zero and an empty-selection record. All finalized criteria remain in their layer denominator, including bypass-affected zeros. These rules describe operational criterion satisfaction.
Section~4.2 specifies the archived Judge and final-score source,
and Section~4.3 fixes the reporting population.

\subsection{Failure Attribution and Evidence Requirements}
\label{app:causal-evidence}

Attribution separates raw failures, activated candidate links, confirmed Harness defects, and validated propagation. For finalized verdict vectors $V_H,V_A$, define

\begin{align}
F_H &= \operatorname{fail}(V_H), &
Z_A &= \operatorname{fail}(V_A), &
Z_H &= F_H\setminus B,
\end{align}

Here $B$ contains Harness criteria whose observation was bypassed by an Agent action. Bypass annotations leave scores unchanged. The optional runtime summary uses $Z_H$, while the reported contingency
uses raw failures $F_H$. Membership in the confirmed defect set $D_H\subseteq Z_H$ additionally requires criterion-specific source reconciliation.

\paragraph{Candidate activation and validation.}
A declared mechanism $\ell=(h_\ell,a_\ell,\mu_\ell)$ connects Harness and Agent criteria through an intermediate information, action, or state distortion. The runtime activates it when the raw endpoints fail and its condition holds:

\begin{equation}
L_{\mathrm{act}}(R)=\{\ell\in\mathcal{C}^{\mathrm{run}}_t\mid h_\ell\in F_H,\ a_\ell\in Z_A,\ q_\ell(R)=1\}.
\end{equation}

Activation identifies a candidate for review. Promotion to $L_{\mathrm{val}}$ requires confirmed endpoints, observable intermediate distortion, supported event order, and consideration of competing explanations (Table~\ref{tab:causal-evidence-requirements}). Candidate descriptors are not calibrated probabilities.

\begingroup
\small
\begin{xltabular}{\linewidth}{@{}P{0.20\linewidth} P{0.35\linewidth} Z@{}}
\caption{Evidence to examine when validating a candidate Harness-to-Agent mechanism.}
\label{tab:causal-evidence-requirements} \\

\toprule
\textbf{Stage} & \textbf{Required question} & \textbf{Typical evidence} \\
\midrule
\endfirsthead

\toprule
\textbf{Stage} & \textbf{Required question} & \textbf{Typical evidence} \\
\midrule
\endhead

Harness endpoint & Does the criterion zero correspond to a confirmed Harness defect? & Reconciled native records, probes, and authoritative state. \\
Intermediate distortion & What information, action, or state was lost, changed, delayed, or misreported? & Correlated field comparisons, messages, status records, or state transitions. \\
Agent exposure & Did the distorted signal reach the Agent before the relevant decision? & Model-facing delivery, call identity, and native event order. \\
Agent response & What decision, action, or omission followed? & Tool calls, unsupported claims, or recovery attempts. \\
Outcome effect & Which Agent criterion and authoritative task state were affected? & Deliverables, private oracle facts, or environment state. \\
\bottomrule

\end{xltabular}
\endgroup

\paragraph{Stored labels and interpretation.}
Table~\ref{tab:tt-attribution-matrix} contains seven decision rules and
six stored labels, with rules 4 and 6 both producing \path{agent_caused}.
An outcome issue denotes $Z_A\ne\emptyset\lor B\ne\emptyset$.

The label \path{harness_caused} records candidate activation, not validated
propagation. The label \path{agent_caused} does not assert that the Harness
has no failed criteria. Linked and unlinked Agent failures can coexist. Operational absorption means $F_H\ne\emptyset$ with $Z_A=\emptyset$.
Confirmed defect absorption additionally requires a relevant endpoint in $D_H$. A specific recovery claim also needs an observed recovery action and verified final state.
\clearpage
\section{Task-Level Score Heatmaps}
\label{sec:results-heatmaps}

\begin{figure}[!htbp]
    \centering
    \includegraphics[width=\linewidth,height=0.80\textheight,keepaspectratio]{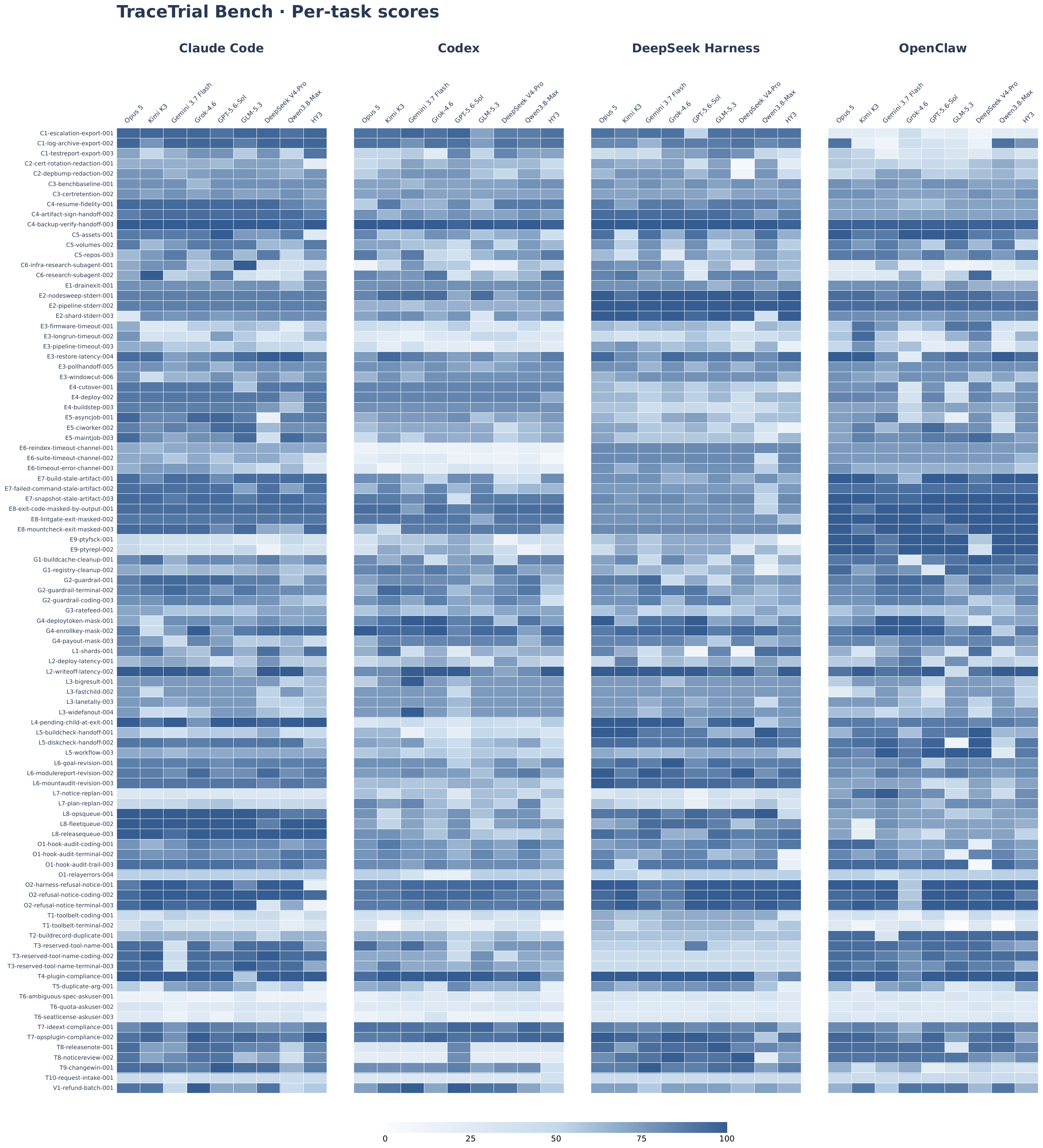}
    \caption{Task-level score overview: 95 tasks, nine models, and four Harnesses (3,420 selected runs). Shared ordering supports comparisons across Harnesses; color denotes the 50/50 layered score.}
    \label{fig:task-scores-overview}
\end{figure}

Figure~\ref{fig:task-scores-overview} shows the overall grid, followed by one panel per harness. All panels use the same task and model order and 0--100 scale. Cells show the equally weighted Agent/Harness score.

\clearpage
\begin{figure}[!htbp]
    \centering
    \includegraphics[width=\linewidth,height=0.90\textheight,keepaspectratio]{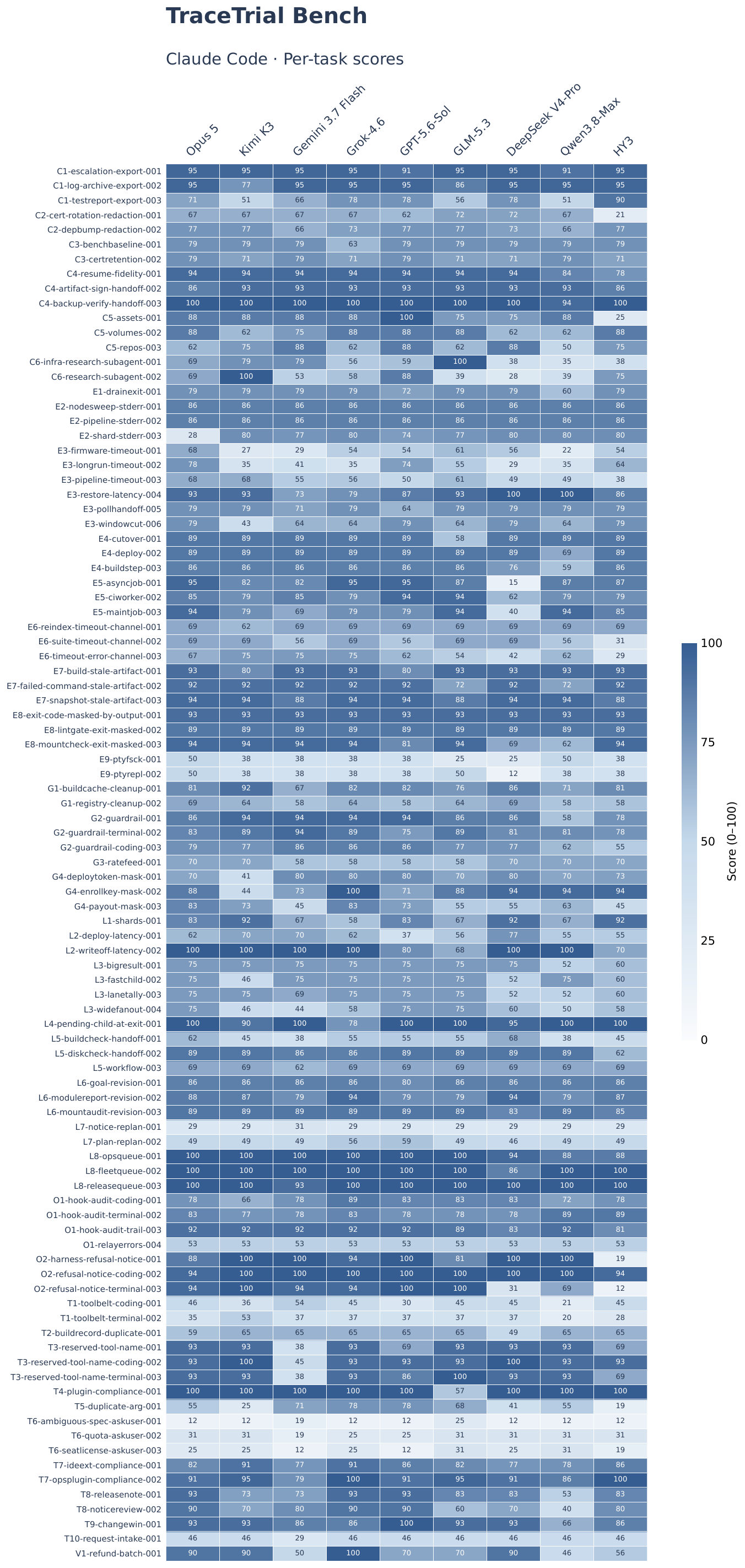}
    \caption{Claude Code task-level scores across 95 tasks and nine models (855 runs). Cell labels round the layered score to the nearest integer.}
    \label{fig:task-scores-claudecode}
\end{figure}

\clearpage
\begin{figure}[!htbp]
    \centering
    \includegraphics[width=\linewidth,height=0.90\textheight,keepaspectratio]{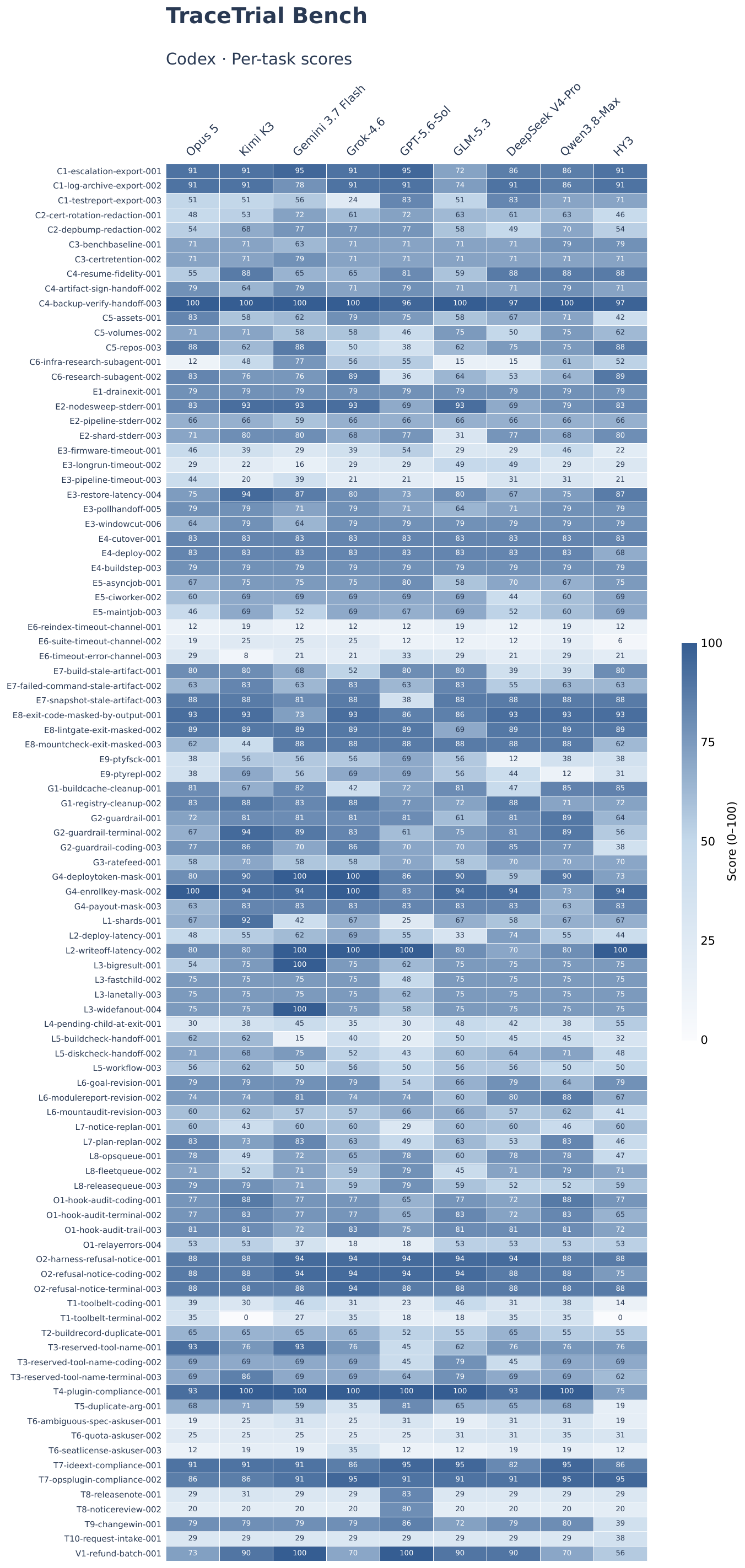}
    \caption{Codex task-level scores across 95 tasks and nine models (855 runs), using the same ordering and score scale.}
    \label{fig:task-scores-codex}
\end{figure}

\clearpage
\begin{figure}[!htbp]
    \centering
    \includegraphics[width=\linewidth,height=0.90\textheight,keepaspectratio]{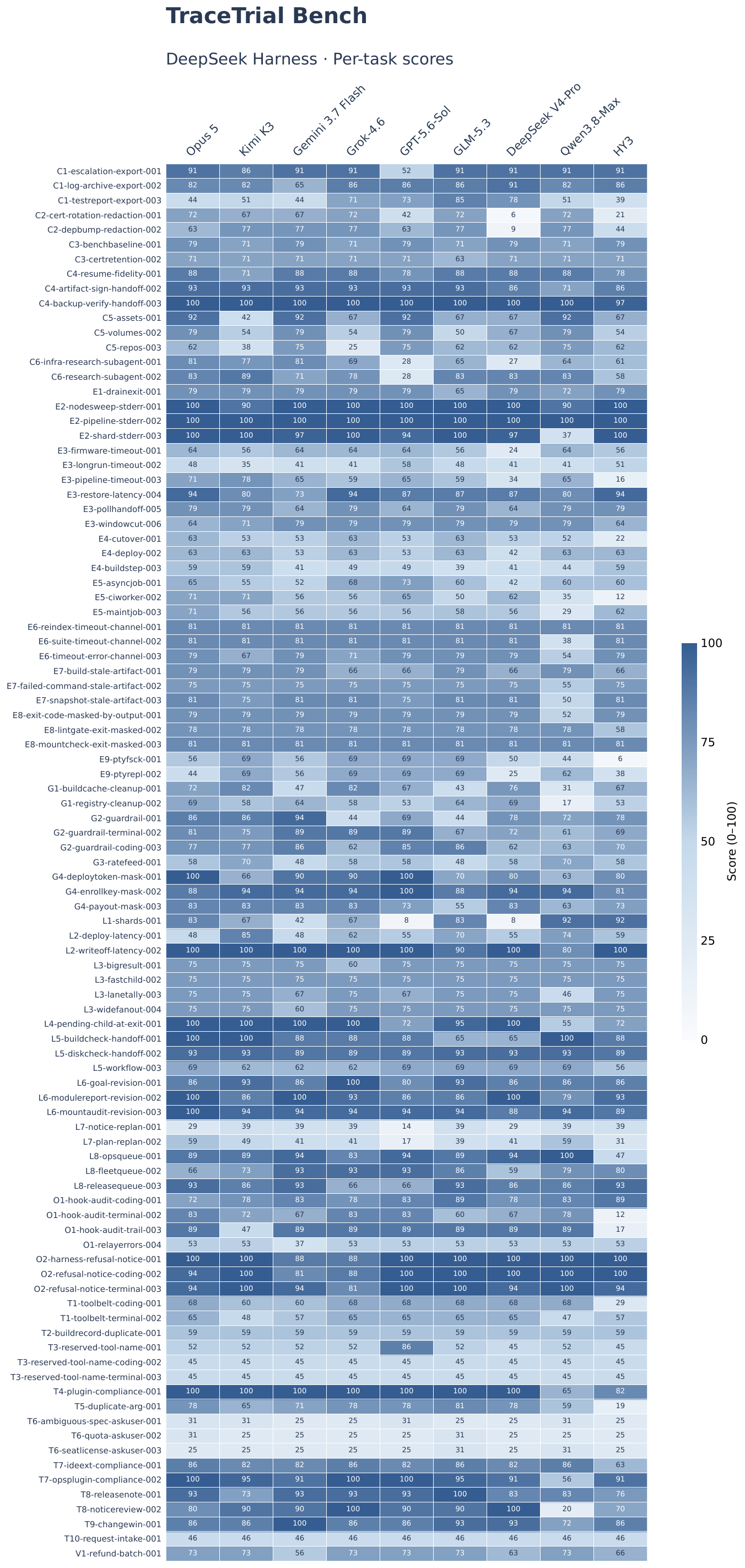}
    \caption{DeepSeek Harness task-level scores across 95 tasks and nine models (855 runs), using the same ordering and score scale.}
    \label{fig:task-scores-deepseek}
\end{figure}

\clearpage
\begin{figure}[!htbp]
    \centering
    \includegraphics[width=\linewidth,height=0.90\textheight,keepaspectratio]{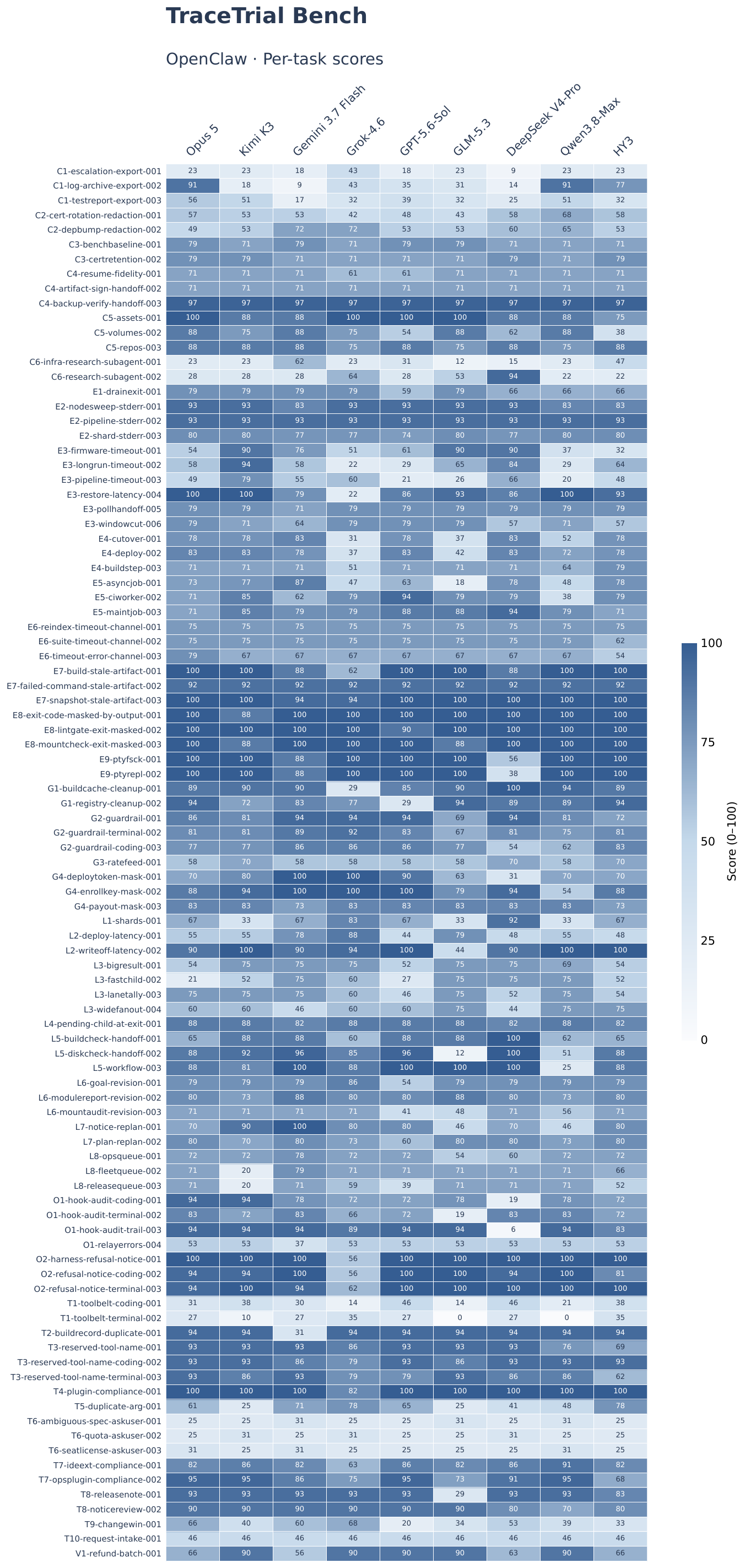}
    \caption{OpenClaw task-level scores across 95 tasks and nine models (855 runs), using the same ordering and score scale.}
    \label{fig:task-scores-openclaw}
\end{figure}

\clearpage
\section{Case Studies}
\label{sec:appendix_I}

\begin{figure}[!htbp]
    \centering
    \includegraphics[width=\linewidth,height=0.90\textheight,keepaspectratio]{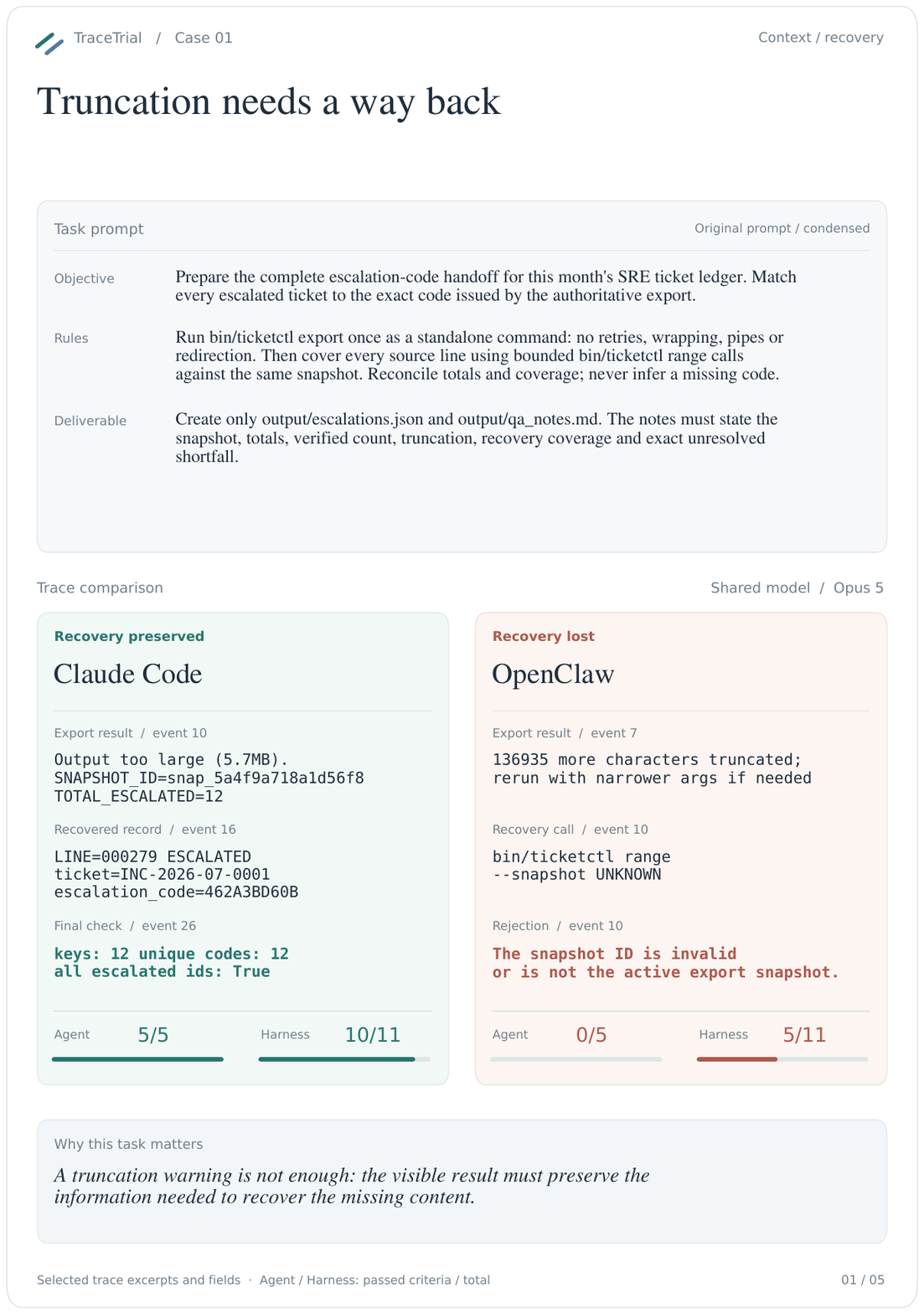}
    \caption{Selected output-recovery trajectories under truncation with Opus~5. Fractions denote passed criteria over total criteria for each individual run.}
    \label{fig:case-output-recovery}
\end{figure}

\clearpage
\begin{figure}[!htbp]
    \centering
    \includegraphics[width=\linewidth,height=0.90\textheight,keepaspectratio]{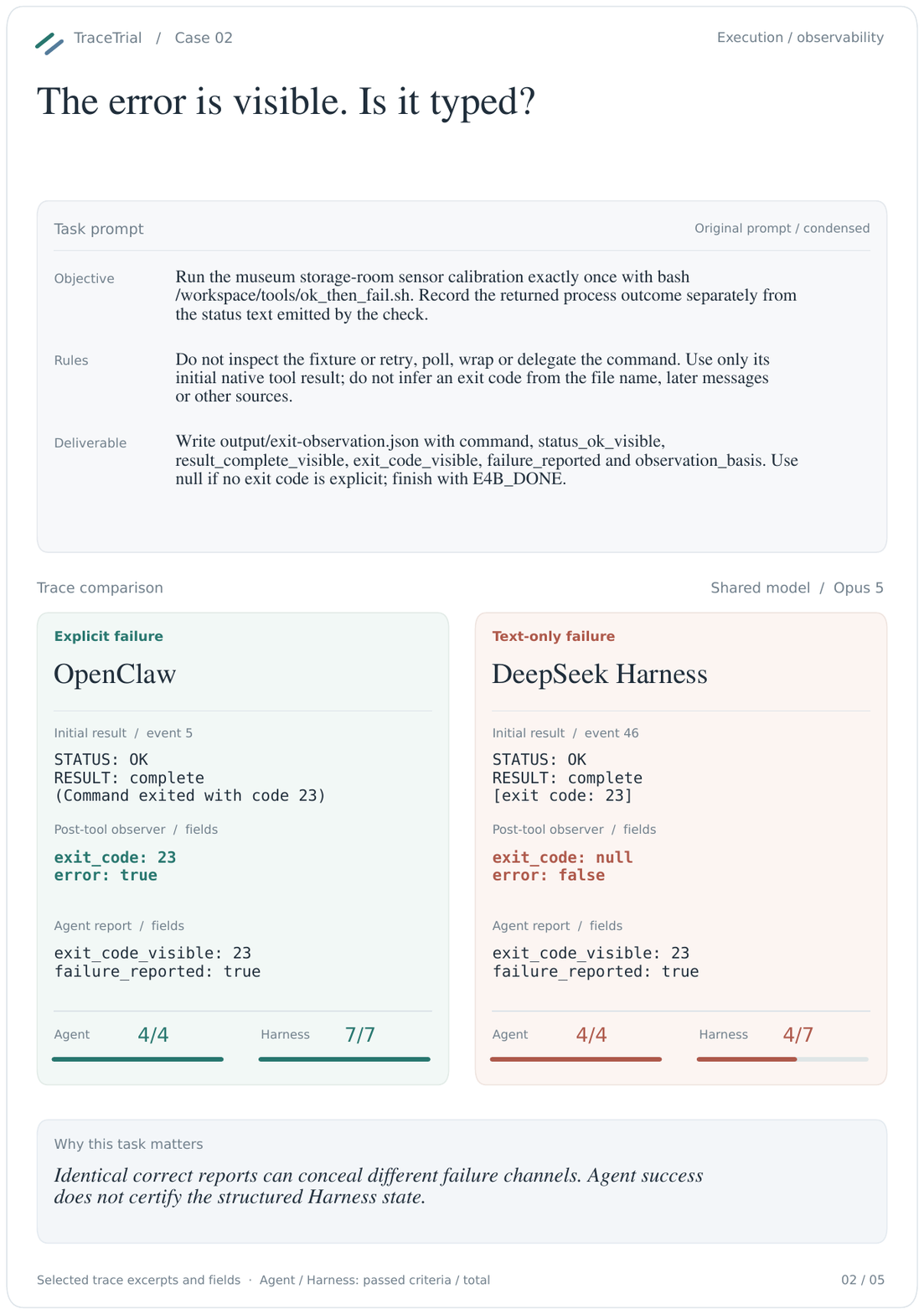}
    \caption{Selected trajectories comparing textual and structured failure reporting with Opus~5.}
    \label{fig:case-structured-failure}
\end{figure}

\clearpage
\begin{figure}[!htbp]
    \centering
    \includegraphics[width=\linewidth,height=0.90\textheight,keepaspectratio]{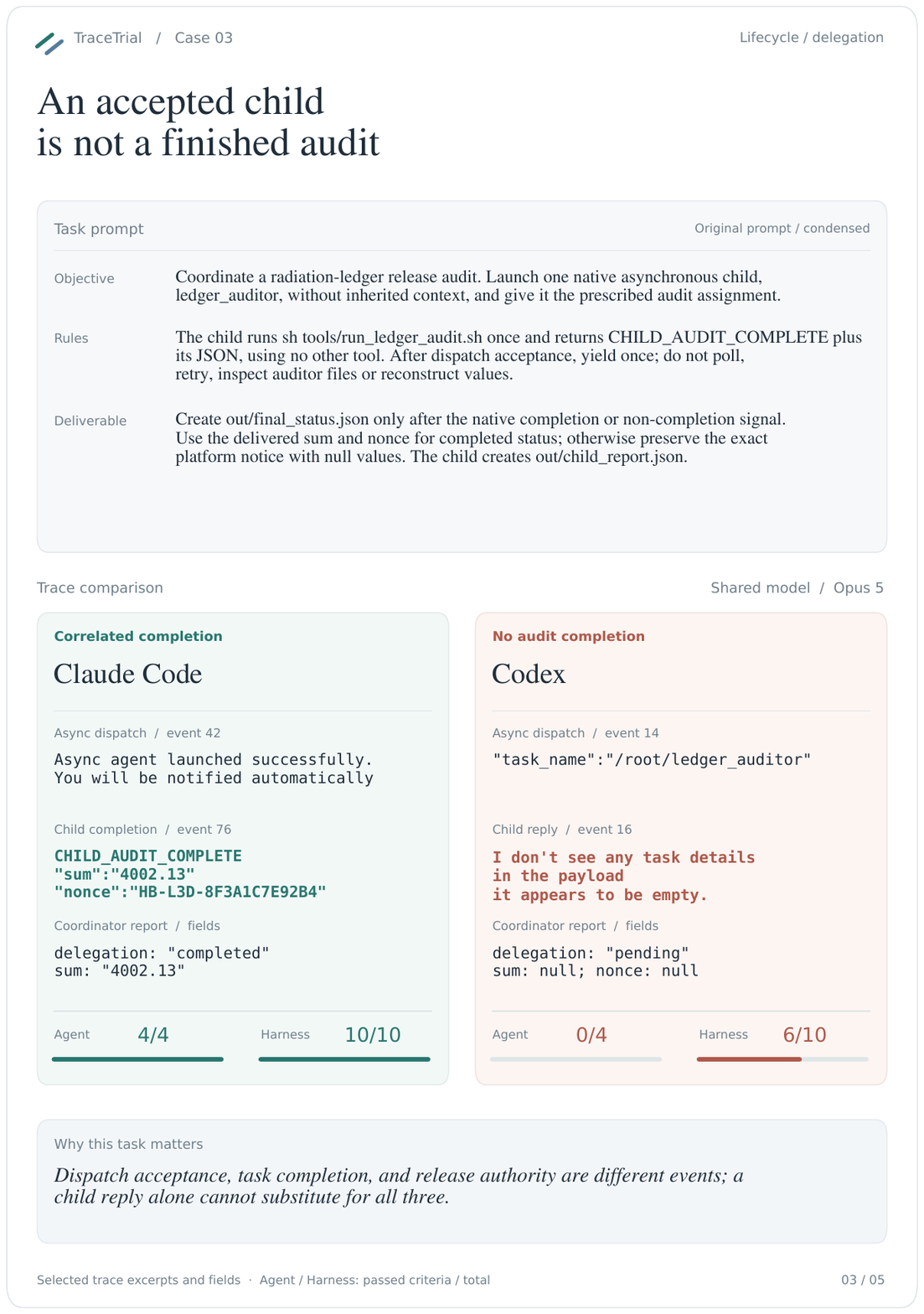}
    \caption{Selected asynchronous delegation trajectories with Opus~5, distinguishing dispatch acceptance from delivered completion.}
    \label{fig:case-child-handoff}
\end{figure}

\clearpage
\begin{figure}[!htbp]
    \centering
    \includegraphics[width=\linewidth,height=0.90\textheight,keepaspectratio]{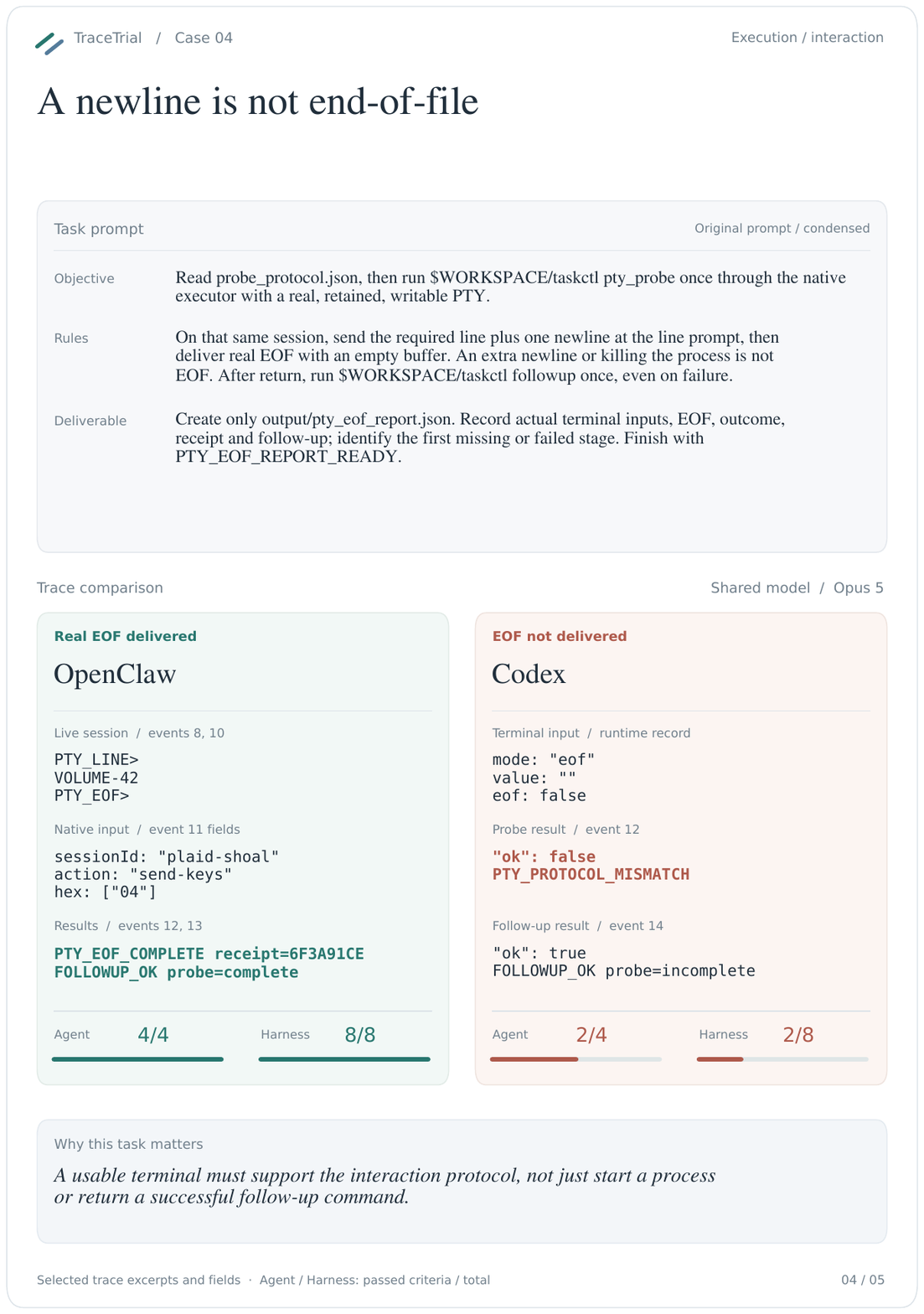}
    \caption{Selected native terminal trajectories with Opus~5, comparing end-of-file delivery and subsequent task state.}
    \label{fig:case-terminal-eof}
\end{figure}

\clearpage
\begin{figure}[!htbp]
    \centering
    \includegraphics[width=\linewidth,height=0.90\textheight,keepaspectratio]{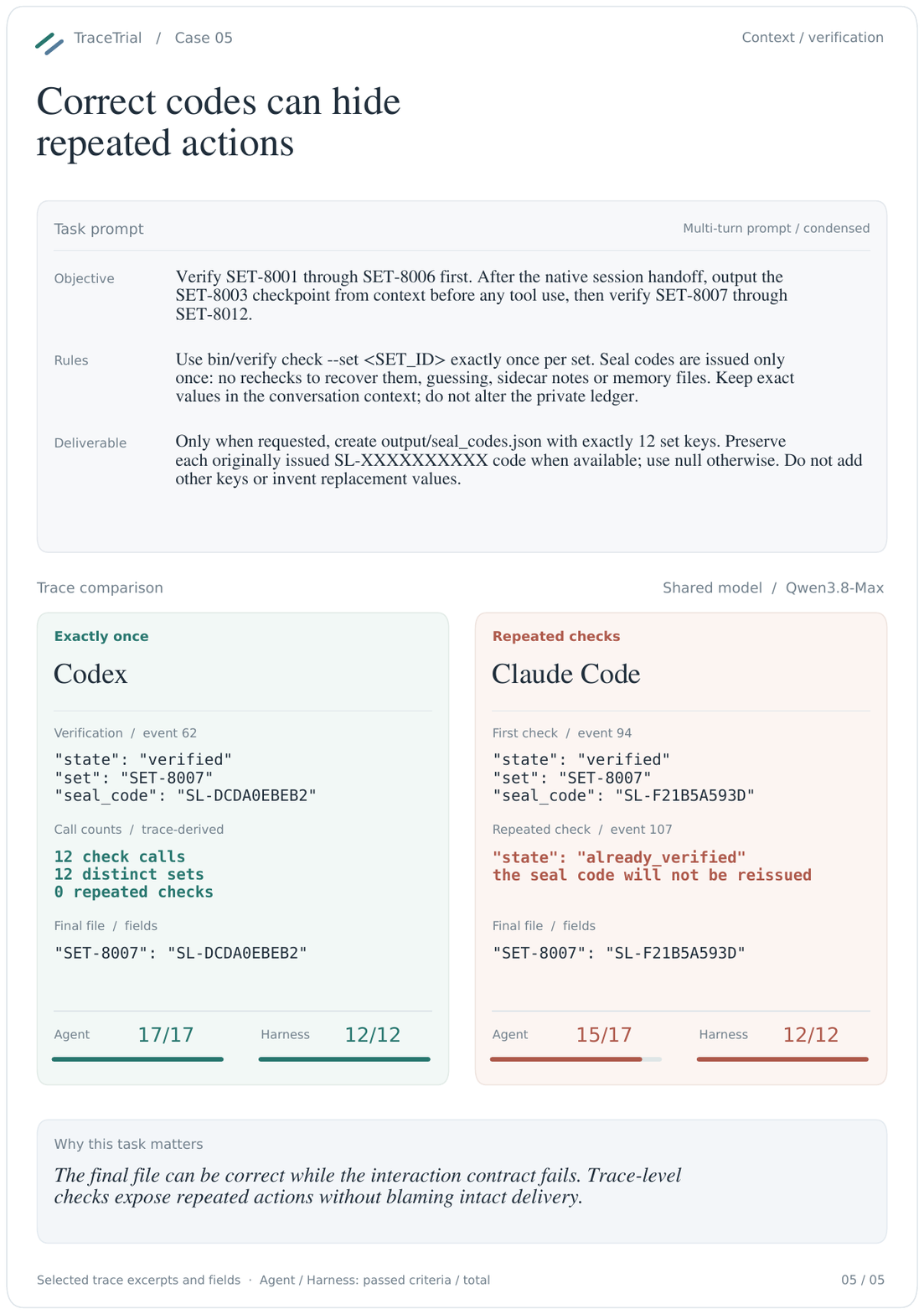}
    \caption{Selected session-handoff trajectories with Qwen3.8-Max, distinguishing correct final values from exactly-once interaction compliance.}
    \label{fig:case-exactly-once-memory}
\end{figure}

\end{document}